\documentclass[print,faculty=firw,department=elt,phddegree=elt]{adsphd}

\title{Designing High-Performing Networks for Multi-Scale Computer Vision}

\author{Cédric}{Picron}

\supervisor{Prof.~dr.~ir.~Tinne~Tuytelaars}{}
\president{Prof.~dr.~ir.~Jean~Berlamont}{}
\jurymember{Prof.~dr.~Matthew~Blaschko}{}
\jurymember{Prof.~dr.~ir.~Toon~Goedemé}{}
\jurymember{Prof.~dr.~ir.~Renaud~Detry}{}
\externaljurymember{Prof.~dr.~Efstratios~Gavves}{Universiteit van Amsterdam}

\researchgroup{Processing speech and images (ESAT-PSI)}
\website{} 
\email{} 

\address{Kasteelpark Arenberg 10 - box 2441}
\addresscity{B-3001 Leuven}

\date{January~\the\year}
\copyyear{\the\year}

\usepackage{glossaries} 
\newcommand{\glossname}{List of Abbreviations}
\newcommand{\myprintglossary}{%
  \renewcommand{\glossaryname}{\glossname}
  \cleardoublepage%
  \printglossary[title=\glossname]
  \chaptermark{\glossname}
  \addcontentsline{toc}{chapter}{\glossname} 
}
\makeglossaries

\usepackage{amsmath}
\usepackage{amssymb}
\usepackage{amsthm}
\usepackage{bibentry}
\usepackage{booktabs}
\usepackage{multirow}
\usepackage{pifont}
\usepackage{textcomp}
\usepackage{xspace}

\nobibliography*

\makeatletter
\DeclareRobustCommand\onedot{\futurelet\@let@token\@onedot}
\def\@onedot{\ifx\@let@token.\else.\null\fi\xspace}
\makeatother

\def\eg{\emph{e.g}\onedot} 

\def\ie{\emph{i.e}\onedot} 

\def\wrt{w.r.t\onedot}

\newcommand{\cmark}{\ding{51}}
\newcommand{\xmark}{\ding{55}}

\begin{document}


\makefrontcoverXII
\maketitle

\frontmatter

\includepreface{preface}
\includeabstract{abstract}
\includeabstractnl{abstractnl}

\newglossaryentry{AP}{name={AP}, description={Average Precision}}
\newglossaryentry{EffSeg}{name={EffSeg}, description={Efficient Segmentation method}}
\newglossaryentry{FFN}{name={FFN}, description={Feedforward network}}
\newglossaryentry{FQDet}{name={FQDet}, description={Fast-converging Query-based Detector}}
\newglossaryentry{FQDetV2}{name={FQDetV2}, description={Fast-converging Query-based Detector (V2)}}
\newglossaryentry{IBBR}{name={IBBR}, description={Iterative bounding box regression}}
\newglossaryentry{IoU}{name={IoU}, description={Intersection over Union}}
\newglossaryentry{MHA}{name={MHA}, description={Multi-head attention}}
\newglossaryentry{MLP}{name={MLP}, description={Multi-layer perceptron}}
\newglossaryentry{MSDA}{name={MSDA}, description={Multi-scale deformable attention}}
\newglossaryentry{NMS}{name={NMS}, description={Non-maximum suppression}}
\newglossaryentry{PQ}{name={PQ}, description={Panoptic Quality}}
\newglossaryentry{RoI}{name={RoI}, description={Region of Interest}}
\newglossaryentry{SOTA}{name={SOTA}, description={State-of-the-art}}
\newglossaryentry{SPS}{name={SPS}, description={Structure-Preserving Sparsity}}
\newglossaryentry{TPN}{name={TPN}, description={Trident Pyramid Network}}
\myprintglossary


\tableofcontents
\listoffigures
\listoftables


\mainmatter

  \graphicspath{{\chaptersdir/introduction/image/}}%
  \chapter{Introduction} \label{ch:introduction}

\section{Research setting} \label{sec:ch1_setting}

\paragraph{Computer vision.} Vision is undoubtedly one of the most important senses of the human species, providing an accurate understanding of its 3D  surroundings. Human vision can be broken down into two main steps: firstly the eyes capture the raw information from the 3D scene to form a 2D image, and secondly the brain processes the image by looking for specific patterns such as objects.

With the invention of digital cameras in the second half of the 20th century, humans found a way to capture the 3D surroundings in the form of a 2D image, similar to the human's eyes. However, a digital camera alone is not very useful as it only produces a 2D image, which in essence is simply a collection of numbers (\ie RGB values) organized in a 2D grid. A \textit{brain} is still required to make sense of these numbers. In some cases this \textit{brain} could be a human brain, but in many cases these images need to be automatically processed by a computer. 

Methods hence need to be designed to enable a computer to interpret the RGB values acquired from digital cameras. The field responsible for designing such methods is called \textit{computer vision}, with the aim to design methods with the highest possible level of image understanding at the lowest possible computational cost. 

Computer vision has been growing in popularity for the past decade, with more and more real-life applications relying on vision-based systems. We identify three main reasons why vision-based systems have become increasingly popular. Firstly, images carry a rich high-resolution signal with plenty of information not available from other physical signals. Secondly, digital cameras are small and cheap, which is different from many other sensors such as LIDAR. Thirdly, computer vision methods have significantly improved over the last decade due to the development of deep learning methods, enabling the use of vision-based systems for a much wider range of applications. 

It goes without saying that this surge in vision-based systems highlights the importance of designing high-performing computer vision methods. In this thesis, we do research on computer vision methods, and more precisely on the network part of computer vision models (see next paragraph).

\paragraph{Deep learning.} \label{par:ch1_deep_learning} Since the introduction of AlexNet~\cite{krizhevsky2012imagenet} in 2012, the field has transitioned from the use of hand-crafted features to deep learning methods, which train a computer vision model on a large set of images. The performance of these computer vision models is determined by three different factors: (1) the data, (2) the network architecture (or network for short), and (3) the training procedure.

The first factor is the \textbf{data}. The data comprises the input images provided to the models during both training and inference. These input images are determined by both the original dataset images as well as the used data augmentation and preprocessing schemes during training and inference. Additionally, the data also contains the ground-truth annotations (\ie the desired image predictions) for the different training images. Three different types of data improvements can hence be distinguished: (1) improving (\eg increasing) the original images dataset, (2) improving the data augmentation and preprocessing schemes, and (3) increasing the annotation quality.

The second factor is the network architecture or \textbf{network} for short. The network performs the processing by computing, from the input image RGB values, the losses during training and the image predictions during inference. The network itself is a combination of multiple base processing layers such as linear, convolutional~\cite{lecun1998gradient} and attention-based~\cite{vaswani2017attention} layers. The network is equipped with tunable knobs called \textit{parameters}, which allow the network to improve its predictions based on the mistakes it made during training.

The third factor is the \textbf{training procedure}. The training procedure is responsible for training the model by updating the network parameters. The training procedure is determined by three elements: (1) the optimizer computing the parameter updates from the parameter gradients, (2) the scheduler determining whether and when optimizer learning rates should be updated, and (3) the total number of training iterations or epochs (\ie the number of passes over the whole training dataset).

In this thesis, we focus on improving the network part of computer vision models, and more precisely on networks used for multi-scale computer vision tasks (see next paragraph).

\paragraph{Computer vision networks.} In many cases, the network design is tailored towards the computer vision task to be solved. We distinguish the networks designed for \textit{single-scale} and \textit{multi-scale} computer vision tasks.

\begin{figure}
    \begin{minipage}{0.29\textwidth}
        \centering
        \includegraphics[scale=0.148]{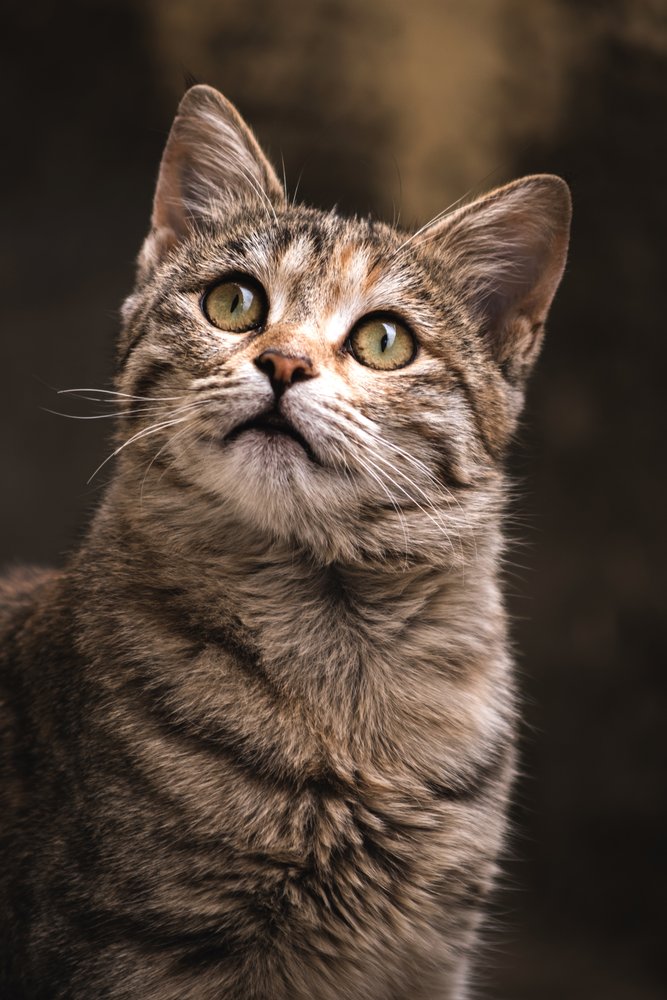}
    \end{minipage} \hfill
    \begin{minipage}{0.69\textwidth}
        \centering
        \includegraphics[scale=0.35]{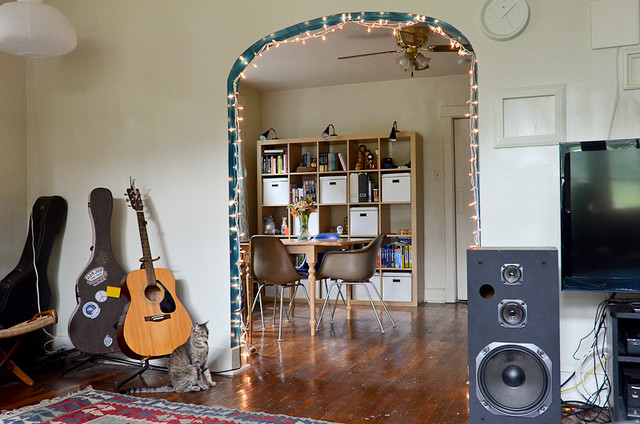}
    \end{minipage} \hfill
    \caption[Introduction - Example object-centric and scene-centric images]{Example object-centric (left) and scene-centric (right) images.}
    \label{fig:ch1_obj_scene}
\end{figure}

In \textbf{single-scale} (or object-centric) computer vision tasks, the input images are expected to be centered and cropped around a single object of interest (see for example left image in Figure~\ref{fig:ch1_obj_scene}), therefore naturally providing the right scale at which to do the computation. The most popular single-scale computer vision task is the image classification task, where models need to assign a class label (from a predefined set) to each of the object-centric input images. Given that all input images are expected to be at the same scale (\ie centered and cropped around the object of interest), network designs can be simplified by determining in advance how much computation is performed at each resolution. In this thesis, we refer to networks designed specifically for single-scale tasks, as \textit{backbones}.

In \textbf{multi-scale} (or scene-centric) computer vision tasks, the input images are general, such as everyday scenes possibly containing multiple objects at various scales (see for example right image in Figure~\ref{fig:ch1_obj_scene}). A wide range of computer vision tasks fall under this category, such as object detection, instance segmentation, and panoptic segmentation~\cite{kirillov2019panoptic}. Given that multiple scales are present in the input images, it is no longer possible to determine in advance how much computation should be performed at each resolution. Therefore, new network designs are required different from backbones, to better handle the multi-scale nature of the problem. We make the distinction between two types of multi-scale network designs: (1) task-independent designs called \textit{necks}, and (2) task-specific designs called \textit{heads}.

\textbf{Necks} are task-independent networks (or network parts to be precise) that can be used for any multi-scale computer vision task. This great flexibility comes at a price though. As necks are oblivious to the end goal, they are unable to focus their computation on those image areas that are important to solve the task. Necks are therefore by definition \textit{dense} processing networks, where the same amount of processing is done for each image region, trading efficiency for flexibility. In Chapter~\ref{ch:neck}, we present our own neck design called Trident Pyramid Network (\gls{TPN}).

\textbf{Heads} are task-specific networks (or network parts to be precise) that are designed for a specific computer vision task. This allows head networks to focus the computation on these image regions required to solve the task, leading to more efficient designs compared to necks. Heads hence trade flexibility for greater efficiency. In Chapter~\ref{ch:detection}, we present our own head design for the object detection task called Fast-converging Query-based Detector (\gls{FQDet}), and in Chapter~\ref{ch:segmentation} we present our own head design for the instance and panoptic segmentation tasks called Efficient Segmentation method (\gls{EffSeg}).

\begin{figure}
    \centering
    \includegraphics[scale=0.60]{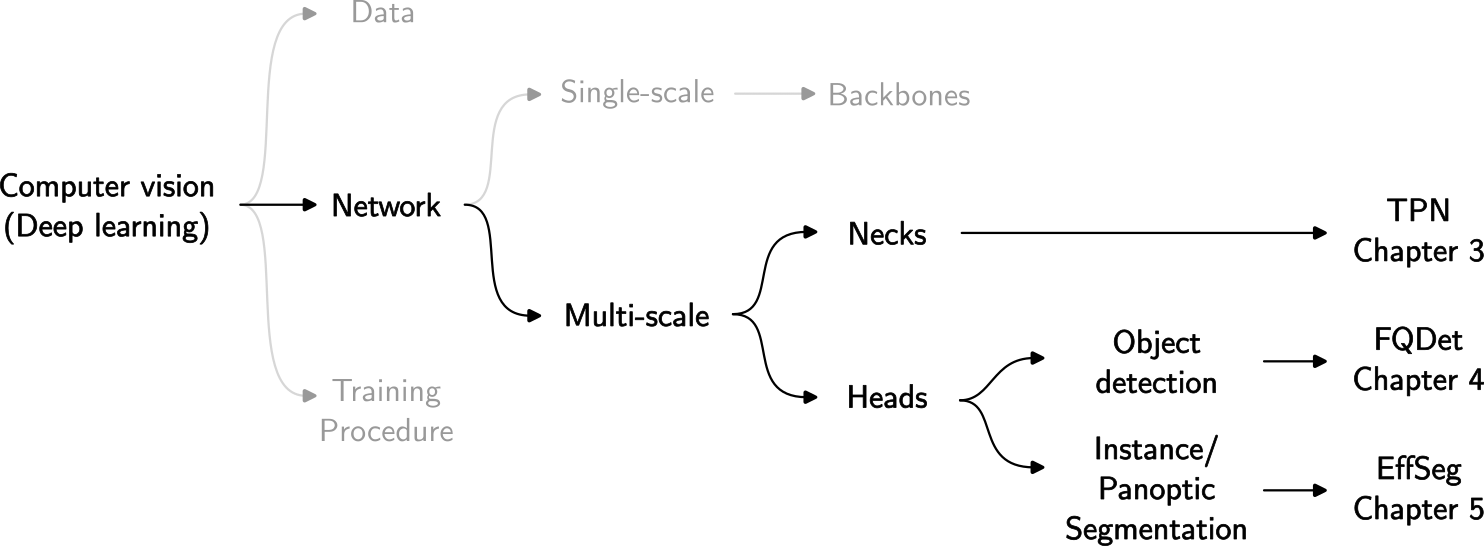}
    \caption[Introduction - Overview of research setting]{Overview of the research setting.}
    \label{fig:ch1_research_setting}
\end{figure}

\textbf{Overview.} An overview of the research setting is found in Figure~\ref{fig:ch1_research_setting}.

\section{General research goal}

The general research objective is to design high-performing networks and compare them with state-of-the-art (\gls{SOTA}) network designs on the multi-scale computer vision tasks of interest. Correctly comparing two distinct networks to assess which one works best, is more delicate than it appears and requires rigour. It is important to adhere to following two principles in particular.

Firstly, one must make sure that both networks use the \textbf{same deep learning settings}, \ie the same data pipeline and training procedure (see paragraph \nameref{par:ch1_deep_learning} in Section~\ref{sec:ch1_setting}). It is only when these settings are exactly the same, that a fair comparison can be made between two network designs.

Secondly, it is important to consider both \textbf{performance and computational cost metrics} when comparing two network designs. Simply comparing the performance metrics does not suffice, as it is often (if not always) possible to improve the performance by adding computation (\eg stacking more layers) to the network without changing its fundamental properties. It is hence possible to outperform a better network design with a worse one, by simply adding computation to the worse design until it outperforms the better one. For this reason, we always provide both performance metrics and computational cost metrics such as the number of FLOPs and the inference speed.

In order to evaluate our network designs, we identify one or multiple \gls{SOTA} baseline networks from the literature to compare with. Due to the varying deep learning settings used by many of these baseline networks, it is often not possible to directly comply with our first principle listed above, meaning that \textit{their} reported baseline performance metrics cannot be used. This is why we retrain the baseline networks using our own standardized deep learning settings, such that the obtained baseline performance metrics can be compared with the performance metrics of our own network designs. We use publicly available implementations for the various baseline networks, such as those provided by MMDetection~\cite{mmdetection}.

\begin{figure}
    \centering
    \includegraphics[scale=0.5]{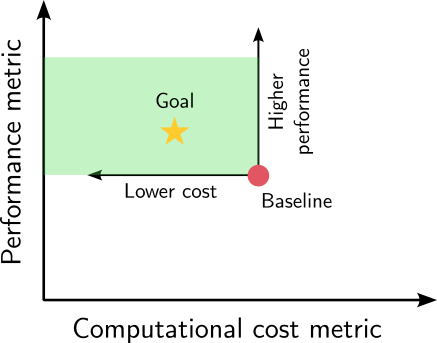}
    \caption[Introduction - General research goal]{The general research goal is to design networks that lie top left (\ie in the green region) of the baseline network in the performance vs. cost graph.}
    \label{fig:ch1_research_goal}
\end{figure}

We say a network design outperforms another one if it achieves similar or better performance with lower or similar computational cost. Our goal is hence to lie top left of the baseline network (or networks) in the performance vs. cost graph (see Figure~\ref{fig:ch1_research_goal}). Note that our new network designs are fundamentally different compared to the baseline designs, by introducing new components or by combining them in a novel way. Techniques that reduce the computational cost of existing networks without fundamental design changes such as network compression~\cite{li2016pruning, molchanov2016pruning} and quantization~\cite{wu2016quantized}, are orthogonal to our research endeavors, and are not used by either our networks or by the baseline networks.

\section{Thesis overview and contributions}

In what follows, we provide an overview of the overall structure of the thesis, together with the main contributions.

\paragraph{Chapter~\ref{ch:background}.} In Chapter~\ref{ch:background}, we provide some \textbf{background} information about the fundamentals on designing and evaluating networks for multi-scale computer vision tasks. The background chapter covers following four topics: (1)~the feature pyramid data structure along with the backbone-neck-head meta architecture, (2)~the multi-scale computer vision tasks of interest with the corresponding datasets, (3)~some general information about the used performance and computational cost metrics, and (4)~the standardized deep learning settings used throughout the thesis.  

\paragraph{Chapter~\ref{ch:neck}.} In Chapter~\ref{ch:neck}, we introduce our own \textbf{neck} network called \gls{TPN}. The TPN design introduces hyperparameters to control the amount of content-based and communication-based processing. We investigate which combination works best and compare with the BiFPN~\cite{tan2020efficientdet} baseline. Additionally, we also perform experiments to evaluate whether it is more beneficial to put additional computation into the neck rather than into the backbone or head. Our findings are published in following conference proceeding:

\begin{quote}
    \bibentry{picron2022trident}
\end{quote}

\paragraph{Chapter~\ref{ch:detection}.} In Chapter~\ref{ch:detection}, we design our own \textbf{object detection head} network called \gls{FQDet}. FQDet eliminates the main limitations of the DETR~\cite{carion2020end} detector (\ie slow convergence and poor performance on small objects), by reintroducing anchors in the context of query-based detectors and by devising a novel top-k matching scheme. Our FQDet detector is compared against four different baseline detectors: Faster R-CNN~\cite{ren2015faster}, Cascade R-CNN~\cite{cai2018cascade}, Deformable DETR~\cite{zhu2020deformable}, and Sparse R-CNN~\cite{sun2021sparse}. Our findings are published at:

\begin{quote}
    \bibentry{picron2022fqdet}
\end{quote}

\paragraph{Chapter~\ref{ch:segmentation}.} In Chapter~\ref{ch:segmentation}, we design our own \textbf{segmentation head} network called \gls{EffSeg}. EffSeg introduces the Structure-Preserving Sparsity (\gls{SPS}) data structure, which allows for structured 2D operations such as 2D convolutions to be applied in sparse set of image locations in an efficient way. Leveraging the SPS data structure enables EffSeg to perform computation only there where it is needed (typically near the segmentation mask boundaries), leading to improved efficiency while maintaining excellent performance. Our EffSeg segmentation method is compared against three different baseline segmentation methods: Mask R-CNN~\cite{he2017mask}, PointRend~\cite{kirillov2020pointrend}, and RefineMask~\cite{zhang2021refinemask}. Our findings are (partly) found in following pre-print:

\begin{quote}
    \bibentry{picron2023effseg}
\end{quote}

\paragraph{Chapter~\ref{ch:conclusion}.} In Chapter~\ref{ch:conclusion}, we \textbf{conclude} the thesis by giving an overview of the main contributions. Additionally, we also discuss the limitations of the work and provide directions for future research.

\cleardoublepage
  \graphicspath{{\chaptersdir/background/image/}}%
  \chapter{Background} \label{ch:background}

In this chapter, we provide some background knowledge related to the main content of the thesis. In Section~\ref{sec:ch2_meta_arch}, we first introduce the feature map and feature pyramid data structures, as well as the backbone-neck-head meta architecture. We continue in Section~\ref{sec:ch2_tasks} by describing the different multi-scale computer vision tasks considered throughout the thesis, alongside the datasets used for each of these tasks. In Section~\ref{sec:ch2_metrics}, we explain various performance and computational cost metrics that are used to evaluate the different network designs. Finally, in Section~\ref{sec:ch2_settings}, we provide some standardized deep learning settings common to many (if not all) experiments performed in this thesis.

\section{Backbone-neck-head meta architecture} \label{sec:ch2_meta_arch}

\paragraph{Feature map.} A feature map is a collection of feature vectors or \textit{features} organized in a 2D grid or \textit{map}, represented by a 4D tensor of shape $[B, \,F, \,M_H, \,M_W]$. Here $B$ is the batch size (\ie the number of images processed in parallel), $F$ the feature (vector) size or the number of channels, $M_H$ the map height, and $M_W$ the map width.

Feature maps play an important role when processing dense 2D data structures such as images. Many base 2D operations namely utilize feature maps by taking a feature map as input and outputting an updated feature map. Popular 2D operations of this sort are convolutional~\cite{lecun1998gradient, he2016deep, liu2022convnet} and attention-based~\cite{dosovitskiy2020image, liu2021swin} operations. Note that the used terminology might differ from the one used in the attention-based literature~\cite{dosovitskiy2020image, liu2021swin}, where features are called \textit{tokens} or \textit{patches} instead.

When combining multiple of these base 2D operations, a network is obtained called the \textbf{backbone}. The input of the backbone is a batch of images, which in itself can be considered as a feature map of shape $[B, \,3, \,I_H, \,I_W]$, with $I_H$ and $I_W$ the image height and width respectively. The intermediate and output data structures of the backbone are also feature maps, obtained by applying various operations or layers on the input feature map, such as linear layers, activation layers~\cite{nair2010rectified, hendrycks2016gaussian}, normalization layers~\cite{ioffe2015batch, ba2016layer}, pooling layers~\cite{krizhevsky2012imagenet, he2016deep}, skip connections~\cite{he2016deep}, convolution layers~\cite{lecun1998gradient, he2016deep, liu2022convnet}, and attention layers~\cite{dosovitskiy2020image, liu2021swin}. Backbones are hence networks operating on feature maps and are also called \textbf{single-scale} networks, as feature maps only contain a single resolution.

The backbone network can be subdivided into a \textit{stem} and multiple consecutive \textit{stages}. The \textbf{stem} is the first part of the backbone network which usually aggressively decreases the resolution of the input images to an appropriate feature map size due to the redundancy inherent to natural images~\cite{liu2022convnet}. This decrease in resolution or downsampling is achieved by either using convolution and pooling layers with stride greater than one~\cite{he2016deep}, or by using a \textit{patchify} approach dividing the image into regions or patches~\cite{dosovitskiy2020image, liu2021swin, liu2022convnet}. However, note that both approaches are actually not that different, as the patchify approach could be considered as a non-overlapping convolution with kernel size equal to the patch resolution~\cite{liu2022convnet}.

After the backbone stem, follow one~\cite{dosovitskiy2020image} or multiple~\cite{he2016deep, liu2021swin, liu2022convnet} backbone \textbf{stages}. Each stage (except the first) starts by downsampling the feature map by a factor two, followed by multiple layers updating the feature map while leaving the resolution unchanged. A stage is hence characterized by the feature map resolution it operates on, with earlier stages working on high-resolution feature maps and later stages working on low-resolution feature maps. Single-stage backbones operating on a single resolution are called \textit{plain} backbones, and multi-stage backbones operating on multiple resolutions are called \textit{hierarchical} backbones~\cite{li2022exploring}.

\paragraph{Feature pyramid.} When working on scene-centric or multi-scale computer vision tasks, it is important that the data structure satisfies following two conditions: (1) sufficiently high-resolution feature maps must be available to deal with fine-grained items such as small objects, and (2) efficient long-range information exchange must be supported to handle image-wide concepts such as large objects.

The most popular data structure satisfying these two conditions, is the \textbf{feature pyramid}~\cite{lin2017feature} data structure. A feature pyramid is a set of feature maps organized from high-resolution to low-resolution, where each map has half the resolution compared to the previous one. To know which resolutions are present in the feature pyramid, feature maps are identified by $P_l$, where $l$ is the number of times the feature map was downsampled (by a factor two) \wrt the original image resolution. A popular choice for the feature pyramid is to consider feature maps $\{P_3, \,P_4, \,P_5, \,P_6, \,P_7\}$, forming the $P_3$-$P_7$ feature pyramid~\cite{lin2017focal}.

Networks operating on feature pyramids are called \textbf{necks}. Necks are hence analogous to backbones, except that they work on feature pyramids instead of feature maps. In addition to the operations and layers found in backbones, necks also use top-down~\cite{lin2017feature}, bottom-up~\cite{liu2018path} and multi-scale~\cite{zhu2020deformable} operations. Given the multiple resolutions present in feature pyramids, necks are also called \textbf{multi-scale} networks.

\begin{figure}
    \centering
    \includegraphics[scale=0.75]{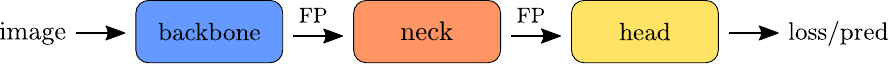}
    \caption[Background - Backbone-neck-head meta architecture]{High-level view of a multi-scale computer vision network following the backbone-neck-head meta architecture. The backbone (\textit{left}) processes the image to output a feature pyramid. The neck (\textit{middle}) takes in a feature pyramid (denoted by FP) and returns an updated feature pyramid. Finally, the task-specific head (\textit{right}) produces the loss during training and makes predictions during inference from the final feature pyramid.}
    \label{fig:ch2_backbone_neck_head}
\end{figure}

\paragraph{Backbone-neck-head meta architecture.} In this thesis, we design networks for multi-scale computer vision tasks such as object detection and instance segmentation. The networks first consist of a task-independent part, followed by a task-specific part.

For the \textbf{task-independent} part, both single-scale backbone or multi-scale neck networks could be used. Given that we are interested in solving multi-scale computer vision tasks, it might seem natural to only use a neck network and no backbone network. However, there is an important reason why backbone networks are still useful in a multi-scale settings. Backbones namely come pretrained, \ie they contain a set of high-quality network weights obtained by solving a different task such as image classification on ImageNet~\cite{russakovsky2015imagenet}. This is in contrast with necks which are not pretrained, and therefore do not have access to high-quality network weights at the beginning of training. Solely relying on a neck network hence means that the whole task-independent part of the network would be randomly initialized, leading to slow convergence during training.

For the task-independent part of the network, in order to avoid training for many epochs, a \textbf{pretrained backbone} is therefore used in conjunction with a randomly initialized \textbf{neck}. However, there is still a problem. The backbone namely outputs a feature map, but the neck expects a feature pyramid as input. To solve this issue, some small modifications are made to the backbone to also output a feature pyramid. For plain backbones, the feature pyramid is constructed by applying convolutions and transposed convolutions on the output feature map~\cite{li2022exploring}. For hierarchical backbones, the feature pyramid is built by collecting the final feature maps from the desired backbone stages and by applying convolutional layers to obtain even lower resolution feature maps if needed~\cite{lin2017feature}.

For the \textbf{task-specific} part of the network, a multi-scale \textbf{head} network is used. The head computes the loss during training and makes predictions during inference from the feature pyramid outputted by the task-independent part. Note that the head could contain multiple parallel sub-heads, such as the classification and bounding box regression sub-heads found in an object detection head. 

When we put both the task-independent and the task-specific part together, we obtain the backbone-neck-head network blueprint or meta architecture. An overview of the backbone-neck-head meta architecture is found in Figure~\ref{fig:ch2_backbone_neck_head}. In Chapter~\ref{ch:neck}, we design our own neck network, and in Chapters~\ref{ch:detection} and \ref{ch:segmentation} we design our own head networks for the object detection and segmentation tasks respectively.

\section{Multi-scale computer vision tasks and datasets} \label{sec:ch2_tasks}

\begin{figure}
    \centering
    \includegraphics[scale=0.19]{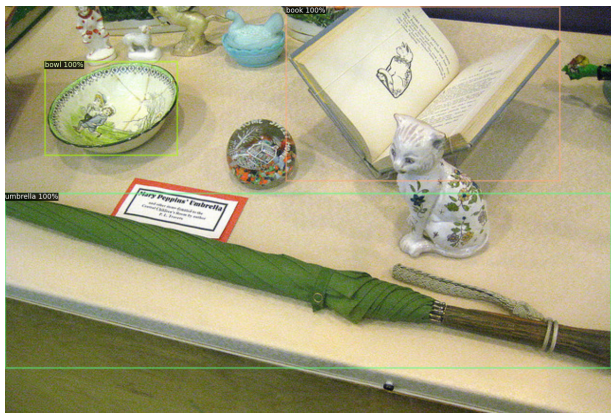}
    \caption[Background - Object detection example image]{Example image with the (ground-truth) object detection annotations.}
    \label{fig:ch2_obj_det}
\end{figure}

\paragraph{Object detection.} Object detection is a multi-scale computer vision task consisting of detecting objects in images by predicting the smallest axis-aligned bounding boxes containing each of the objects, together with their corresponding class labels. In this thesis, we consider the \textbf{closed-set object detection} setting, where only objects from a predefined set of object classes need to be detected. This is in contrast with open-set object detection which also attempts to detect objects from unknown or novel object classes not annotated during the training process~\cite{joseph2021towards, zheng2022towards}.

An example image together with the (ground-truth) object detection annotations is found in Figure~\ref{fig:ch2_obj_det}. Note that each object detection prediction also contains a confidence score between $0$ and $1$. This allows the different predictions to be ranked from high-confident to low-confident during the computation of the object detection performance metrics (see Section~\ref{sec:ch2_metrics}), and to prune low-confident predictions during \eg visualization.

Our object detection experiments are performed on the popular \textbf{COCO} (Common Objects in Context) dataset~\cite{lin2014microsoft}. The 2017 COCO dataset contains roughly $123$k annotated images of everyday scenes with a total of $896$k bounding box detections corresponding to $80$ different object classes~\cite{shao2019objects365}. Figure~\ref{fig:ch2_obj_det} contains an example image from the COCO dataset together with the provided ground-truth object detection annotations. During our experiments, we train on the 2017 training split containing roughly 118k images, and evaluate on the 2017 validation and test-dev splits containing 5k and 20k images respectively. The object detection task is featured in Chapters~\ref{ch:neck} and \ref{ch:detection} of the thesis.

\begin{figure}
    \centering
    \includegraphics[scale=0.19]{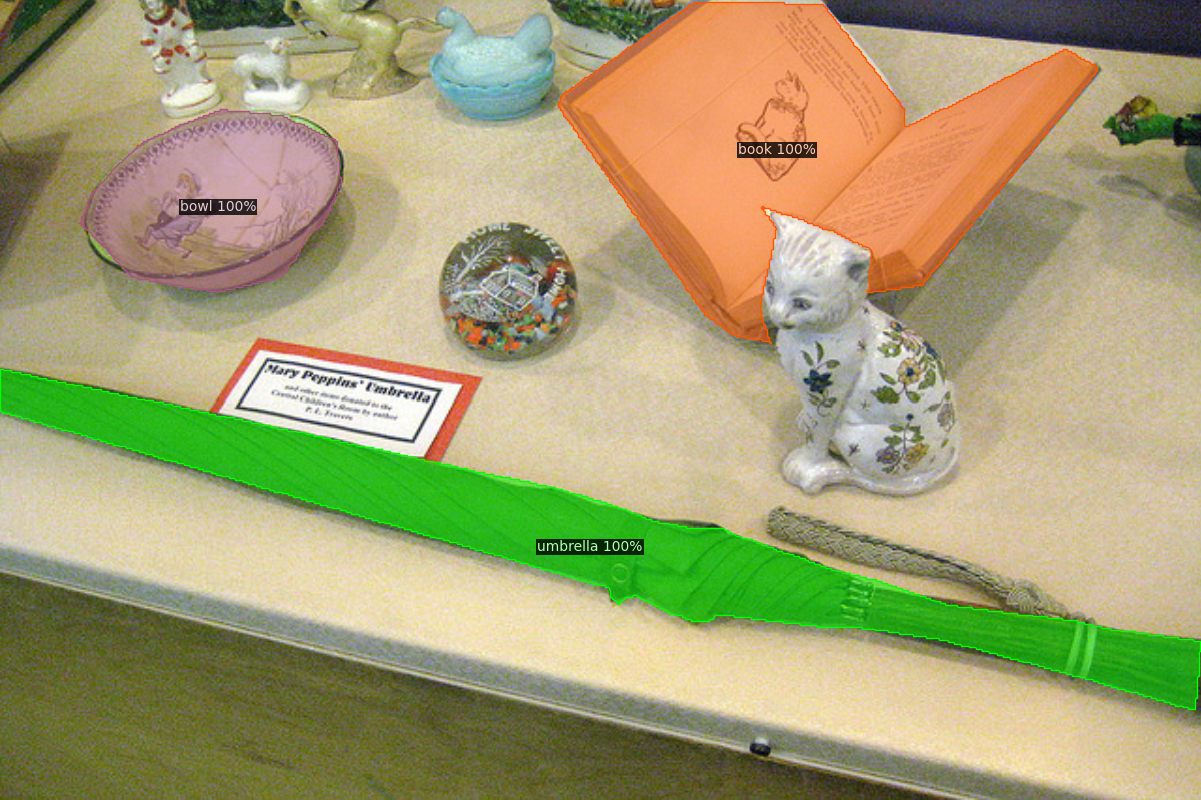}
    \caption[Background - Instance segmentation example image]{Example image with the (ground-truth) instance segmentation annotations.}
    \label{fig:ch2_ins_seg}
\end{figure}

\paragraph{Instance segmentation.} Instance segmentation is a multi-scale computer vision task consisting of detecting objects in images by predicting the object segmentation masks, together with their corresponding class labels. The instance segmentation task is hence similar to the object detection task, except that it identifies each object with a segmentation mask instead of with the smallest axis-aligned bounding box containing the object. For the instance segmentation task, we also consider the closed-set setting where the object classes to be detected are fixed and known in advance. The same example image as in Figure~\ref{fig:ch2_obj_det} but this time with (ground-truth) instance segmentation annotations, is found in Figure~\ref{fig:ch2_ins_seg}.

Our instance segmentation experiments are also be performed on the 2017 \textbf{COCO} dataset~\cite{lin2014microsoft}. Similar to the object detection experiments, we train on the 2017 training split containing roughly 118k images, and evaluate on the 2017 validation split containing 5k images.

\begin{figure}
    \begin{minipage}{0.49\textwidth}
        \centering
        \includegraphics[scale=0.2]{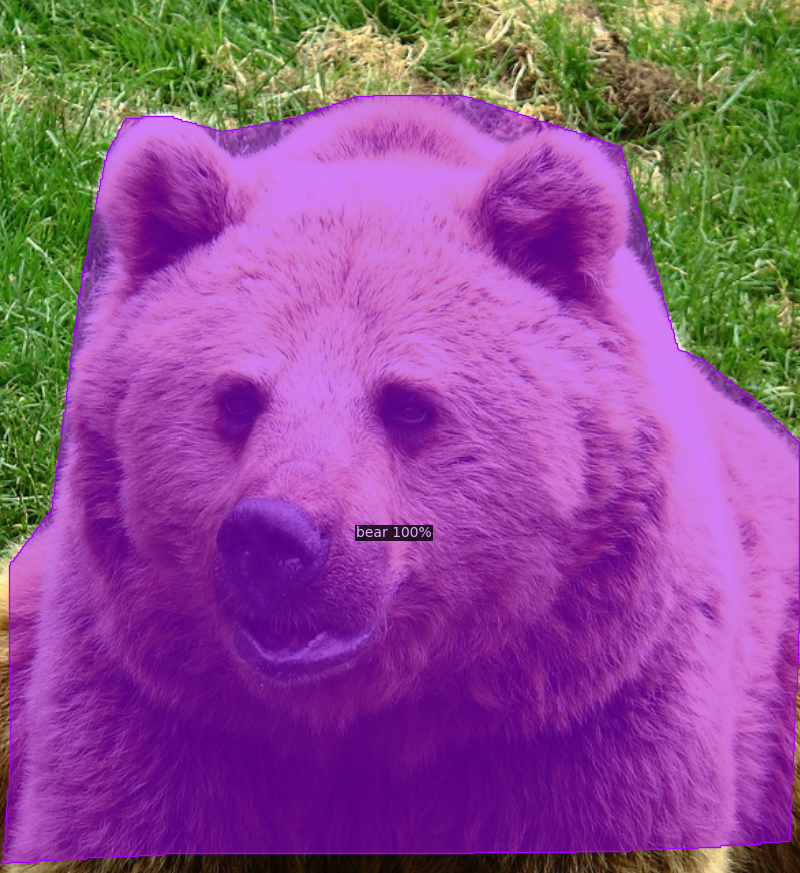}
    \end{minipage} \hfill
    \begin{minipage}{0.49\textwidth}
        \centering
        \includegraphics[scale=0.2]{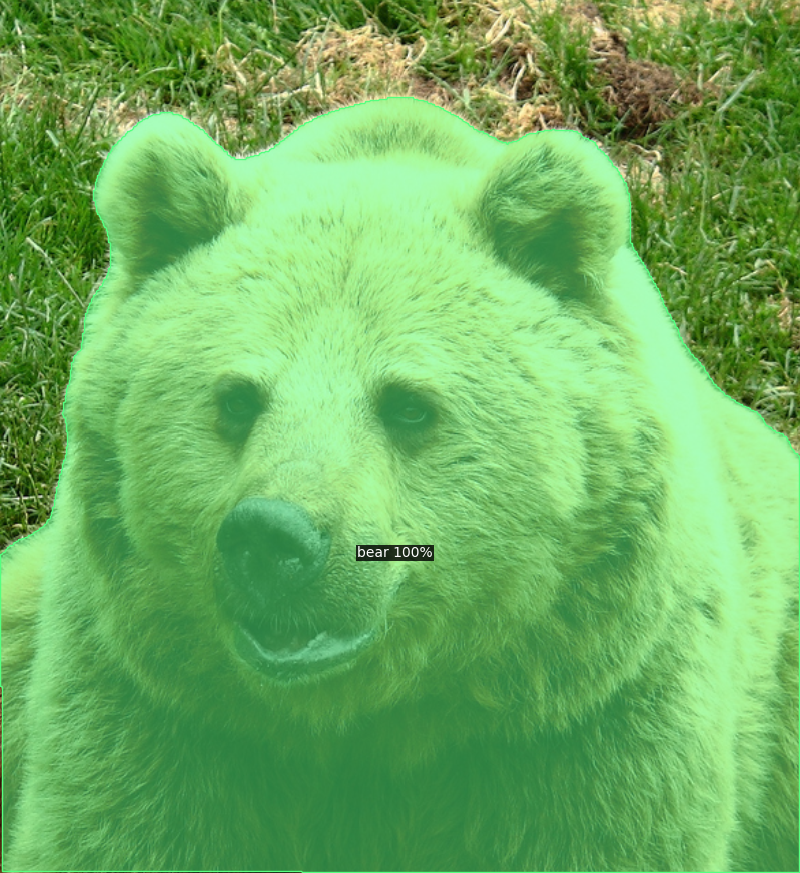}
    \end{minipage} \hfill
    \caption[Background - COCO vs. LVIS example mask]{Example images showing the difference in mask quality between the low-quality COCO mask (\textit{left}) and the high-quality LVIS mask (\textit{right}).}
    \label{fig:ch2_coco_lvis_mask}
\end{figure}

Additionally, we also evaluate on the 2017 COCO validation images using \textbf{LVIS}~\cite{gupta2019lvis} annotations as opposed to the original COCO annotations. This is because the LVIS annotations provide higher-quality segmentation masks compared to the COCO segmentation masks (see Figure~\ref{fig:ch2_coco_lvis_mask}), enabling a better evaluation on the fine-grained quality of the predicted segmentation masks. The instance segmentation task is featured in Chapter~\ref{ch:segmentation} of the thesis.

\begin{figure}
    \centering
    \includegraphics[scale=0.21]{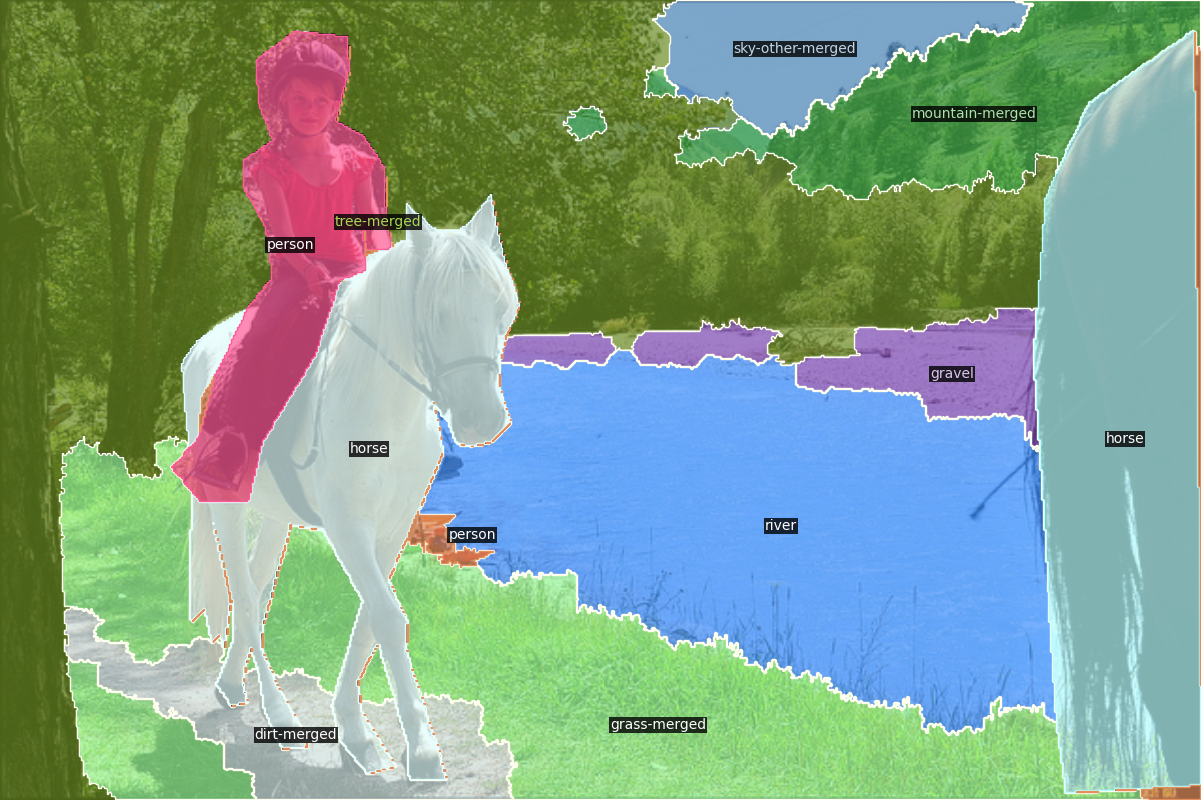}
    \caption[Background - Panoptic segmentation example image]{Example image with the (ground-truth) panoptic segmentation annotations.}
    \label{fig:ch2_pan_seg}
\end{figure}

\paragraph{Panoptic segmentation.} Panoptic segmentation~\cite{kirillov2019panoptic} is a multi-scale computer vision task which consists in assigning each image pixel either to an instance of a thing class, or to a stuff class. Here, \textit{thing} classes correspond to countable objects, such as people, animals, and tools, whereas \textit{stuff} classes correspond to stuff that is uncountable, such as grass, sky, and dirt. Similar to previous tasks, we consider the closed-set setting where the thing and stuff classes are fixed and known in advance. In Figure~\ref{fig:ch2_pan_seg}, an example image is shown with the (ground-truth) panoptic segmentation annotations.

The panoptic segmentation task can be considered as an extension of both the instance and semantic segmentation tasks. On the one hand, panoptic segmentation extends the instance segmentation task by not only considering thing classes, but also stuff classes, allowing for a more accurate segmentation of the background region (compared to leaving to background region as is). On the other hand, panoptic segmentation extends the semantic segmentation task by making the distinction between different instances of the same thing class (\eg between different persons). Hence, the panoptic segmentation task can be considered as the most general image segmentation task, segmenting each pixel in the image (as in semantic segmentation), and distinguishing between different objects of the same class (as in instance segmentation).

The panoptic segmentation experiments are also performed on the 2017 COCO dataset~\cite{lin2014microsoft}, where we train on the 2017 training split containing roughly 118k images, and evaluate on the 2017 validation split containing 5k images. The panoptic segmentation task is featured in Chapter~\ref{ch:segmentation} of the thesis.

\section{Performance and computational cost metrics} \label{sec:ch2_metrics}

\paragraph{Performance metrics.} For \textbf{object detection}, the main performance metric is the Average Precision (\gls{AP}) metric from the COCO benchmark~\cite{lin2014microsoft}. The COCO AP metric is computed by averaging the per-class AP for the different classes, where the per-class AP is obtained by averaging the APs for 10 different Intersection over Union (\gls{IoU}) thresholds, ranging from $0.5$ to $0.95$ in steps of $0.05$. Here, the IoU is defined as

\begin{equation}
    \text{IoU} = \frac{\text{Area of Overlap}}{\text{Area of Union}}.
\end{equation}

The AP for a specific class and IoU threshold, is computed in the following way:

\begin{enumerate}
    \item All predictions for the class of interest are gathered and ranked according to their prediction score.
    \item For each prediction selected in descending order (\ie higher-ranked predictions are picked first), the (currently unmatched) target is identified resulting in the highest IoU value. If this IoU value surpasses the IoU threshold, the prediction is considered a \textit{true positive} (TP), otherwise it is labeled as \textit{false positive} (FP). In case of a true positive, the corresponding target is removed from the matching process, meaning that lower-ranked predictions can no longer match with this target. Targets that have not yet been matched are considered \textit{false negatives} (FN).
    \item The number of true positives (TP), false positives (FP), and false negatives (FN) is computed for each of the top-$k$ predictions, \ie for $k$ ranging from one to the total number of class predictions.
    \item The precision and recall values are computed for each $k$ using the TP, FP and FN values from previous step, where
    \begin{equation}
        \text{Precision} = \frac{\text{TP}}{\text{TP}+\text{FP}} \hspace{1cm} \text{and} \hspace{1cm} \text{Recall} = \frac{\text{TP}}{\text{TP}+\text{FN}}.
    \end{equation}
    \item The (recall, precision) points are plotted in a precision-recall graph for each $k$, from which the precision-recall curve is constructed by linearly connecting points between neighboring values of $k$.
    \item The precision-recall curve is modified to a monotonically decreasing function via
    \begin{equation}
        p_{\text{mod}}(r) = \max_{\hat{r}\geq r} \, p_{\text{orig}}(\hat{r}),
    \end{equation}
    with $p_{\text{mod}}(r)$ and $p_{\text{orig}}(r)$ the modified and original precision-recall curves respectively.
    \item The AP is obtained by computing the area under the modified precision-recall curve.
\end{enumerate}

The AP metric is reported as AP when computed on the validation set, and as AP$_\text{test}$ when computed on the test set.

Other metrics are also considered in addition to the main AP metric. Instead of averaging over $10$ different IoU thresholds, the AP$_{50}$ and AP$_{75}$ metrics only use the $0.50$ and $0.75$ IoU thresholds respectively. The tolerant AP$_{50}$ metric gives an indication of how many objects were found, whereas the stricter AP$_{75}$ metric indicates how many high-quality detections were made. Additionally, we also report the AP$_S$, AP$_M$ and AP$_L$ metrics, which are the same metrics as the main AP metric, but only computed on the \textit{small}, \textit{medium} and \textit{large} ground-truth objects respectively. A ground-truth object is said to be small when its area is smaller than $32^2$ pixels, medium when its area is between $32^2$ and $96^2$ pixels, and large when its area is larger than $96^2$ pixels. Here, the area is measured as the number of pixels in the ground-truth segmentation mask. On COCO, approximately $41\%$ of the objects are small, $34\%$ are medium, and $25\%$ are large.

For the \textbf{instance segmentation} task, we use the same metrics as for the object detection task. The AP computation occurs in exactly the same way as before, except that for instance segmentation the IoU values are computed based on the mask overlap instead of the box overlap. We additionally use following two metrics specifically for the instance segmentation task. 

Firstly, we have the AP$^*$ metric~\cite{kirillov2020pointrend}, which is exactly the same as the AP metric except that AP$^*$ is computed using LVIS~\cite{gupta2019lvis} target annotations and AP using COCO~\cite{lin2014microsoft} target annotations. As mentioned in Section~\ref{sec:ch2_tasks} and shown in Figure~\ref{fig:ch2_coco_lvis_mask}, the LVIS mask annotations are of significantly higher quality than the COCO mask annotations. Providing the AP$^*$ metric hence allows us to better evaluate the quality of the predicted segmentation masks.

\begin{figure}
    \centering
    \includegraphics[scale=0.14]{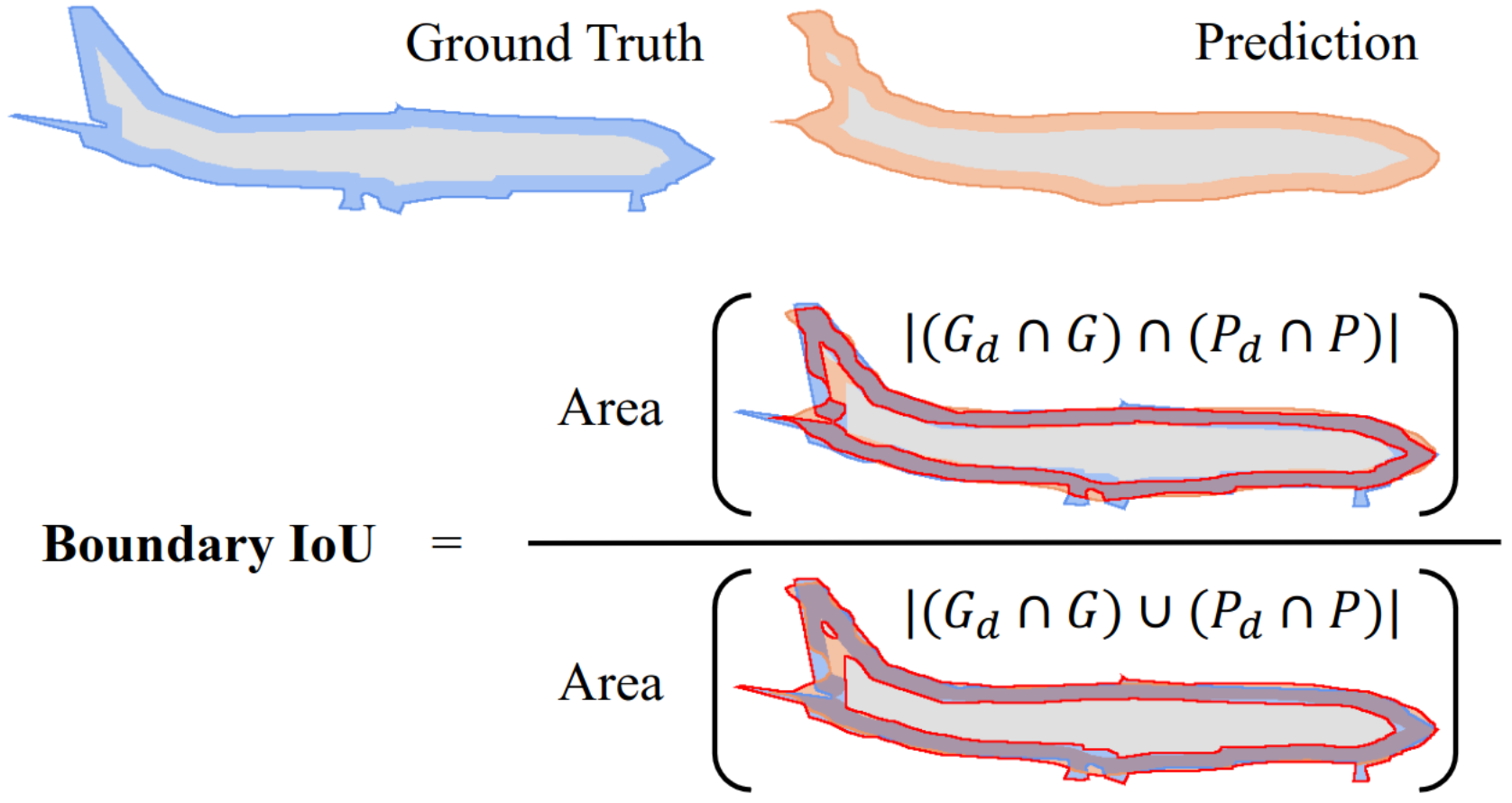}
    \caption[Background - Boundary IoU example image]{Example illustrating how the boundary IoU is computed (from \cite{cheng2021boundary}).}
    \label{fig:ch2_boundary_iou}
\end{figure}

Secondly, we have the AP$^{B*}$ metric~\cite{cheng2021boundary}, which is the same as the AP$^*$ metric except that AP$^{B*}$ uses IoU values computed with boundary IoU~\cite{cheng2021boundary} instead of with dense mask IoU. Boundary IoU computes IoU values only based on segmentation predictions made within the joint predicted and target mask boundary region (see Figure~\ref{fig:ch2_boundary_iou}), as opposed to dense mask IoU which computes IoU values based on all segmentation predictions made over the whole image. Here, the boundary region is constructed by dilating the contour line by $d$~pixels, with $d$ corresponding to $2$\% of the image diagonal. The AP$^{B*}$ metric hence allows us to further evaluate the fine-grained quality of segmentation masks by only measuring the segmentation quality near the mask boundaries.

\begin{figure}
    \centering
    \includegraphics[scale=0.6]{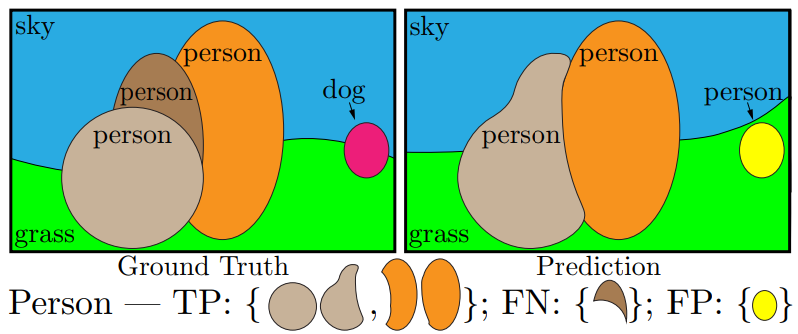}
    \caption[Background - PQ segment matching example image]{Toy example illustrating the PQ segment matching (from \cite{kirillov2019panoptic}).}
    \label{fig:ch2_segment_matching}
\end{figure}

For the \textbf{panoptic segmentation} task, we use the Panoptic Quality (\gls{PQ}) performance metric~\cite{kirillov2019panoptic}. The PQ metric is especially designed to treat the segmentations of both thing and stuff classes in a uniform way. The computation of the PQ metric can be split in two main steps:

\begin{enumerate}
    \vspace{-0.1cm}
    \item Segment matching: The first step consists in matching the predicted segments (\ie segmentation masks) with the ground-truth segments, where predicted and ground-truth segments match if their \gls{IoU} is greater than~$0.5$. Note that these segments are non-overlapping, as the panoptic segmentation task only allows for a single prediction per image pixel. Given the non-overlapping segments, it can easily be shown that this matching procedure results in unique pairings~\cite{kirillov2019panoptic}, \ie a single predicted segment can only be matched with a single ground-truth segment (and vice versa). Matched segments are categorized as true positives (TP), unmatched ground-truth segments as false negatives (FN), and unmatched predicted segments as false positives (FP). See Figure~\ref{fig:ch2_segment_matching} for a toy example illustrating the segment matching procedure taken from~\cite{kirillov2019panoptic}.
    
    \item PQ computation: The second step computes the PQ metric. The PQ metric is calculated for each class independently and averaged, such that the PQ metric is insensitive to class imbalance. Based on the (class-specific) TP, FN, and FP sets from step~1, the (class-specific) PQ metric is computed as

    \begin{equation}
        \text{PQ} = \frac{\sum_{(p,g) \in \text{TP}} \text{IoU}(p,g)}{|\text{TP}| + \frac{1}{2}|\text{FP}| + \frac{1}{2}|\text{FN}|}.
    \end{equation}
\end{enumerate}

The PQ metric is computed for all classes, but it can also be computed for the thing and stuff classes separately, giving the PQ$^{\text{Th}}$ and PQ$^{\text{St}}$ metrics respectively. Additionally, the PQ metric can be decomposed into the segmentation quality (SQ) and recognition quality (RQ) metrics as follows

\begin{equation}
    \text{PQ} = \underbrace{\frac{\sum_{(p,g) \in \text{TP}} \text{IoU}(p,g)}{|\text{TP}|}}_{\text{SQ}} \,\times\, \underbrace{\frac{|\text{TP}|}{|\text{TP}| + \frac{1}{2}|\text{FP}| + \frac{1}{2}|\text{FN}|}}_{\text{RQ}}.
\end{equation}

This simple decomposition provides useful insight by disentangling the segmentation mask performance SQ and the recognition performance RQ.

\paragraph{Computational cost metrics.} Next to the performance metrics, we also report various computational cost metrics of the different network designs. We distinguish following four types of computational cost metrics:

\begin{enumerate}
    \vspace{-0.1cm}
    \itemsep 0cm
    \item The number of network parameters (Params).
    \item The number of network floating point operations (FLOPs).
    \item The number of frames-per-second (FPS).
    \item The network peak GPU memory utilization (Mem).
\end{enumerate}

For the FLOPs, FPS and memory cost metrics, we additionally make the distinction between training and inference. In ambiguous cases, the training cost metrics are also denoted by tFLOPS, tFPS and tMem, and the inference cost metrics by iFLOPS, iFPS and iMem. Note that when computing the training cost metrics, we consider both the forward and backward passes, with the parameter update from the optimizer.

Moreover, the obtained FLOPs, FPS and memory cost metrics also depend on the input provided to the network. During the thesis, we consider two different settings. In Chapter~\ref{ch:neck} and in the first part of Chapter~\ref{ch:detection}, we compute these cost metrics by supplying the network with a batch of two $800\times800$ images, each containing ten target objects during training. In the second part of Chapter~\ref{ch:detection} and in Chapter~\ref{ch:segmentation}, we compute the cost metrics on the first $100$ images of the validation set and take the average, similar to~\cite{carion2020end}. Both approaches yield equivalent results when working with static networks, \ie networks performing the same computation regardless of the image content. However, when working with dynamic networks as in Chapter~\ref{ch:segmentation} where the cost metrics may vary from image to image, the second approach is to be preferred as the cost metrics are averaged over multiple images.

In order to count the number of FLOPs, we use the tool from Detectron2~\cite{wu2019detectron2}. Note that this tool in fact computes the number of multiply-accumulate operations (MACs), instead of the number of FLOPs. An addition followed by a multiplication is hence considered as one operation, and not two. Note that some operations are also neglected during the computation of the number of FLOPs, such as simple additions not followed by multiplications. See \cite{wu2019supported, wu2019ignored, picron2023extra} for an overview on the supported and ignored operations during FLOPs computation.

Note that considering different cost metrics is necessary in order to get a complete overview of the computational characteristics of a network. A network that has a lower cost \wrt another network for \textit{one} cost metric, does not automatically have a lower cost for \textit{all} cost metrics. It is for example possible that a network uses fewer FLOPs (lower cost), but also has lower FPS (higher cost) compared to another network.

\section{Standardized deep learning settings} \label{sec:ch2_settings}

\paragraph{Backbone settings.} When using a \textbf{ResNet}~\cite{he2016deep} backbone, we initialize the backbone network with ImageNet1k~\cite{russakovsky2015imagenet} pretrained weights provided by TorchVision~\cite{torchvision2016} (version~1) and freeze the stem, stage~1 and BatchNorm~\cite{ioffe2015batch} layers (see Section~\ref{sec:ch2_meta_arch} for the used terminology).

For experiments with the \textbf{Swin-L}~\cite{liu2021swin} backbone, we use the weights from the official GitHub repository~\cite{liu2021github} that were obtained after pretraining on ImageNet22k~\cite{deng2009imagenet} with a $384\times384$ input image resolution.

\paragraph{Neck settings.} Throughout our experiments, we either use a $P_3$-$P_7$ feature pyramid or a $P_2$-$P_7$ feature pyramid, containing five and six feature maps respectively, each having feature size $256$.

The initial $P_3$-$P_7$ feature pyramid is constructed based on the backbone output feature maps $C_3$ to $C_5$ from stages $2$, $3$ and $4$ respectively. Remember that the subscript denotes how many times the feature map was downsampled with factor $2$ compared to the input image. The initial $P_3$ to $P_5$ maps are obtained by applying simple linear projections on $C_3$ to $C_5$, whereas the initial $P_6$ and $P_7$ maps are obtained by applying a simple network on $C_5$ (or $P_5$) consisting of $2$ convolutions with stride $2$, and a ReLU activation in between (similar to \cite{lin2017focal}).

The initial $P_2$-$P_7$ feature pyramid is constructed in a similar way, where additionally a linear projection is applied on the backbone feature map $C_2$ from stage~$1$ to obtain the initial $P_2$ map.

\paragraph{Data settings.} During \textbf{training}, we use the multi-scale data augmentation scheme from DETR~\cite{carion2020end}. The scheme consists of two main steps. Firstly, the scheme starts by flipping the input image horizontally with a $50\%$ chance. Secondly, the scheme proceeds either with a \textit{simple} resize operation, or with a \textit{cropped} resize operation, each chosen with equal probability. The simple resize operation resizes the shortest image edge between $480$ and $800$, while limiting the largest image edge to $1333$ and leaving the image aspect ratio unchanged. The cropped resize operation first resizes the shortest image edge between $400$ and $600$, then crops a region of width and height between $384$ and $600$, and finally resizes the cropped image using the simple resize operation described previously.

During \textbf{inference}, the input images are resized to have a shortest image edge of $800$, while limiting the largest image edge to $1333$ and leaving the image aspect ratio unchanged.

\paragraph{Training settings.} Unless specified otherwise, we train the networks using the AdamW optimizer~\cite{loshchilov2017decoupled} with weight decay~$10^{-4}$. We use an initial learning rate of $10^{-5}$ for the backbone parameters and for the linear projection modules computing the multi-scale deformable attention (\gls{MSDA}) sampling offsets~\cite{zhu2020deformable}. For the remaining model parameters, we use an initial learning rate of $10^{-4}$. The models are trained and evaluated on two GPUs, each having a batch size of one or two depending on the experiment.

\cleardoublepage
  \graphicspath{{\chaptersdir/neck/image/}}%
  \chapter{Trident Pyramid Networks for Object Detection} \label{ch:neck}

\definecolor{shadecolor}{gray}{0.85}
\begin{shaded}
    The work in this chapter was adapted from the following conference proceeding: \vspace{-0.1cm} \\
    \bibentry{picron2022trident}
\end{shaded}

\section{Introduction}
Many computer vision tasks such as object detection and instance segmentation require strong features both at low and high resolution to detect both large and small objects respectively. This is in contrast to the image classification task where low resolution features are sufficient as usually only a single object is present in the center of the image. Networks developed specifically for the image classification task (\eg \cite{simonyan2014very, he2016deep, xie2017aggregated}), denoted by \textit{backbones}, are therefore insufficient for multi-scale vision tasks. Especially poor performance is to be expected on small objects, as shown in \cite{lin2017feature}.

In order to alleviate this problem, sometimes named the \textit{feature fusion problem}, top-down mechanisms are added \cite{lin2017feature} to propagate semantically strong information from the low resolution to the high resolution feature maps, with improved performance on small objects as a result. Additionally, bottom-up mechanisms can also be appended \cite{liu2018path} such that the lower resolution maps can benefit from the freshly updated higher resolution maps. These top-down and bottom-up mechanisms can now be grouped into a layer, after which multiple of these layers can be concatenated, as done in \cite{tan2020efficientdet}. This part of a computer vision network is called the \textit{neck}, laying in between the \textit{backbone} and the task-specific \textit{head}. See Section~\ref{sec:ch2_meta_arch} and Figure~\ref{fig:ch2_backbone_neck_head} in particular for more information about the backbone-neck-head meta architecture.

The top-down and bottom-up operations can be regarded as \textit{communication}-based processing operating on two feature maps, as opposed to \textit{self}-processing operating on a single feature map. Existing necks such as FPN \cite{lin2017feature}, PANet \cite{liu2018path} and BiFPN \cite{tan2020efficientdet} mostly focus on communication-based processing, as this nicely supplements the backbone merely consisting of self-processing. However, when having multiple communication-based operations in a row, communication tends to saturate (everyone is up to date) and hence becomes superfluous. We argue it is therefore more effective to alternate communication-based processing with sufficient self-processing, allowing the feature maps to come up with new findings to be communicated.

\paragraph{First contribution.} Based on this observation, we design the Trident Pyramid Network (\gls{TPN}) neck consisting of sequential top-down and bottom-up operations alternated with parallel self-processing mechanisms. The TPN neck is equipped with hyperparameters controlling the amount of communication-based processing and self-processing. These hyperparameters enable our TPN neck to find a better balance between both types of processing compared to other necks as BiFPN \cite{tan2020efficientdet}, outperforming the latter by $0.5$ AP when using the ResNet-50 backbone.

\paragraph{Second contribution.} When having additional compute to improve performance, practitioners typically decide to replace their backbone with a heavier one. A ResNet-50+FPN network for example gets traded with the heavier ResNet-101+FPN network. Yet, one might wonder whether it is not more beneficial to add additional computation into the neck (\ie at the feature pyramid level) by using a ResNet-50+TPN network, rather than into the backbone by using a ResNet-101+FPN network. When comparing both options under similar computational characteristics, we show a $1.7$ AP improvement of the ResNet-50+TPN network over the ResNet-101+FPN network. This empirically shows that it is more beneficial to add additional computation into the neck, highlighting the importance of performing computation at the feature pyramid level in modern-day object detection systems.

\paragraph{Code.} The code is available at \url{https://github.com/CedricPicron/TPN}.

\section{Related work}
In order to obtain multi-scale features, early detectors performed predictions on feature maps directly coming from the backbone, such as MS-CNN \cite{cai2016unified} and SSD \cite{liu2016ssd}. As the higher resolution maps from the backbone contain relatively weak semantic information, top-down mechanisms were added to propagate semantically strong information from lower resolution maps back to the higher resolution maps as in FPN \cite{lin2017feature} and TDM \cite{shrivastava2016beyond}. Since, many variants and additions have been proposed: PANet \cite{liu2018path} appends bottom-up connections, M2det \cite{zhao2019m2det} uses a U-shape feature interaction architecture, ZigZagNet \cite{lin2019zigzagnet} adds additional pathways between different levels of the top-down and bottom-up hierarchies, NAS-FPN \cite{ghiasi2019fpn} and Hit-Detector \cite{guo2020hit} use Neural Architecture Search (NAS) to automatically design a feature interaction topology, and BiFPN \cite{tan2020efficientdet} modifies PANet by removing some connections, adding skip connections and using weighted feature map aggregation. All of the above variants focus on improving the communication between the different feature maps. We argue however that to be effective, extra self-processing is needed in between the communication flow.

However, not all methods use a feature pyramid to deal with scale variation. TridentNet \cite{li2019scale} applies parallel branches of convolutional blocks with different dilations on a single feature map to obtain scale-aware features. In DetectoRS \cite{qiao2021detectors}, they combine this idea with feature pyramids, by applying their switchable atrous convolutions (SAC) inside their recursive feature pyramids (RFP). Note that to avoid any name confusion with TridentNet, we call our neck by its abbreviated name TPN as opposed to Trident Pyramid Network.

Our TPN neck is also related to networks typically used for segmentation such as U-Net \cite{ronneberger2015u} and stacked hourglass networks \cite{newell2016stacked}, as these networks also use a combination of top-down, self-processing and bottom-up operations. However, a major difference between these networks and our TPN neck, is that they do not operate on a feature pyramid in the sense that lower resolution maps are only generated and used within a single layer (\eg within a single hourglass) and are not shared across layers (\eg across two neighboring hourglasses).

\section{Method}
\subsection{TPN neck architecture}
Recall from Section~\ref{sec:ch2_meta_arch} that a neck outputs an updated feature pyramid from its input feature pyramid, and that a feature pyramid is defined as a collection of feature maps, with feature maps defined as a collection of feature vectors (called features) organized in a two-dimensional map. More specifically, feature map~$P_l$ denotes a feature map of level $l$, which is $2^l$ times smaller in width and height compared to the initial image resolution. A popular choice for the feature pyramid \cite{lin2017focal} is to consider feature maps $\{P_3, P_4, P_5, P_6, P_7\}$, which is used as the default setting for the various TPN experiments found in this chapter.

\begin{figure}
    \centering
    \includegraphics[scale=0.6]{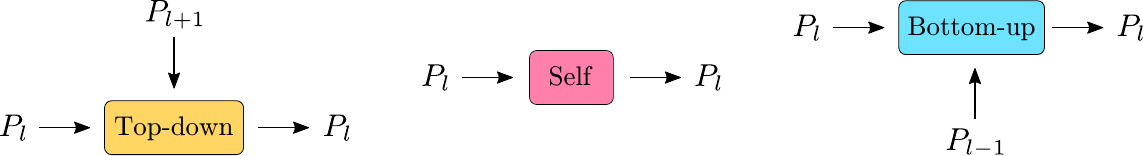}
    \caption[TPN design - Building blocks]{Collection of building blocks for neck architecture design. Here $P_l$ denotes feature map of level~$l$, which is $2^l$ times smaller compared to the initial image resolution. (\textit{Left}) General top-down operation updating feature map $P_l$ with information from lower resolution map $P_{l+1}$. (\textit{Middle}) General self-processing operation updating feature map $P_l$ with information from itself, \ie from feature map $P_l$. (\textit{Right}) General bottom-up operation updating feature map $P_l$ with information from higher resolution map $P_{l-1}$.}
    \label{fig:ch3_operations}
\end{figure}

The neck is constructed from three building blocks: top-down operations, self-processing operations and bottom-up operations (see Figure~\ref{fig:ch3_operations}). In this subsection, we focus on how these operations are best combined, independently of their precise implementations. We call this configuration of operations making up a neck, the \textit{neck architecture}. The specific implementations corresponding to the top-down, self-processing and bottom-up operations are discussed in Subsection~\ref{sec:ch3_self} and Subsection~\ref{sec:ch3_td_bu}.

\begin{figure}
    \centering
    \includegraphics[scale=0.6]{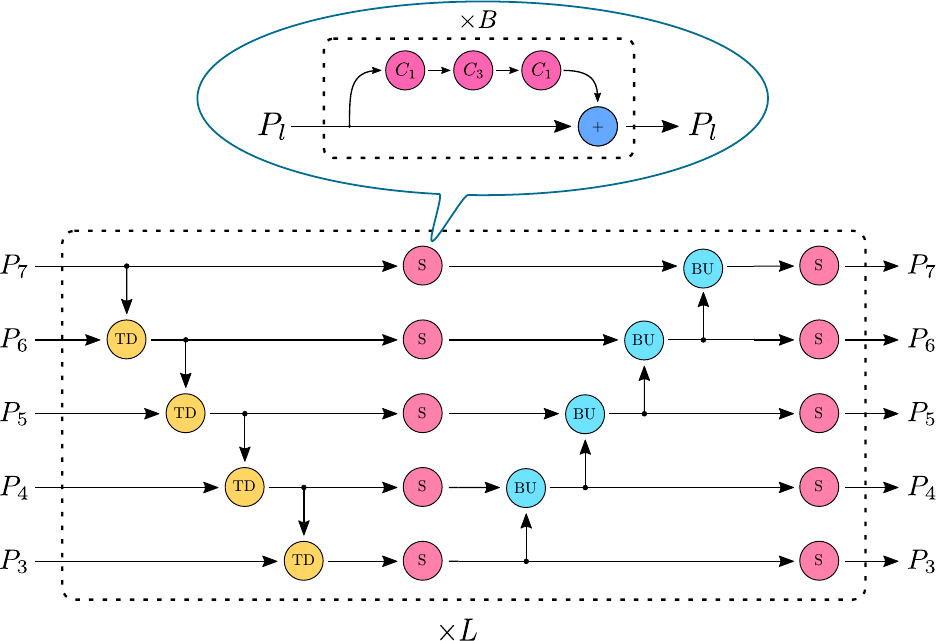}
    \caption[TPN design - TPN architecture]{Our TPN neck architecture consisting of $L$ consecutive TPN neck layers (\textit{bottom}), with each self-processing operation consisting of $B$ consecutive bottleneck layers (\textit{top}). The name `Trident Pyramid Network' is inspired by the top-down, first self-processing and bottom-up operations resembling a trident.}
    \label{fig:ch3_tpn_detailed}
\end{figure}

The TPN neck architecture is displayed in the lower part of Figure~\ref{fig:ch3_tpn_detailed} (\ie without text balloon). The TPN neck consists of $L$ consecutive TPN layers, with each layer consisting of sequential top-down and bottom-up operations, alternated by parallel self-processing operations. The TPN neck architecture in itself does not differ much from those of existing necks such as PANet \cite{liu2018path} or BiFPN \cite{tan2020efficientdet}. However, our TPN neck architecture explicitly incorporates self-processing operations which could be of any kind (\eg convolutional or attention-based) and depth (\ie one layer or more). This is in contrast with existing necks as PANet and BiFPN, that use a single convolution layer after each top-down and bottom-up operation. By doing so, these existing necks are unable to balance the amount of communication-based processing and self-processing, a balance which turns out to be sub-optimal in their case (see Subsection~\ref{sec:ch3_balance}). The TPN neck removes this limitation by considering abstract self-processing operations, which could be of any depth, allowing the optimal balance to be found between communication-based processing and self-processing.

\subsection{Self-processing operation} \label{sec:ch3_self}
\begin{figure}
    \centering
    \includegraphics[scale=0.35]{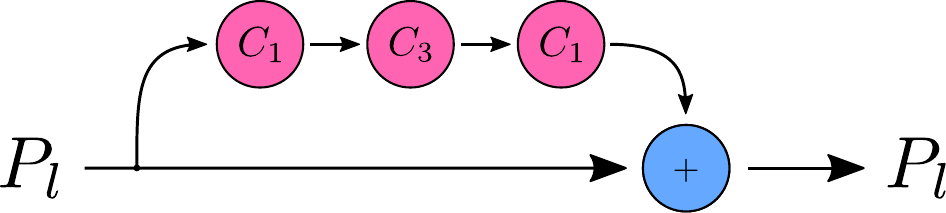}
    \caption[TPN design - Bottleneck layer]{Bottleneck layer used as base self-processing layer. It is a skip-connection operation with a residual branch consisting of three convolution operations: a convolution operation of kernel size~$1$ reducing the original feature size to the hidden feature size, a content convolution operation of kernel size~$3$ applied on the hidden feature size, and finally a convolution operation of kernel size~$1$ expanding the hidden feature size back to the original feature size. Note that each convolution operation (\ie pink convolution node) consists of the actual convolution preceded \cite{he2016identity} by group normalization \cite{wu2018group} and a ReLU activation function.}
    \label{fig:ch3_self_bottleneck}
\end{figure}

In general, we consider the self-processing operation to consist of a sequence of $B$ base self-processing layers, where the base self-processing layer could be any convolution or attention-based operation. In practical implementations, we use the bottleneck layer from \cite{he2016deep} as the base self-processing layer (see Figure~\ref{fig:ch3_self_bottleneck}).

The TPN neck hence consists of $L$ TPN layers, with each TPN layer consisting of top-down and bottom-up operations alternated by $B$ base self-processing layers (see also Figure~\ref{fig:ch3_tpn_detailed}). The hyperparameter pair $(L, B)$ balances the amount of communication-based processing and self-processing, with $L$ determining the amount of communication-based processing and $B$ tuning the amount of self-processing. In what follows, we also refer to the hyperparameter pair $(L, B)$ as the TPN configuration. In Subsection~\ref{sec:ch3_balance}, we empirically find out which TPN configurations work best for various computation budgets.

\subsection{Top-down and bottom-up operations} \label{sec:ch3_td_bu}
\begin{figure}
    \centering
    \includegraphics[scale=0.3]{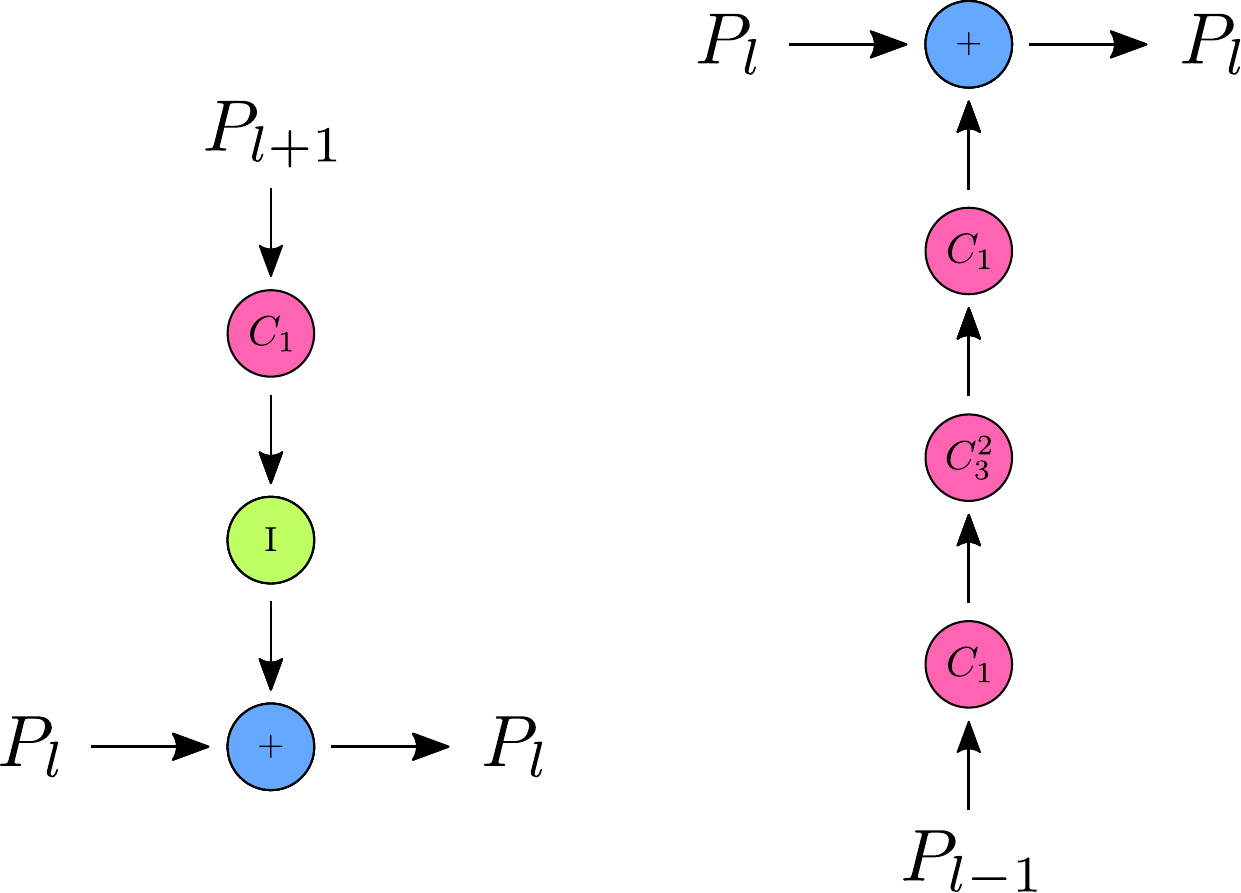}
    \caption[TPN design - Top-down and bottom-up layers]{Implementation of the top-down (\textit{left}) and bottom-up (\textit{right}) operations. The pink convolution nodes are defined as in Figure~\ref{fig:ch3_self_bottleneck}, with the subscript denoting the kernel size and the superscript denoting the stride (stride~$1$ when omitted). The green node is an interpolation node resizing the input feature map to the required resolution by using bilinear interpolation.}
    \label{fig:ch3_def_comp}
\end{figure}

Let us now take a closer look at the top-down and bottom-up operations. Generally speaking, these operations update a feature map based on a second feature map, either having a lower resolution (top-down case) or a higher resolution (bottom-up case). Our implementations of the top-down and bottom-up operations are shown in Figure~\ref{fig:ch3_def_comp}. The operations consist of adding a modified version of $P_{l\pm1}$ to $P_l$. This is similar to traditional skip-connection operations, with the exception that the \textit{residual features} originate from a different feature map. The residual branch of the top-down operation consists of a linear projection followed by bilinear interpolation. The presence of the linear projection is important here, as it makes the expectation of the residual features zero at initialization. Failing to do so can be detrimental, especially when building deeper neck modules, as correlated features add up without constraints. An alternative consists in replacing the blue addition nodes with averaging nodes. However, this fails to keep the skip connection computation free (due to the $0.5$ factor), which is undesired \cite{he2016identity}. The residual branch of the bottom-up operation is similar to the bottleneck residual branch in Figure~\ref{fig:ch3_self_bottleneck}. Only the middle $3 \times 3$ convolution has stride~$2$ instead of stride~$1$, avoiding the need for an interpolation step later in the residual branch.

\section{Experiments} \label{sec:ch3_experiments}
\subsection{Setup} \label{sec:ch3_setup}
\paragraph{Dataset.} 
We perform object detection experiments on the COCO detection dataset~\cite{lin2014microsoft}, where we train on the 2017 COCO training set and evaluate on the 2017 COCO validation set. See Section~\ref{sec:ch2_tasks} for more information on the object detection task and the COCO dataset.

\paragraph{Implementation details.}
We follow the standardized settings stipulated in Section~\ref{sec:ch2_settings} for the backbone settings, neck settings, data settings, and (some of the) training settings. In what follows, we provide the TPN settings, the detection head settings, and the remaining training settings.

For our \textbf{TPN} necks, we use group normalization~\cite{wu2018group} with $8$ groups, and we use a hidden feature size of $64$ inside the bottleneck layers (see Figure~\ref{fig:ch3_self_bottleneck}).

For the object detection \textbf{head}, we use the one-stage detector head from RetinaNet~\cite{lin2017focal}, with $C=1$ or $C=4$ hidden layers in both classification and bounding box subnets. We follow the implementation and settings from Detectron2~\cite{wu2019detectron2}, except that the last layer of the subnets has kernel size $1$ (instead of $3$) and that we normalize the losses per feature map (instead of over the whole feature pyramid).

During \textbf{training}, we use the 3x schedule for our main experiments in Subsection~\ref{sec:ch3_main_experiments}, consisting of $36$ training epochs with learning rate drops after the $27$th and $33$rd epoch with a factor $0.1$. For the experiments in Subsection~\ref{sec:ch3_balance}, we use the 1x schedule instead, consisting of $12$ training epochs with learning rate drops after the $9$th and $11$th epoch with a factor $0.1$. The models are trained on $2$ GPUs, each having a batch size of $2$.

\subsection{Main experiments} \label{sec:ch3_main_experiments}
\paragraph{Baselines.}
In this subsection, we perform experiments to evaluate the TPN neck. As baseline, we consider the BiFPN neck from \cite{tan2020efficientdet} with batch normalization layers \cite{ioffe2015batch} replaced by group normalization layers \cite{wu2018group}, and with Swish-1 activation functions \cite{ramachandran2017searching} replaced by ReLU activation functions. Multiple of these BiFPN layers are concatenated such that the BiFPN neck shares similar computational characteristics compared to the tested TPN necks.

\begin{figure}
    \centering
    \includegraphics[scale=0.6]{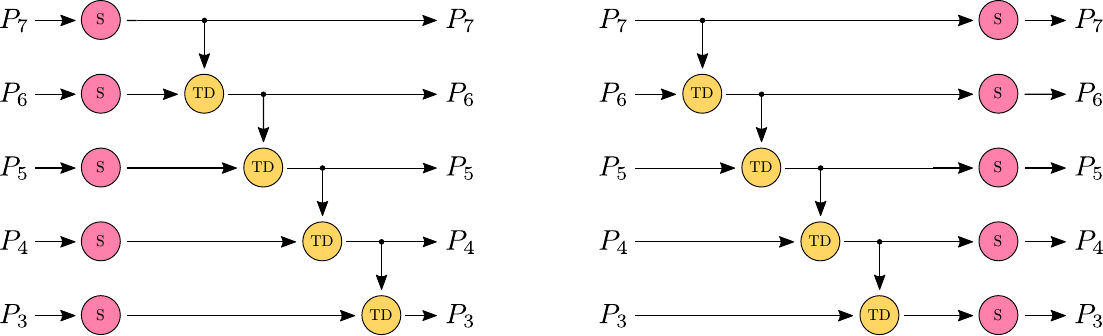}
    \caption[TPN - bFPN and hFPN baselines]{(\textit{Left}) The baseline bFPN neck architecture simulating a heavier backbone followed by a single FPN layer. (\textit{Right}) The baseline hFPN neck architecture simulating a single FPN layer followed by a heavier head.}
    \label{fig:ch3_arch_bfpn_hfpn}
\end{figure}

Additionally, we would like to compare our TPN neck with the popular FPN neck under similar computation budgets. As the FPN layer was not designed to be concatenated many times, we instead provide the bFPN and hFPN baselines (see Figure~\ref{fig:ch3_arch_bfpn_hfpn}). Here, the bFPN baseline performs additional self-processing before the FPN simulating a heavier backbone, whereas the hFPN baseline performs additional self-processing after the FPN simulating a heavier head. As such, we are not only be able to evaluate whether the TPN neck outperforms other necks, but also whether it outperforms detection networks using a simple FPN neck with heavier backbones or heads, while operating under similar computation budgets. 

Finally, we compare ResNet-101+FPN and ResNet-101+TPN networks with a ResNet-50+TPN network of similar computation budget, to further assess whether it is more beneficial to put additional computation into the backbone or into the neck.

\paragraph{Metrics.} As performance metrics, we use the COCO AP metrics, and as computational cost metrics, we use the Params, tFPS, tMem and iFPS metrics. The computational cost metrics are obtained using a GeForce GTX 1660 Ti GPU, by applying the network on a batch of two $800\times800$ images, each containing ten ground-truth objects during training. See Subsection~\ref{sec:ch2_metrics} for more information about these metrics.

\paragraph{Results.}
The experiment results evaluating four different TPN configurations and the five baselines, are found in Table~\ref{tab:ch3_tpn}. We make following four sets of observations.

\begin{table*}
    \centering
    \caption[TPN - Experiment results]{Experiment results on the 2017 COCO validation set of different TPN necks (top four rows) and its baselines (bottom five rows). The five leftmost columns specify the network (Back = Backbone, Neck, $L$ = Number of layers, $B$ = Number of bottleneck layers per self-processing operation, $C$~= Number of hidden layers in classification and bounding box subnets), the middle six columns the performance metrics, and the four rightmost columns the computational cost metrics. \vspace{0.2cm}}
    \resizebox{\columnwidth}{!}{
        \begin{tabular}{c c c c c|c c c c c c|c c c c}
            \toprule
            Back & Neck & $L$ & $B$ & $C$ & AP & AP$_{50}$ & AP$_{75}$ & AP$_S$ & AP$_M$ & AP$_L$ & Params & tFPS & tMem & iFPS \\ \midrule
            R50 & TPN & $1$ & $7$ & $1$ & $41.3$ & $60.5$ & $44.2$ & $26.3$ & $45.9$ & $52.5$ & $36.3$ M & $1.7$ & $3.31$ GB & $5.3$ \\
            R50 & TPN & $2$ & $3$ & $1$ & $41.6$ & $60.9$ & $44.6$ & $26.4$ & $45.8$ & $53.2$ & $36.2$ M & $1.7$ & $3.21$ GB & $5.5$ \\
            R50 & TPN & $3$ & $2$ & $1$ & $\mathbf{41.8}$ & $61.1$ & $44.4$ & $26.2$ & $46.1$ & $53.7$ & $36.7$ M & $1.6$ & $3.27$ GB & $5.3$ \\
            R50 & TPN & $5$ & $1$ & $1$ & $\mathbf{41.8}$ & $\mathbf{61.2}$ & $\mathbf{45.0}$ & $26.0$ & $\mathbf{46.3}$ & $53.4$ & $37.1$ M & $1.6$ & $3.22$ GB & $5.3$ \\
            \midrule
            R50 & BiFPN & $7$ & $-$ & $1$ & $41.3$ & $\mathbf{61.2}$ & $43.7$ & $\mathbf{27.1}$ & $45.2$ & $\mathbf{53.8}$ & $34.7$ M & $1.8$ & $3.22$ GB & $6.0$ \\
            \midrule
            R50 & bFPN & $1$ & $14$ & $1$ & $39.6$ & $60.3$ & $42.4$ & $24.2$ & $43.5$ & $51.3$ & $36.1$ M & $1.7$ & $3.26$ GB & $5.4$ \\
            R50 & hFPN & $1$ & $14$ & $1$ & $40.0$ & $60.2$ & $43.0$ & $25.6$ & $43.9$ & $51.1$ & $36.1$ M & $1.7$ & $3.26$ GB & $5.4$ \\
            \midrule
            R101 & FPN & $1$ & $-$ & $4$ & $40.1$ & $60.1$ & $42.8$ & $24.0$ & $44.0$ & $52.7$ & $55.1$ M & $1.4$ & $3.20$ GB & $4.1$ \\
            R101 & TPN & $1$ & $2$ & $1$ & $40.9$ & $61.0$ & $44.2$ & $25.0$ & $45.3$ & $52.6$ & $51.7$ M & $1.6$ & $3.20$ GB & $4.6$ \\
            \bottomrule
        \end{tabular}}
    \label{tab:ch3_tpn}
\end{table*}

Firstly, notice how the $L$ and $B$ hyperparameters (defining the TPN configuration) are chosen in order to obtain models with similar computational costs. Here, hyperparameter~$L$ denotes the number of consecutive neck layers, while hyperparameter~$B$ denotes the number of bottleneck layers per self-processing operation (see also Figure~\ref{fig:ch3_tpn_detailed}). These similar computational cost metrics ensure us that a fair comparison is made between the different TPN networks.

Secondly, we observe the similar performance between the four different TPN configurations, all four obtaining between $41.3$ and $41.8$ AP. At first glance, it appears that having more TPN neck layers $L$ is slightly more beneficial than having more bottleneck layers $B$ under similar computational budgets. In Subsection~\ref{sec:ch3_balance}, we further investigate whether this is indeed the case by analyzing which TPN configurations yield the best accuracy vs. efficiency trade-off at various computation budgets.

Thirdly, when comparing our TPN necks (top four rows) with the BiFPN neck (fifth row), we observe that one TPN configuration performs on par with the BiFPN neck, whereas the remaining three TPN configurations outperform the BiFPN neck by up to $0.5$ AP. We especially notice improvements on the AP$_{75}$ metric, where all four TPN configurations outperform the BiFPN neck by $0.5$ up to $1.3$ AP$_{75}$. This hence shows that our TPN necks provide more accurate detections than the BiFPN neck.

Fourthly, when comparing the ResNet-50+TPN networks (top four rows) with the ResNet-50+bFPN, ResNet-50+hFPN, ResNet-101+FPN and ResNet-101+TPN baselines (bottom four rows), we again see that the ResNet-50+TPN networks work best. The best-performing baseline from this category (\ie the ResNet-101+TPN network), is outperformed by all four ResNet-50+TPN configurations with $0.4$ up to $0.9$ AP. Moreover, note that the ResNet-101+TPN baseline has considerably more parameters and is clearly slower at inference, and still does not match the performance of the ResNet-50+TPN networks despite the higher computational cost. This hence shows that the TPN neck not only outperforms other necks such as BiFPN, but also other detection networks using heavier backbones or heads while operating under similar overall computation budgets. This highlights the importance and effectiveness of neck networks within general object detection networks, by showing that it is more beneficial to put additional computation into the neck rather than into the backbone or head.

\subsection{Comparison between different TPN configurations} \label{sec:ch3_balance}

\begin{figure}
    \centering
    \includegraphics[scale=0.4]{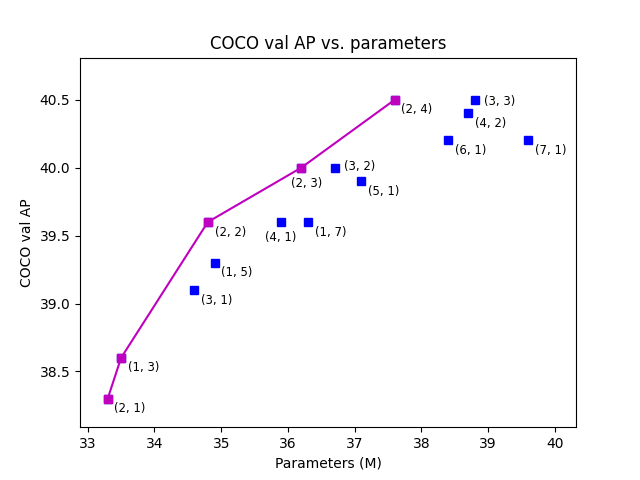} \\
    \vspace{0.15cm}
    \includegraphics[scale=0.4]{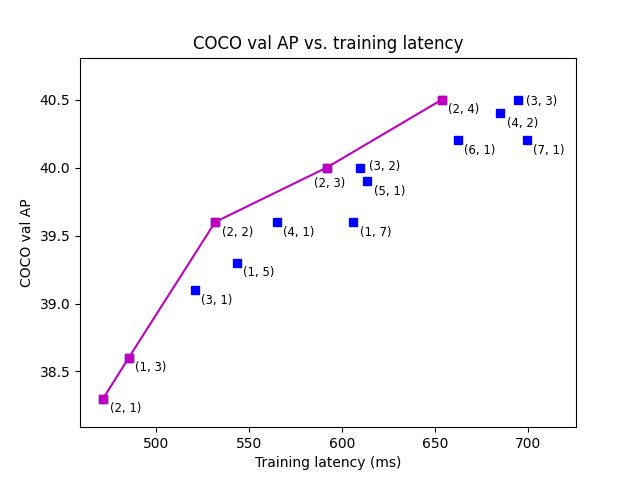} \\
    \vspace{0.15cm}
    \includegraphics[scale=0.4]{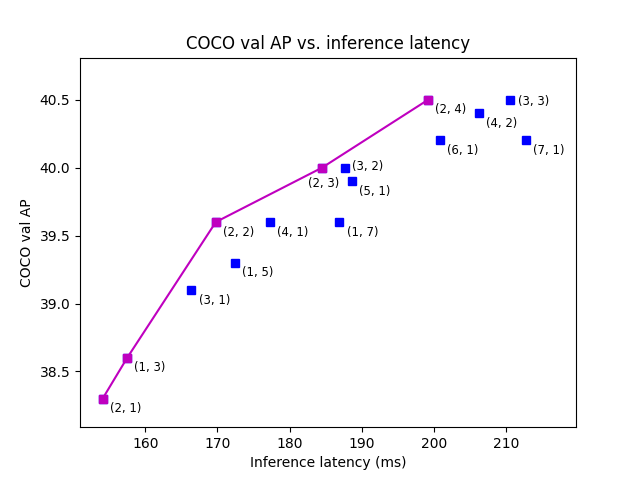}
    \caption[TPN - Comparison of different TPN configurations]{Accuracy vs. efficiency comparisons between $15$ different $(L, B)$ TPN configurations using the `parameters' (\textit{top}), `training latency' (\textit{middle}) and `inference latency' (\textit{bottom}) metrics. The accuracies correspond to the COCO validation APs, obtained after training the models for $12$ epochs using the 1x schedule. The TPN configurations yielding the best accuracy vs. efficiency trade-off at various computation budgets, are highlighted in magenta.}
    \label{fig:ch3_balance}
\end{figure}

In this subsection, we investigate which TPN configurations yield the best accuracy vs. efficiency trade-off at various computation budgets. Here, a TPN configuration is determined by the hyperparameter pair $(L, B)$, respectively denoting the number of TPN layers and the number of bottleneck layers per self-processing operation. In Figure~\ref{fig:ch3_balance}, we compare $15$ different $(L, B)$ TPN configurations using the `parameters', `training latency' and `inference latency' cost metrics, with the latency metrics corresponding to the time it takes to process a batch of two $800\times800$ images (\ie 1/FPS).

We can see from the magenta curves yielding the optimal TPN configurations at various computation budgets, that having a good balance between communication-based processing (in the form of TPN layers $L$) and self-processing (in the form of bottleneck layers per self-processing operation $B$) is crucial. We can for example see that the balanced $(2, 2)$ configuration outperforms the unbalanced $(3, 1)$ and $(1, 5)$ configurations. The same observation can also be made at higher computation budgets, where the balanced $(2, 4)$, $(4, 2)$ and $(3, 3)$ configurations outperform the unbalanced $(6, 1)$ and $(7, 1)$ configurations.

We hence empirically show that balanced $(L, B)$ configurations with $L \geq 2$ and $B \geq 2$ are to be preferred over unbalanced configurations such as $(L, 1)$ and $(1, B)$. By having only one self-processing operation in between each communication-based operation, existing necks with a $(L, 1)$ configuration such as PANet \cite{liu2018path} and BiFPN \cite{tan2020efficientdet} crucially lack self-processing. The TPN neck solves this problem by introducing the $(L, B)$ hyperparameter pair, such that a better balance between communication-based processing and self-processing can be chosen.

\section{Future developments} \label{sec:ch3_developments}
In this section, we discuss some future developments in the literature related to the neck design and to the balance between the size of the backbone, neck and head in object detection networks.

\paragraph{Neck design.} When designing the TPN neck, we considered three distinct building blocks, namely the top-down, self-processing and bottom-up operation, similar to existing works at the time. However, it is also possible to combine these three operations into an all-in-one \textbf{multi-scale operation}, as done in recent works~\cite{zhu2020deformable, tan2021giraffedet}. These multi-scale operations can be subdivided into two categories, namely \textit{local} and \textit{global} multi-scale operations. Local multi-scale operations update each feature map $P_l$ only using the neighboring feature maps $P_{l-1}$, $P_l$ and $P_{l+1}$, whereas global multi-scale operations use all feature maps from the feature pyramid. A popular local multi-scale operation is the Queen-fusion operation used in GFPN~\cite{tan2021giraffedet}, and a popular global multi-scale operation is the multi-scale deformable attention (\gls{MSDA}) operation from DefEncoder~\cite{zhu2020deformable}.

\begin{figure}
    \centering
    \includegraphics[scale=1.0]{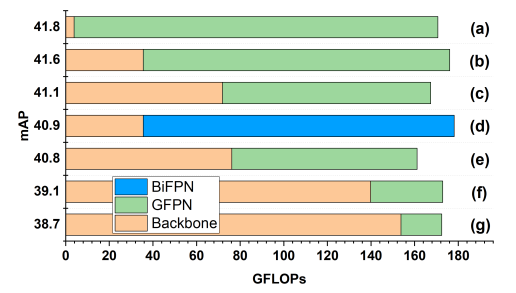}
    \caption[GiraffeDet - Comparison of backbone-neck combinations]{Performance for networks with different backbone and neck sizes under similar computation budgets (taken from \cite{tan2021giraffedet}). The best performance is obtained when using a small backbone in combination with a large neck. See~\cite{tan2021giraffedet} for more information about the different networks.}
    \label{fig:ch3_giraffedet}
\end{figure}

\paragraph{Backbone-neck-head balance.} In this chapter, we show using the TPN neck that it is more beneficial to put additional computation into the neck rather than into the backbone or head. Recent works~\cite{tan2021giraffedet, xu2022damo, lyu2022rtmdet} reach similar conclusions, while using different types of backbone, neck and head networks. This hence shows that this phenomenon is general and not specific to the used networks. In~\cite{tan2021giraffedet}, it is moreover shown that this principle of using large necks  rather than large backbones, can be pushed to the extreme, by obtaining the best accuracy vs. efficiency trade-off when using a very small backbone in combination with a large neck (see Figure~\ref{fig:ch3_giraffedet}). The same is shown for heads in~\cite{xu2022damo}, where the best accuracy vs. efficiency trade-off is obtained when using a large neck in combination with a small head consisting of only a single linear layer for both classification and bounding box regression networks.

\section{Future work}
In this section, we discuss some potential directions for future work. We start off by discussing some aspects closely related to the conducted research, and continue with more general research questions which naturally arise from our results and from the recent findings in the literature.

\paragraph{More extensive experiments.} More extensive experiments could be performed in the future. Currently, the experiments are limited by only providing results for one type of self-processing operation (\ie the bottleneck layer from~\cite{he2016deep}), for one type of backbone (\ie ResNet-like backbones~\cite{he2016deep}), for one type of head (\ie RetinaNet~\cite{lin2017focal}), and on a single dataset (\ie COCO~\cite{lin2014microsoft}).

\paragraph{Comparison with recent necks.} In this work, we compare our TPN neck with the BiFPN neck, which was the state-of-the-art neck design at the time this research was conducted. However, many new necks have been developed since, such as the DefEncoder~\cite{zhu2020deformable} and the GFPN~\cite{tan2021giraffedet} necks. It would hence be useful to compare the TPN neck with these newer necks, and especially with the DefEncoder neck which has become the go-to neck design for many popular object detection~\cite{zhang2022dino} and instance segmentation~\cite{cheng2022masked} models.

\paragraph{TPN-like multi-scale operation.} As pointed out in Section~\ref{sec:ch3_developments}, some recent works~\cite{zhu2020deformable, tan2021giraffedet} use an all-in-one multi-scale operation, as opposed to the distinct top-down, self-processing and bottom-up operations used in TPN. However, as opposed to the TPN design, the multi-scale operations introduced by these works do not contain hyperparameters which enable the balancing between communication-based processing and self-processing. A future research direction hence consists in designing a TPN-like multi-scale operation, where hyperparameters are introduced that would allow the balancing of communication-based processing and self-processing.

\paragraph{Computation balance for multi-scale vision.} It has been empirically shown with our TPN experiments and with experiments from~\cite{tan2021giraffedet, xu2022damo, lyu2022rtmdet}, that the best accuracy vs. efficiency trade-offs are obtained when using a small backbone, a large neck, and a small head. However, so far, these findings have only been made for the object detection task, and only on the COCO dataset~\cite{lin2014microsoft}. A future research direction would be to investigate whether this also holds on other datasets and for other multi-scale computer vision tasks.

If it indeed turns out that this paradigm results in the best accuracy vs. efficiency trade-off for many multi-scale computer vision tasks on various datasets, then this could have some far-reaching consequences for the computer vision field. Currently, many research efforts namely go towards the design of high-performing backbones~\cite{liu2021swin, liu2022convnet} by optimizing their performance on the ImageNet~\cite{deng2009imagenet} classification benchmark, with the hypothesis that these backbone networks are also optimal for the downstream multi-scale computer vision tasks found in practice. If this hypothesis hence turns out to be false (as recent findings seem to suggest), then the computer vision field should move beyond the design of optimal networks for the image classification task, and instead transition to designing optimal (neck) networks for a more fundamental (multi-scale) computer vision task, with better generalization to other multi-scale computer vision tasks found in practice.

\section{Conclusion}
In this chapter, we introduce the TPN neck consisting of top-down, self-processing and bottom-up operations. By considering the self-processing operation as an independent entity separated from the top-down and bottom-up operations, the TPN neck is able to find a better balance between communication-based processing and self-processing compared to existing necks as BiFPN. We validate our findings on the COCO object detection benchmark where we show the superiority of the TPN neck compared to the BiFPN neck. We additionally observe that moving computation from the TPN neck to the backbone or head decreases performance, highlighting the importance and effectiveness of the neck component within object detection networks.

\section*{Acknowledgements}
This work was partly supported by the KU Leuven C1 MACCHINA project.

\cleardoublepage
  \graphicspath{{\chaptersdir/detection/image/}}%
  \chapter{FQDet: Fast-converging Query-based Detector} \label{ch:detection}

\definecolor{shadecolor}{gray}{0.85}
\begin{shaded}
    The work in this chapter was adapted from the following conference proceeding: \vspace{-0.1cm} \\
    \bibentry{picron2022fqdet}
\end{shaded}

\section{Introduction}
In this chapter, we design a head network for the object detection task (see Section~\ref{sec:ch2_meta_arch} for more information about the backbone-neck-head meta architecture). The object detection head is defined as taking in a feature pyramid from the neck, and outputting object detection predictions, with each prediction consisting of an axis-aligned box together with a corresponding class label and confidence score. Object detection heads are commonly subdivided into \textit{one-stage} and \textit{two-stage} heads. One-stage heads make an object detection prediction for \textit{every feature} from its input feature pyramid. Two-stage heads first evaluate for every feature whether it is related to an object, and then use these results to \textit{focus the processing} on object-related parts of the feature pyramid.

In what follows, we exclusively work and compare with two-stage heads. Two-stage heads can further be subdivided in \textit{region-based} and \textit{query-based} two-stage heads, based on how the second stage of the two-stage head is implemented. 

Region-based two-stage heads pool a rectangular grid of features from the neck feature pyramid and further process these using convolutional neural networks (CNNs) in order to obtain their final object predictions. Region-based heads are found in many well-established two-stage object detectors such as Faster R-CNN~\cite{ren2015faster} and Cascade R-CNN~\cite{cai2018cascade}.

Query-based two-stage heads were recently introduced in the two-stage variant of Deformable DETR~\cite{zhu2020deformable}. While region-based heads extract a \textit{grid of features} from the neck feature pyramid per detection, only a \textit{single feature} called the query is selected in query-based heads. These queries are then further processed using operations commonly found in a transformer decoder~\cite{vaswani2017attention}, namely cross-attention, self-attention, and feedforward operations. Here, the cross-attention operation is defined between the neck feature pyramid and each query, the self-attention operation is applied between the different queries, and the feedforward operations are performed on each query individually.

In this chapter, we propose new query-based two-stage heads, called the \gls{FQDet} (Fast-converging Query-based Detector) and \gls{FQDetV2} (FQDet Version~2) heads. The FQDet heads were obtained by combining the query-based paradigm from two-stage Deformable DETR~\cite{zhu2020deformable} with classical object detection techniques such as anchor generation and static (\ie non-Hungarian) matching.

When evaluated on the 2017 COCO object detection benchmark~\cite{lin2014microsoft} after training for $12$ epochs, our FQDet head with ResNet-50 backbone~\cite{he2016deep} and TPN neck (see Chapter~\ref{ch:neck}) achieves $45.4$ AP, outperforming other high-performing two-stage heads such as Faster R-CNN~\cite{ren2015faster}, Cascade R-CNN~\cite{cai2018cascade}, two-stage Deformable DETR~\cite{zhu2020deformable}, and Sparse R-CNN~\cite{sun2021sparse}, while using the same backbone and neck networks. Additionally, when using the DefEnc-P3~\cite{zhu2020deformable} neck and after training for $12$~epochs, the enhanced FQDetV2 head obtains $50.8$~AP and $58.2$~AP with the ResNet-50~\cite{he2016deep} and Swin-L~\cite{liu2021swin} backbones respectively, outperforming the state-of-the-art (\gls{SOTA}) DINO~\cite{zhang2022dino} and Stable-DINO~\cite{liu2023detection} object detection heads. The code is released at \url{https://github.com/CedricPicron/FQDet}.

\section{Related work}

\paragraph{DETR and its variants.} With its unique way of approaching the object detection task, DETR~\cite{carion2020end} quickly gained a lot of popularity. Since, many variants such as SMCA~\cite{gao2021fast}, Conditional DETR~\cite{meng2021conditional}, Deformable DETR~\cite{zhu2020deformable}, Anchor DETR~\cite{wang2022anchor}, DAB-DETR~\cite{liu2022dab} and DN-DETR~\cite{li2022dn} have been proposed to improve the main two shortcomings of DETR, namely its slow convergence speed and its poor performance on small objects. In doing so, Deformable DETR~\cite{zhu2020deformable} introduces the query-based two-stage head, a new type of two-stage head now also used in other works such as DINO~\cite{zhang2022dino}.

\paragraph{Anchors.} Our FQDet head differs from other query-based two-stage heads by introducing anchors within the query-based two-stage paradigm. Faster R-CNN~\cite{ren2015faster} was one of the first works to use anchors. Anchors are axis-aligned boxes of different sizes and aspect ratios that are attached to feature pyramid locations and are refined to yield the final bounding box predictions. Anchors are found in both one-stage detectors (SSD~\cite{liu2016ssd}, YOLO~\cite{redmon2016you}, RetinaNet~\cite{lin2017focal}) and two-stage detectors (Faster R-CNN~\cite{ren2015faster}, Cascade R-CNN~\cite{cai2018cascade}). Over the years, many anchor-free detectors have also been proposed such as CornerNet~\cite{law2018cornernet}, FCOS~\cite{tian2019fcos}, CenterNet~\cite{zhou2019objects}, FoveaBox~\cite{kong2020foveabox}, DETR~\cite{carion2020end} and Deformable DETR~\cite{zhu2020deformable}.

\paragraph{Matching.} Our FQDet head additionally differs from other query-based two-stage heads by using a static top-k matching scheme as opposed to the dynamic Hungarian matching scheme. Matching is the process responsible of assigning ground-truth labels to the different head outputs during training. Most detectors have a \textit{static} matching scheme, meaning that the matching scheme stays the same throughout the whole training process. Static matching schemes are common when using anchors, where label assignment is done based on the IoU overlap between anchors and ground-truth boxes (\eg Faster R-CNN~\cite{ren2015faster}, RetinaNet~\cite{lin2017focal} and Cascade R-CNN~\cite{cai2018cascade}). Some works make use of a \textit{dynamic} matching scheme, meaning that the matching scheme changes as training progresses. In Dynamic R-CNN~\cite{zhang2020dynamic}, the minimum IoU thresholds are increased throughout the training process. In DETR \cite{carion2020end}, a dynamic Hungarian matching scheme is used, matching every ground-truth to a single prediction (one-to-one matching). In OTA \cite{ge2021ota}, the one-to-one Hungarian matching scheme from DETR is extended to one-to-many Hungarian matching, where a single ground-truth could be assigned to multiple predictions.

\section{Method}

\subsection{Revisiting DETR and Deformable DETR}
\paragraph{DETR.} DETR~\cite{carion2020end} solves the object detection task using transformers~\cite{vaswani2017attention}, originally designed to solve natural language processing (NLP) tasks. By doing so, DETR removes many hand-crafted components such as anchor generation and non-maximum suppression (\gls{NMS}), though it comes at a cost. DETR needs more than 100 epochs to converge, which are many more than the dozen of epochs traditional detectors such as Faster R-CNN~\cite{ren2015faster} need. Additionally, DETR does not perform well on small objects, as the quadratic complexity of self-attention layers prohibit the use of feature pyramids~\cite{lin2017feature} inside the transformer encoder.

\paragraph{Vanilla Deformable DETR.} Not much later, Deformable DETR~\cite{zhu2020deformable} is proposed. Deformable DETR is a DETR-like object detector mitigating the two main shortcomings of DETR, namely the slow convergence and the poor performance on small objects. Most of Deformable DETR's improvements can be attributed to the replacement of vanilla self-attention and cross-attention operations with new self-attention and cross-attention operations based on \textit{multi-scale deformable attention} (\gls{MSDA}). MSDA could be considered as an extension of the deformable convolution operations~\cite{dai2017deformable, zhu2019deformable} by introducing a weighting mechanism similar to the one used in the attention-based literature~\cite{vaswani2017attention}. Fundamentally, MSDA remains a local operation like traditional convolutions, in contrast to the global attention-based operation. By leveraging the local priors from MSDA, Deformable DETR is able to significantly speed up the convergence compared to DETR. Additionally, MSDA is a \textit{multi-scale} operation, meaning that it can operate on feature pyramids. This allows Deformable DETR to use feature pyramids at the encoder level by introducing the DefEncoder neck, and to apply the cross-attention operation on the neck feature pyramid at the decoder level, considerably improving the performance on small objects.

\paragraph{Two-stage Deformable DETR.} In DETR-like detectors, the information of objects is captured in object queries or queries for short. Queries are updated within the decoder using cross-attention with the neck output feature pyramid (\ie the encoder output), self-attention between other queries, and feedforward computation on themselves. In DETR and vanilla Deformable DETR, these queries are initialized independently of the content of the image. This is sub-optimal as these queries should focus on regions containing objects, which could be inferred from the neck feature pyramid. Two-stage Deformable DETR addresses this issue by selecting features from the neck feature pyramid as initialization for the queries, further boosting the performance and convergence speed compared to vanilla Deformable DETR.

\subsection{Overview and motivation} \label{sec:ch4_overview_motiv}
\paragraph{Overview.} Below, we give an overview of the main improvements of our FQDet head over two-stage Deformable DETR:
\begin{enumerate}
    \item We use multiple anchors of different sizes and aspect ratios~\cite{lin2017focal}, improving the cross-attention prior.
    \item We encode the bounding boxes relative to these anchors, similar to Faster R-CNN~\cite{ren2015faster}.
    \item We only use an L1 loss for bounding box regression~\cite{ren2015faster} without GIoU loss~\cite{rezatofighi2019generalized}.
    \item We remove the auxiliary decoder losses and predictions as used in DETR~\cite{carion2020end}.
    \item We do not perform iterative bounding box regression as in Deformable DETR~\cite{zhu2020deformable}.
    \item We replace Hungarian matching~\cite{carion2020end, zhu2020deformable} with static top-k matching (\textit{ours}).
\end{enumerate}

\paragraph{Motivation.} In what follows, we motivate each of the above FQDet design choices. The impact of these design choices on the performance will be analyzed in the ablation studies of Subsection~\ref{sec:ch4_base_abl}.

\textit{(Item~1)} We first motivate the use of anchors in our FQDet head. To understand this, we must take a look at the cross-attention operation from Deformable DETR. This operation updates a query based on features sampled from the neck feature pyramid, where the sample coordinates are regressed from the query itself. These sample coordinates are defined \wrt to a reference frame based on the query box prior, where the origin of the reference frame is placed at the center of the box prior, and where the unit lengths correspond to half the width and height of the box prior. By using anchors of various sizes and aspect ratios in our FQDet head, we significantly improve these cross-attention box priors resulting in more accurate and robust sampling, and eventually in better performance.

Items 2-5 from above list now in fact all follow from the use of anchors. \textit{(Item~2)} Given that we use anchors, it is logical to also encode our bounding box predictions relative to these anchors as done in Faster R-CNN~\cite{ren2015faster}. \textit{(Item~3)} Since we encode our boxes relative to the anchors, we only use an L1 loss (without GIoU loss) for the bounding box regression, as commonly used for this box encoding~\cite{ren2015faster}. \textit{(Item~4)} Since we improved our box priors with the introduction of anchors, there is no need to further improve these by introducing intermediate box predictions supervised by auxiliary losses as in Deformable DETR~\cite{zhu2020deformable}. In our setting, updating the box priors with these intermediate box predictions even hurts the performance, as a fixed sample reference frame for the different decoder layers, is to be preferred over a changing one. \textit{(Item~5)} Given that we do not make intermediate box predictions, there is also no need for iterative bounding box regression as used in Deformable DETR~\cite{zhu2020deformable}.

\textit{(Item~6)} Our final design choice involves replacing the Hungarian matching with our static top-k matching. At the beginning of training, Hungarian matching might assign a ground-truth detection to a query which sampled from a very different region compared to the location of the ground-truth detection. Hungarian matching hence produces many low-quality matches at the beginning of training, significantly slowing down convergence. Instead, we propose our static (\ie non-Hungarian) top-k matching scheme consisting of matching each ground-truth detection with its top-k anchors. Queries corresponding to these anchors are then also automatically matched. This matching scheme guarantees that each matched query indeed did processing in the neighborhood of the matched ground-truth detection.

\subsection{FQDet in detail}
\begin{figure}
    \centering
    \includegraphics[angle=90, scale=1.17]{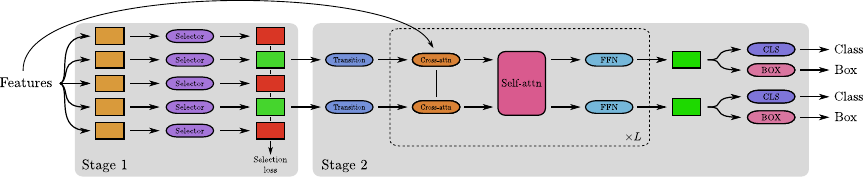}
    \caption[FQDet - Architectural overview]{Architectural overview of our two-stage FQDet head.}
    \label{fig:FQDet}
\end{figure}

An architectural overview of our FQDet head is displayed in Figure~\ref{fig:FQDet}. The head consists of two stages: a first stage applied to all input features and a second stage applied only to those features (\ie queries) selected from Stage~1. We now discuss both stages in more depth.

\paragraph{Stage 1.} In Stage~1, the goal of our FQDet head is to determine which input features belong to an object, such that these can be selected for Stage~2. At its input, the head receives the neck output feature pyramid. First, we attach anchors of various sizes and aspect ratios to each neck feature of the input feature pyramid. A lightweight selector network is then applied to each neck feature outputting a score for each feature-anchor combination, with top-scoring feature-anchor combinations selected for Stage~2. The selected features for Stage~2 are called queries and the corresponding anchors form the query box priors for the MSDA cross-attention operation.

A selection loss is applied during training to learn which features-anchor combinations should be selected. The selection targets are determined by our static top-k matching scheme, where anchors with a top-k highest IoU with one of the ground-truth boxes receive a positive label, and anchors with no top-k highest IoU with any of the ground-truth boxes receive a negative label.

\paragraph{Stage 2.} In Stage~2, the goal of our FQDet head is to make an object detection prediction for each of the queries selected from Stage~1. First, the queries are passed through a lightweight transition network. This transition network is meant to change the queries from the neck feature space to the head feature space, as
both feature spaces might have a different size.

Once processed by the transition network, the queries are passed through $L$ consecutive transformer decoder layers. Each layer consists of a sequence of a cross-attention operations, a self-attention operations, and a feedforward network (\gls{FFN}). The \textit{cross-attention} operation updates each query by sampling features from the neck feature pyramid using the multi-scale deformable attention (\gls{MSDA}) operation from Deformable DETR~\cite{zhu2020deformable}. The \textit{self-attention} operation updates each query by interacting with other queries using a standard multi-head attention (\gls{MHA}) operation \cite{vaswani2017attention}. Finally, each query is also processed individually by a \textit{feedforward network} (FFN)~\cite{vaswani2017attention}, which can be considered as a specific type of multi-layer perceptron (\gls{MLP}).

After the decoder layers, the queries are processed by the classification and bounding box regression networks yielding the object detection predictions. The classification network outputs a score for each of the classes within the dataset, along with a score for the non-object class. The bounding box regression network outputs the bounding box deltas \wrt the anchor, using the same box encoding scheme as in Faster R-CNN~\cite{ren2015faster}. Both classification and bounding box regression networks are implemented using lightweight MLPs. 

We use the same static top-k matching scheme in Stage~2 as in Stage~1, except that we allow for a different $k$ in both stages.

\section{FQDet Experiments}
\subsection{Setup} \label{sec:ch4_setup_v1}

\paragraph{Dataset.} We perform object detection experiments on the COCO detection dataset~\cite{lin2014microsoft}, where we train on the 2017 COCO training set and evaluate on the 2017 COCO validation set. See Section~\ref{sec:ch2_tasks} for more information about the object detection task and the COCO dataset.

\paragraph{Implementation details.} We follow the standardized settings mentioned in Section~\ref{sec:ch2_settings} for the backbone settings, neck settings, data settings, and (some of the) training settings. In what follows, we provide the TPN settings, the FQDet head settings, and the remaining training settings.

For the \textbf{TPN network} (see Chapter~\ref{ch:neck}), we use the TPN configuration with $L=3$ TPN layers and $B=2$ bottleneck layers per self-processing operation. The TPN network operates on a feature pyramid containing the $P_3$-$P_7$ feature maps. Recall that feature map~$P_l$ corresponds to the map which is $2^l$ times smaller in width and height compared to the input image resolution.

For the selector network of our \textbf{FQDet head}, we use one hidden layer and one output layer. The hidden layer consists of a bottleneck layer~\cite{he2016deep} with the same settings as used in Section~\ref{sec:ch3_setup}. The output layer is a linear projection layer projecting each feature to a selection score for each of the anchor types. We use $9$ anchor types of different size and aspect ratio as defined in RetinaNet~\cite{lin2017focal}. For the selection loss, we use a sigmoid focal loss~\cite{lin2017focal} with $\alpha=0.25$ and $\gamma=2.0$ as hyperparameters. The selection targets are determined using our static top-k matching scheme with $k=5$.

The $300$~feature-anchor combinations (\ie query and query prior boxes) with the highest score are selected for the second stage. The queries pass first through one of the $9$~transition networks, depending on their corresponding anchor type. Each of these transition networks consist of a linear projection layer with output feature size $256$, preceded by a LayerNorm~\cite{ba2016layer} and a ReLU activation layer.

The transformer decoder of the FQDet head consists of $6$ consecutive decoder layers, with each layer a sequence of cross-attention, self-attention and feedforward operations. Each of the operations updates the queries by using skip connections, with each residual branch starting with a LayerNorm layer~\cite{ba2016layer}.

The cross-attention operation is implemented by the multi-scale deformable attention (MSDA) operation from Deformable DETR~\cite{zhu2020deformable}. We use $8$~MSDA attention heads and $4$~MSDA sampling points per head and feature pyramid level. For the self-attention operation, we use the multi-head attention operation (MHA) from \cite{vaswani2017attention} with $8$ attention heads. The feedforward network (FFN) is implemented as in \cite{vaswani2017attention} with a hidden feature size of $1024$, except that we remove the dropout layers.

Additionally, box encodings are added to the queries when passed to the residual branches of the cross-attention and self-attention operations. These box encodings could be interpreted as enhanced positional encodings containing an additional notion of width and height. The box encodings are learned from the query box priors using an MLP with one hidden layer.

For the FQDet classification and bounding box regression networks, we use in both cases an MLP with one hidden layer. The classification and bounding box ground-truth labels are obtained using our static top-k matching scheme with $k=15$. Unmatched queries receive the \textit{non-object} label. For the classification loss, we use a sigmoid focal loss \cite{lin2017focal} with the same hyperparameters as for the selection loss. For the bounding box regression loss, we use an L1 loss in the delta box encoding space as done in Faster R-CNN~\cite{ren2015faster}. We remove duplicate detections during inference by using non-maximum suppression (\gls{NMS}) with an IoU threshold of $0.5$.

We \textbf{train} our FQDet models using the 1x schedule, consisting of $12$ training epochs with learning rate drops after the $9$th and $11$th epoch with a factor $0.1$. The models are trained on $2$ GPUs, each having a batch size of $2$.

\paragraph{Metrics.} As performance metrics, we use the COCO \gls{AP} metrics, and as computational cost metrics, we use the Params, tFPS, tMem and iFPS metrics. The computational cost metrics are obtained on a GeForce GTX 1660 Ti GPU, by applying the network on a batch of two $800\times800$ images, each containing ten ground-truth objects during training. See Subsection~\ref{sec:ch2_metrics} for more information about these metrics.

\subsection{Base FQDet results and ablation studies} \label{sec:ch4_base_abl}
In this subsection, we provide the results of the base FQDet head (\ie the FQDet head with the settings as explained above), and perform various ablation studies \wrt this head. In Subsection~\ref{sec:ch4_improving}, some small modifications are made to the base FQDet head yielding improved performance, and in Subsection~\ref{sec:ch4_comp} the improved FQDet head is compared with four different baseline two-stage heads from the literature.

\paragraph{Base FQDet} In Table~\ref{tab:ch4_FQDet_base}, we report the results of our base FQDet head, obtaining $44.3$~AP after only $12$~epochs of training. In what follows, we conduct a series of ablation studies investigating various FQDet design choices.

\begin{table*}[t!]
    \centering
    \caption[FQDet - Base results]{Base FQDet results on the COCO validation set.}
    \vspace{0.2cm}
    \resizebox{\columnwidth}{!}{
        \begin{tabular}{c c|c c c c c c|c c c c}
            \toprule
            Head & Epochs & AP & AP$_{50}$ & AP$_{75}$ & AP$_S$ & AP$_M$ & AP$_L$ & Params & tFPS & tMem & iFPS \\ \midrule
            FQDet & $12$ & $44.3$ & $62.5$ & $47.8$ & $25.3$ & $48.3$ & $59.0$ & $42.6$ M & $1.6$ & $2.86$ GB & $4.8$ \\
            \bottomrule
        \end{tabular}}
    \label{tab:ch4_FQDet_base}
\end{table*}

\paragraph{Anchors ablation study.} In the anchors ablation study, we vary the number of anchor types used in the FQDet selector. In our base FQDet head, we use $9$ anchor types composed of $3$~anchor sizes and $3$~anchor aspect ratios. In Table~\ref{tab:ch4_anchors}, we show the FQDet results when the number of anchor sizes and/or the number of anchor aspect ratios is reduced to~$1$. We can see that the number of anchor types play an important role in the resulting performance, with a performance drop up to $3.7$ AP when only a single anchor type is used. Moreover, we observe that the additional computational cost of using multiple anchor types is very limited. These results support our analysis from Subsection~\ref{sec:ch4_overview_motiv}, where we highlight the importance of having good query box priors (\ie anchors) for the MSDA cross-attention operation.

\begin{table*}[t!]
    \centering
    \caption[FQDet ablation - Anchors]{FQDet anchors ablation study on the COCO validation set.}
    \vspace{0.2cm}
    \resizebox{\columnwidth}{!}{
        \begin{tabular}{c c c|c c c c c c|c c c c}
            \toprule
            Head & Sizes & Ratios & AP & AP$_{50}$ & AP$_{75}$ & AP$_S$ & AP$_M$ & AP$_L$ & Params & tFPS & tMem & iFPS \\ \midrule
            FQDet & $1$ & $1$ & $40.6$ & $60.8$ & $42.4$ & $24.1$ & $44.6$ & $53.4$ & $42.0$ M & $1.7$ & $2.84$ GB & $5.0$ \\
            FQDet & $3$ & $1$ & $42.9$ & $61.4$ & $46.2$ & $24.4$ & $47.3$ & $57.5$ & $42.2$ M & $1.7$ & $2.84$ GB & $5.0$ \\
            FQDet & $1$ & $3$ & $43.1$ & $62.6$ & $46.1$ & $25.8$ & $46.8$ & $57.0$ & $42.2$ M & $1.7$ & $2.84$ GB & $5.0$ \\
            FQDet & $3$ & $3$ & $44.3$ & $62.5$ & $47.8$ & $25.3$ & $48.3$ & $59.0$ & $42.6$ M & $1.6$ & $2.86$ GB & $4.8$ \\
            \bottomrule
        \end{tabular}}
    \label{tab:ch4_anchors}
\end{table*}

\paragraph{Auxiliary losses ablation study.} In the auxiliary losses ablation study, we investigate whether making intermediate predictions after each decoder layer supervised by auxiliary losses as in Deformable DETR \cite{zhu2020deformable}, improves the performance of our base FQDet head. In Table~\ref{tab:ch4_aux}, we show the performance of our FQDet head where auxiliary losses are added, both with and without shared classification and bounding box regression networks. We observe that the performance decreases between $0.2$ and $0.6$ AP, depending on whether the regression networks are shared or not. We argue that as our FQDet head already provides good box priors through its Stage~1 anchors for the Stage~2 decoder, there is no need to further improve these through intermediate box predictions supervised by auxiliary losses.

\begin{table*}[t!]
    \centering
    \caption[FQDet ablation - Auxiliary losses]{FQDet auxiliary losses ablation study on the COCO validation set.}
    \vspace{0.2cm}
    \resizebox{\columnwidth}{!}{
        \begin{tabular}{c c c|c c c c c c|c c c c}
            \toprule
            Head & Aux. & Shared & AP & AP$_{50}$ & AP$_{75}$ & AP$_S$ & AP$_M$ & AP$_L$ & Params & tFPS & tMem & iFPS \\ \midrule
            FQDet & \cmark & \cmark & $44.1$ & $62.4$ & $47.5$ & $25.2$ & $48.0$ & $59.3$ & $42.6$ M & $1.5$ & $2.88$ GB & $4.5$ \\
            FQDet & \cmark & \xmark & $43.7$ & $62.1$ & $46.7$ & $25.3$ & $47.6$ & $58.4$ & $43.5$ M & $1.5$ & $2.89$ GB & $4.5$ \\
            FQDet & \xmark & $-$ & $44.3$ & $62.5$ & $47.8$ & $25.3$ & $48.3$ & $59.0$ & $42.6$ M & $1.6$ & $2.86$ GB & $4.8$ \\
            \bottomrule
        \end{tabular}}
    \label{tab:ch4_aux}
\end{table*}

\paragraph{IBBR ablation study.} In the iterative bounding box regression (\gls{IBBR}) ablation study, we again investigate the setting with intermediate predictions supervised by auxiliary losses as in previous ablation study, except that we also do iterative bounding box regression (IBBR) as introduced in Deformable DETR \cite{zhu2020deformable}. IBBR consists in iteratively updating the bounding box predictions relative to the previous bounding box predictions (or anchors). The results of our IBBR ablation study are found in Table~\ref{tab:ch4_ibbr}, where we again consider the cases with and without shared classification and bounding box regression networks. We can see that the performance again decreases compared to our base FQDet head, this time between $0.5$ and $0.8$ AP depending on whether the regression networks are shared or not. Adding IBBR to the setting of intermediate predictions supervised by auxiliary losses, hence does not yield improvements for our base FQDet head.

\begin{table*}[t!]
    \centering
    \caption[FQDet ablation - IBBR]{FQDet IBBR ablation study on the COCO validation set.}
    \vspace{0.2cm}
    \resizebox{\columnwidth}{!}{
        \begin{tabular}{c c c c|c c c c c c|c c c c}
            \toprule
            Head & Aux. & IBBR & Shared & AP & AP$_{50}$ & AP$_{75}$ & AP$_S$ & AP$_M$ & AP$_L$ & Params & tFPS & tMem & iFPS \\ \midrule
            FQDet & \cmark & \cmark & \cmark & $43.5$ & $62.0$ & $46.5$ & $24.7$ & $47.6$ & $58.1$ & $42.6$ M & $1.5$ & $2.89$ GB & $4.4$ \\
            FQDet & \cmark & \cmark & \xmark & $43.8$ & $62.2$ & $46.7$ & $24.7$ & $47.7$ & $58.8$ & $43.5$ M & $1.5$ & $2.90$ GB & $4.4$ \\
            FQDet & \xmark & \xmark & $-$ & $44.3$ & $62.5$ & $47.8$ & $25.3$ & $48.3$ & $59.0$ & $42.6$ M & $1.6$ & $2.86$ GB & $4.8$ \\
            \bottomrule
        \end{tabular}}
    \label{tab:ch4_ibbr}
\end{table*}

\paragraph{Matching ablation study.} In the matching ablation study, we compare our static top-k matching scheme with the static absolute matching scheme from Faster R-CNN~\cite{ren2015faster}, and the Hungarian matching scheme from Deformable DETR~\cite{zhu2020deformable}. In Table~\ref{tab:ch4_matching}, we show the results when using our FQDet head with each of these three matching schemes. We can see that our top-k matching scheme works best, outperforming the absolute matching scheme by $1.0$ AP and the Hungarian scheme by $9.2$ AP. These 12-epoch results are in line with the analysis from Subsection~\ref{sec:ch4_overview_motiv}, where we argue that Hungarian matching leads to slow convergence due to the low-quality matches at the start of training.

\begin{table*}[t!]
    \centering
    \caption[FQDet ablation - Matching]{FQDet matching ablation study on the COCO validation set.}
    \vspace{0.2cm}
    \resizebox{\columnwidth}{!}{
        \begin{tabular}{c c|c c c c c c|c c c c}
            \toprule
            Head & Matching & AP & AP$_{50}$ & AP$_{75}$ & AP$_S$ & AP$_M$ & AP$_L$ & Params & tFPS & tMem & iFPS \\ \midrule
            FQDet & Hungarian & $35.1$ & $54.2$ & $37.5$ & $20.3$ & $38.5$ & $43.8$ & $42.6$ M & $1.5$ & $2.86$ GB & $4.8$ \\
            FQDet & Absolute & $43.3$ & $61.0$ & $46.8$ & $25.5$ & $47.4$ & $57.6$ & $42.6$ M & $1.7$ & $2.86$ GB & $4.8$ \\
            FQDet & Top-k (\textit{ours}) & $44.3$ & $62.5$ & $47.8$ & $25.3$ & $48.3$ & $59.0$ & $42.6$ M & $1.6$ & $2.86$ GB & $4.8$ \\
            \bottomrule
        \end{tabular}}
    \label{tab:ch4_matching}
\end{table*}

\subsection{Improving the base FQDet head} \label{sec:ch4_improving}
In what follows, we make following three improvements to the base FQDet head. The results after applying each of those improvements sequentially, are shown in Table~\ref{tab:ch4_improvements}.

\begin{table*}[t!]
    \centering
    \caption[FQDet - Improving base FQDet]{Improving the base FQDet head by changing the number of MSDA sampling points (Pts), altering the inference strategy (Inf), and training longer (evaluated on the COCO validation set).}
    \vspace{0.2cm}
    \resizebox{\columnwidth}{!}{
        \begin{tabular}{c c c c|c c c c c c|c c c}
            \toprule
            Head & Pts & Inf & Epochs & AP & AP$_{50}$ & AP$_{75}$ & AP$_S$ & AP$_M$ & AP$_L$ & Params & tFPS & iFPS \\ \midrule
            FQDet & $4$ & Old & $12$ & $44.3$ & $62.5$ & $47.8$ & $25.3$ & $48.3$ & $59.0$ & $42.6$ M & $1.6$ & $4.8$ \\
            FQDet & $1$ & Old & $12$ & $44.5$ & $62.9$ & $47.7$ & $26.1$ & $48.7$ & $59.3$ & $42.0$ M & $1.7$ & $5.0$ \\
            FQDet & $1$ & New & $12$ & $45.4$ & $64.3$ & $48.9$ & $27.4$ & $49.6$ & $60.3$ & $42.0$ M & $1.7$ & $5.0$ \\
            \midrule
            FQDet & $1$ & New & $24$ & $46.2$ & $65.3$ & $49.2$ & $29.1$ & $50.3$ & $61.0$ & $42.0$ M & $1.7$ & $5.0$ \\
            \bottomrule
        \end{tabular}}
    \label{tab:ch4_improvements}
\end{table*}

\paragraph{MSDA sampling points improvement.} The first improvement consists in changing the number of MSDA sampling points per head and per feature pyramid level from $4$ to $1$, increasing the performance by $0.2$~AP while decreasing the computational cost.

\paragraph{Inference strategy improvement.} The second improvement consists in altering the inference strategy. In the current inference strategy, we only assign a single classification label to each predicted bounding box, \ie the label with the highest score. This is sub-optimal as the head might be hesitant about the class, with multiple classes containing similar scores. In the new inference strategy, we therefore consider all classification labels with their scores for each predicted bounding box, and feed these to the NMS algorithm to only keep the top~$100$ predictions. Replacing the old inference strategy with the new one improves the performance by $0.9$~AP, while leaving the computational cost virtually unchanged.

\paragraph{Training longer.} The performance of our FQDet head can further be improved by training with the 2x schedule, increasing the performance by $0.8$ AP. Here, the 2x schedule consists in training for $24$~epochs, with learning rate drops after the $18$th and $22$nd epochs with a factor~$0.1$.

\subsection{Comparison with other two-stage heads} \label{sec:ch4_comp}
In Table~\ref{tab:ch4_comparison}, we compare our improved FQDet head from Subsection~\ref{sec:ch4_improving} with other baseline two-stage heads from the literature. We compare with the two-stage heads from Faster R-CNN~\cite{ren2015faster}, Cascade R-CNN~\cite{cai2018cascade}, Deformable DETR~\cite{zhu2020deformable}, and Sparse R-CNN~\cite{sun2021sparse}, while using the same backbone, neck, data and training settings. We used their MMDetection implementations~\cite{mmdetection}, where we adapted the config files to be compatible with the $P_3$ to $P_7$ feature pyramid outputted by the neck.

\begin{table*}[t!]
    \centering
    \caption[FQDet - Comparison of FQDet with baselines]{Comparison of our FQDet head with other baseline two-stage heads on the COCO validation set.}
    \vspace{0.2cm}
    \resizebox{\columnwidth}{!}{
        \begin{tabular}{c c|c c c c c c|c c c c}
            \toprule
            Head & Epochs & AP & AP$_{50}$ & AP$_{75}$ & AP$_S$ & AP$_M$ & AP$_L$ & Params & tFPS & tMem & iFPS \\ \midrule
            Faster R-CNN~\cite{ren2015faster} & $12$ & $40.7$ & $60.9$ & $44.4$ & $23.9$ & $46.0$ & $54.1$ & $50.3$ M & $1.8$ & $2.62$ GB & $4.0$ \\
            Cascade R-CNN~\cite{cai2018cascade} & $12$ & $43.9$ & $61.9$ & $47.6$ & $25.8$ & $49.3$ & $58.5$ & $77.9$ M & $1.4$ & $3.15$ GB & $2.5$ \\
            Deformable DETR~\cite{zhu2020deformable} & $12$ & $38.9$ & $58.1$ & $41.6$ & $23.6$ & $41.7$ & $52.3$ & $43.0$ M & $1.6$ & $3.02$ GB & $4.9$ \\
            Sparse R-CNN~\cite{sun2021sparse} & $12$ & $39.8$ & $57.4$ & $43.1$ & $23.8$ & $42.5$ & $53.4$ & $114.8$ M & $1.3$ & $4.67$ GB & $4.0$ \\
            \midrule
            FQDet (ours) & $12$ & $\mathbf{45.4}$ & $\mathbf{64.3}$ & $\mathbf{48.9}$ & $\mathbf{27.4}$ & $\mathbf{49.6}$ & $\mathbf{60.3}$ & $42.0$ M & $1.7$ & $2.84$ GB & $5.0$ \\
            \bottomrule
        \end{tabular}}
    \label{tab:ch4_comparison}
\end{table*}

From Table~\ref{tab:ch4_comparison}, we observe the clear superiority of our FQDet head. The FQDet head outperforms all other two-stage heads with $1.5$ up to $6.5$~AP, not only outperforming the query-based two-stage Deformable DETR head, but also well-established region-based heads such as Cascade R-CNN. Additionally, our FQDet head has a considerably lower computational cost compared to most of the heads, only being slightly more expensive during training than the Faster R-CNN head.

\begin{figure}
    \centering
    \begin{minipage}{0.49\textwidth}
        \centering
        \includegraphics[scale=0.38]{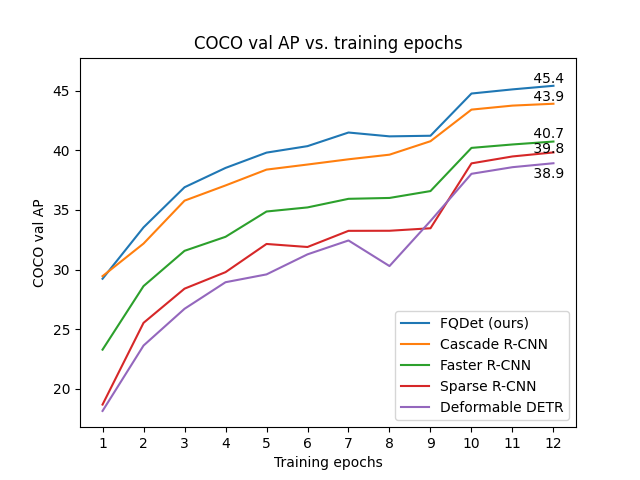}
    \end{minipage} \hfill
    \begin{minipage}{0.49\textwidth}
        \centering
        \includegraphics[scale=0.38]{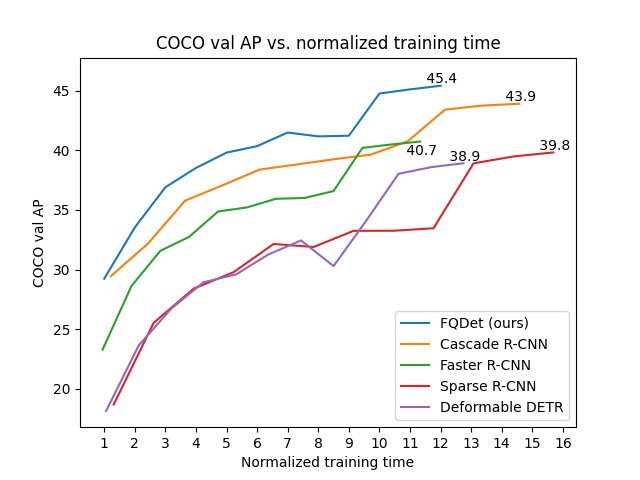}
    \end{minipage}
    \caption[FQDet - Convergence graphs]{(\textit{Left}) Convergence graph displaying the COCO validation AP after every training epoch for different two-stage heads. (\textit{Right}) Convergence graph displaying the COCO validation AP as a function of normalized training time. Here, one unit of normalized training time is defined as the time it takes to train our FQDet head for one epoch.}
    \label{fig:ch4_two-stage-comp}
\end{figure}

In Figure~\ref{fig:ch4_two-stage-comp}, we show two convergence graphs containing our FQDet two-stage head and other prominent two-stage heads using their MMDetection \cite{mmdetection} implementation. Here, the left convergence graph displays the COCO validation AP after every training epoch, and the right convergence graph shows the COCO validation AP after each unit of normalized time, with this normalized time unit defined as the time it takes to train our FQDet head for one epoch.

When looking at the left convergence graph, we can see that our FQDet head performs better throughout the whole training process, highlighting its fast convergence. Only the Cascade R-CNN head trails by a small amount. However, it takes substantially more time for the Cascade R-CNN head to complete one epoch compared to our FQDet head. Hence, when taking the actual training time into account in the right convergence graph, we now see that our FQDet head clearly outperforms the Cascade R-CNN head at any point during the training process. All in all, our FQDet head demonstrates excellent object detection performance after only a few training epochs, while remaining computationally cheap.

\section{FQDetV2 experiments}

\subsection{Setup} \label{sec:ch4_setup_v2}

\paragraph{Datset.} As before, we perform object detection experiments on the COCO detection dataset~\cite{lin2014microsoft}, where we train on the 2017 COCO training set and evaluate on the 2017 COCO validation set.

\paragraph{Implementation details.} Unless specified otherwise, we use the (improved) FQDet settings mentioned in Subsection~\ref{sec:ch4_setup_v1} and Subsection~\ref{sec:ch4_improving}. In what follows, we provide the FPN neck settings, the DefEncoder neck settings, and some updated training settings.

For the \textbf{FPN neck}~\cite{lin2017feature}, we use the implementation from MMDetection~\cite{mmdetection}, where we add extra convolutions on the lateral feature map $P_5$ and apply a ReLU activation function before each extra convolution. The FPN neck is used in the experiments of Subsection~\ref{sec:ch4_design_v2}, operating on the $P_3$-$P_7$ feature pyramid.

For the \textbf{DefEncoder neck}~\cite{zhu2020deformable}, we use the implementation from MMDetection~\cite{mmdetection}, following the same settings as in DINO~\cite{zhang2022dino}. The DefEncoder neck is used in the experiments of Subsection~\ref{sec:ch4_main_v2}, where we consider both the DefEnc-P2 and DefEnc-P3 variants operating on the $P_2$-$P_7$ and $P_3$-$P_7$ feature pyramids respectively. 

We \textbf{train} our FQDetV2 models using a 1x or 2x training schedule. The 1x schedule consists of 12 training epochs with a learning rate decrease after the 9th epoch with a factor~$0.1$, and the 2x schedule contains 24 training epochs with learning rate drops after the 18th and 23rd epochs with a factor~$0.1$.

For our design experiments in Subsection~\ref{sec:ch4_design_v2}, the remaining training settings are the same as the standardized settings from Subsection~\ref{sec:ch2_settings} and the additional FQDet settings from Subsection~\ref{sec:ch4_setup_v1}.

For our main experiments in Subsection~\ref{sec:ch4_main_v2}, we make following modifications to the training settings. We use an initial learning rate of $2\cdot10^{-5}$ for the ResNet-50~\cite{he2016deep} backbone parameters, an initial learning rate of $10^{-5}$ for the Swin-L~\cite{liu2021swin} backbone parameters, and an initial learning rate of $10^{-4}$ the linear projection parameters computing the \gls{MSDA} sampling offsets. For the remaining model parameters, we use an initial learning rate of $2\cdot10^{-4}$. We also apply gradient clipping, with a maximum gradient norm of $0.1$, similar to \cite{carion2020end}. The models are trained on 8 GPUs with a batch size of 2 each, and evaluated with a batch size of 1 each.

\paragraph{Metrics.} As performance metrics, we use the COCO AP metrics~\cite{lin2014microsoft}, and as computational cost metrics, we use the Params, iFLOPs and iFPS metrics. As no confusion is possible with training cost metrics, the iFLOPs and iFPS metrics are abbreviated as FLOPs and FPS respectively. The computational cost metrics are obtained on an NVIDIA GeForce RTX 3060 Ti GPU for the design experiments in Subsection~\ref{sec:ch4_design_v2}, and on an NVIDIA A100 GPU for the main experiments in Subsection~\ref{sec:ch4_main_v2}. Here, the cost metrics are computed with a batch size of~1 by averaging the costs over the first 100 images of the COCO validation set. See Subsection~\ref{sec:ch2_metrics} for more information about these metrics.

\subsection{FQDetV2 design experiments} \label{sec:ch4_design_v2}

In this subsection, we perform small-scale experiments to validate the \gls{FQDetV2} design choices, exclusively using the ResNet-50~\cite{he2016deep} backbone and the FPN~\cite{lin2017feature} neck. Then, in Subsection~\ref{sec:ch4_main_v2}, we perform larger-scale experiments with the DefEncoder~\cite{zhu2020deformable} neck and the ResNet-50~\cite{he2016deep} or Swin-L~\cite{liu2021swin} backbone, and compare with \gls{SOTA} object detection heads~\cite{zhang2022dino, liu2023detection, zong2022detrs}. Finally, in Subsection~\ref{sec:ch4_qualitative_v2}, we provide some qualitative results obtained with FQDetV2.

\begin{table*}[t!]
    \centering
    \caption[FQDetV2 design - FQDet Baseline]{FQDet baseline for the FQDetV2 design experiments.}
    \vspace{0.2cm}
    \resizebox{\columnwidth}{!}{
        \begin{tabular}{c c c c|c c c c c c|c c c}
            \toprule
            Bbone & Neck & Head & Ep. & AP & AP$_{50}$ & AP$_{75}$ & AP$_S$ & AP$_M$ & AP$_L$ & Params & FLOPs & FPS \\ \midrule
            R50 & FPN & FQDet & $12$ & $43.0$ & $61.8$ & $46.1$ & $24.9$ & $46.8$ & $58.5$ & $33.9$ M & $99.0$ G & $21.6$ \\
            \bottomrule
        \end{tabular}}
    \label{tab:ch4_baseline_v2}
\end{table*}

\paragraph{Baseline.} First, we establish our baseline model which we try to improve upon. The baseline model consists of the ResNet-50 backbone, the FPN neck, and the (improved) FQDet head from Subsection~\ref{sec:ch4_improving}. The results of our baseline model are shown in Table~\ref{tab:ch4_baseline_v2}. Note that the performance is lower compared to the FQDet performance reported in Table~\ref{tab:ch4_comparison}, as we use the computationally cheaper FPN neck instead of the TPN neck. 

\begin{table*}[t!]
    \centering
    \caption[FQDetV2 design - FQDet Baseline Variation]{Statistical variation of the FQDet baseline performance. The run with the median performance corresponds to the FQDet baseline from Table~\ref{tab:ch4_baseline_v2}.}
    \vspace{0.2cm}
    \resizebox{\columnwidth}{!}{
        \begin{tabular}{c c|c c c c c c|c c c}
            \toprule
            Head & Run & AP & AP$_{50}$ & AP$_{75}$ & AP$_S$ & AP$_M$ & AP$_L$ & Params & FLOPs & FPS \\ \midrule
            FQDet & Min & $42.9$ & $61.8$ & $46.1$ & $24.9$ & $46.4$ & $58.4$ & $33.9$ M & $99.0$ G & $21.6$ \\
            FQDet & Median & $43.0$ & $61.8$ & $46.1$ & $24.9$ & $46.8$ & $58.5$ & $33.9$ M & $99.0$ G & $21.6$ \\
            FQDet & Max & $43.3$ & $62.0$ & $46.4$ & $24.9$ & $47.0$ & $59.0$ & $33.9$ M & $99.0$ G & $21.6$ \\
            \bottomrule
        \end{tabular}}
    \label{tab:ch4_baseline_v2_var}
\end{table*}

Moreover, we have trained the FQDet baseline model multiple times in order to assess the statistical variation of the FQDet baseline performance. This statistical variation occurs due to following three stochastic elements: (1) the random order in which the training data is presented to the model during an epoch, (2) the randomness of the data augmentation operations during training, and (3) the random initialization of the neck and head network parameters. In total, we have trained the FQDet baseline model $9$~times. The results from the runs with the minimum, median, and maximum performance, are shown in Table~\ref{tab:ch4_baseline_v2_var}. We can see that the performance difference between the runs is small, with a performance difference of only $0.4$~AP between the best and the worst run. Additionally, the standard deviation between the runs is only $0.12$~AP. In general, despite not explicitly computing the statistical significance, we can conclude that improvements of $0.4$~AP or more are unlikely to be caused by chance.

In following paragraphs, we propose a series of modifications improving the baseline FQDet head. Note that these experiments will only be performed once, given the small statistical variation in performance metrics and as running each experiment multiple times would be computationally infeasible. Finally, we bring all of these modifications together into a single head called \gls{FQDetV2}.

\begin{table*}[t!]
    \centering
    \caption[FQDetV2 design - Anchor aspect ratios]{FQDetV2 design experiments varying the anchor aspect ratios.}
    \vspace{0.2cm}
    \resizebox{\columnwidth}{!}{
        \begin{tabular}{c c|c c c c c c|c c c}
            \toprule
            Head & Ratios & AP & AP$_{50}$ & AP$_{75}$ & AP$_S$ & AP$_M$ & AP$_L$ & Params & FLOPs & FPS \\ 
            \midrule
            FQDet & $3$ & $43.0$ & $61.8$ & $46.1$ & $24.9$ & $46.8$ & $58.5$ & $33.9$ M & $99.0$ G & $21.6$ \\
            FQDet & $5$ & $43.1$ & $61.8$ & $46.1$ & $26.3$ & $47.0$ & $58.2$ & $34.3$ M & $99.0$ G & $21.0$ \\
            FQDet & $\mathbf{7}$ & $43.1$ & $61.8$ & $46.1$ & $25.1$ & $47.0$ & $58.9$ & $34.7$ M & $99.0$ G & $20.8$ \\
            \bottomrule
        \end{tabular}}
    \label{tab:ch4_ratios_v2}
\end{table*}

\paragraph{Anchor aspect ratios.} The results of the FQDetV2 design experiments varying the number of anchor aspect ratios, are found in Table~\ref{tab:ch4_ratios_v2}. The top row contains the FQDet baseline which uses $3$ different aspect ratios, and in the two rows below, we additionally provide results when using $5$ and $7$ different aspect ratios. We observe a small performance gain from $43.0$~AP to $43.1$~AP when increasing the number of aspect ratios, but also a small increase in computational cost. Based on some experiments not shown here, we still decide to increase the number of aspect ratios to $7$ for the FQDetV2 head, though we admit the difference is small and not statistically significant. Note that in Table~\ref{tab:ch4_ratios_v2} and subsequent design experiment tables, we highlight the chosen setting for the FQDetV2 head in bold.

\begin{table*}[t!]
    \centering
    \caption[FQDetV2 design - Number of queries]{FQDetV2 design experiments varying the number of queries.}
    \vspace{0.2cm}
    \resizebox{\columnwidth}{!}{
        \begin{tabular}{c c|c c c c c c|c c c}
            \toprule
            Head & Queries & AP & AP$_{50}$ & AP$_{75}$ & AP$_S$ & AP$_M$ & AP$_L$ & Params & FLOPs & FPS \\ \midrule
            FQDet & $300$ & $43.0$ & $61.8$ & $46.1$ & $24.9$ & $46.8$ & $58.5$ & $33.9$ M & $99.0$ G & $21.6$ \\
            \midrule
            FQDet & $500$ & $44.1$ & $63.6$ & $47.3$ & $26.6$ & $48.0$ & $59.1$ & $33.9$ M & $100.6$ G & $21.5$ \\
            FQDet & $700$ & $44.0$ & $63.8$ & $47.0$ & $26.4$ & $48.0$ & $58.6$ & $33.9$ M & $102.5$ G & $21.2$ \\
            FQDet & $900$ & $44.1$ & $63.9$ & $47.6$ & $27.6$ & $48.0$ & $58.4$ & $33.9$ M & $104.7$ G & $20.8$ \\
            FQDet & $1100$ & $44.5$ & $64.5$ & $47.7$ & $27.3$ & $48.4$ & $59.0$ & $33.9$ M & $107.0$ G & $20.3$ \\
            FQDet & $1300$ & $44.5$ & $64.5$ & $47.7$ & $27.6$ & $48.3$ & $58.7$ & $33.9$ M & $109.7$ G & $19.4$ \\
            \midrule
            FQDet & $\mathbf{1500}$ & $44.7$ & $64.8$ & $48.0$ & $28.4$ & $48.8$ & $58.6$ & $33.9$ M & $112.6$ G & $18.7$ \\
            \midrule
            FQDet & $1800$ & $44.6$ & $64.5$ & $48.0$ & $27.3$ & $48.4$ & $58.9$ & $33.9$ M & $117.3$ G & $17.3$ \\
            FQDet & $2100$ & $44.7$ & $64.7$ & $48.2$ & $27.4$ & $48.7$ & $58.4$ & $33.9$ M & $122.7$ G & $16.5$ \\
            \bottomrule
        \end{tabular}}
    \label{tab:ch4_num_qrys_v2}
\end{table*}

\paragraph{Number of queries.} In Table~\ref{tab:ch4_num_qrys_v2}, we provide the results of the FQDetV2 design experiments varying the number of queries. In the top row, we have the FQDet baseline which uses $300$ queries, and in the rows below, we increase the number of queries from $500$ up until $2100$. Already when increasing the number of queries from $300$ to $500$, we observe a significant boost in performance with the AP rising from $43.0$ to $44.1$. The performance can further be increased to $44.5$~AP when using $1100$ queries, and to $44.7$~AP when using $1500$ queries. However, no further performance gains are observed when increasing the number of queries up to $2100$, while increasing the computational cost. For the FQDetV2 head, we therefore decide to increase the number of queries to $1500$, giving maximal performance with a reasonable increase in computational cost. 

We believe the use of a large number of queries (much larger than the number of objects in the image), is advantageous for two reasons. Firstly, more objects are found when using more queries. We think this is the main mechanism behind the $1.1$~AP performance increase when going from $300$ to $500$ queries. Secondly, as we allow multiple queries to match a single object, adding queries allows more matches per object, leading to more robust bounding box predictions. We believe this is an additional reason why performance gains are still seen when going from $500$ to $1500$ queries.

\begin{table*}[t!]
    \centering
    \caption[FQDetV2 design - FFN hidden size]{FQDetV2 design experiments varying the FFN hidden size.}
    \vspace{0.2cm}
    \resizebox{\columnwidth}{!}{
        \begin{tabular}{c c|c c c c c c|c c c}
            \toprule
            Head & Size & AP & AP$_{50}$ & AP$_{75}$ & AP$_S$ & AP$_M$ & AP$_L$ & Params & FLOPs & FPS \\ 
            \midrule
            FQDet & $1024$ & $43.0$ & $61.8$ & $46.1$ & $24.9$ & $46.8$ & $58.5$ & $33.9$ M & $99.0$ G & $21.6$ \\
            FQDet & $\mathbf{2048}$ & $43.1$ & $62.1$ & $45.8$ & $24.9$ & $46.8$ & $58.5$ & $37.1$ M & $99.9$ G & $21.6$ \\
            \bottomrule
        \end{tabular}}
    \label{tab:ch4_hidden_size_v2}
\end{table*}

\paragraph{FFN hidden size.} The results of the FQDetV2 design experiments varying the decoder \gls{FFN} hidden size, are found in Table~\ref{tab:ch4_hidden_size_v2}. When increasing the baseline FFN hidden size from $1024$ to $2048$, we observe a marginal gain in performance from $43.0$~AP to $43.1$~AP. The corresponding increase in computational cost is small for the FLOPs and FPS metrics, but is rather large for the Params metric rising from $33.9$~M to $37.1$~M. Despite the small, statistically insignificant performance gain and the increase in the number of parameters, we still decide to up the FFN hidden size to $2048$ for the FQDetV2 head, in order to be consistent with other object detection heads~\cite{zhang2022dino, liu2023detection, zong2022detrs}.

\begin{table*}[t!]
    \centering
    \caption[FQDetV2 design - Classification network]{FQDetV2 design experiments varying the classification network.}
    \vspace{0.2cm}
    \resizebox{\columnwidth}{!}{
        \begin{tabular}{c c|c c c c c c|c c c}
            \toprule
            Head & Cls net & AP & AP$_{50}$ & AP$_{75}$ & AP$_S$ & AP$_M$ & AP$_L$ & Params & FLOPs & FPS \\ 
            \midrule
            FQDet & V1 & $43.0$ & $61.8$ & $46.1$ & $24.9$ & $46.8$ & $58.5$ & $33.9$ M & $99.0$ G & $21.6$ \\
            FQDet & \textbf{V2} & $43.3$ & $62.4$ & $46.6$ & $26.0$ & $46.9$ & $59.2$ & $33.8$ M & $98.9$ G & $21.6$ \\
            \bottomrule
        \end{tabular}}
    \label{tab:ch4_cls_net_v2}
\end{table*}

\paragraph{Classification network.} In Table~\ref{tab:ch4_cls_net_v2}, we find the results of the FQDetV2 design experiments varying the classification network. Here, the baseline V1 classification network consists of an MLP with one hidden layer, and the new V2 classification network simply contains a single linear layer, similar to~\cite{zhang2022dino}. We can see from Table~\ref{tab:ch4_cls_net_v2} that the new V2 classification network increases the performance from $43.0$~AP to $43.3$~AP, while slightly decreasing the computational cost. Given these excellent results, we decide to incorporate the new V2 classification network in the FQDetV2 head.

\begin{table*}[t!]
    \centering
    \caption[FQDetV2 design - Classification loss]{FQDetV2 design experiments varying the classification loss.}
    \vspace{0.2cm}
    \resizebox{\columnwidth}{!}{
        \begin{tabular}{c c|c c c c c c|c c c}
            \toprule
            Head & Cls loss & AP & AP$_{50}$ & AP$_{75}$ & AP$_S$ & AP$_M$ & AP$_L$ & Params & FLOPs & FPS \\ 
            \midrule
            FQDet & V1 & $43.0$ & $61.8$ & $46.1$ & $24.9$ & $46.8$ & $58.5$ & $33.9$ M & $99.0$ G & $21.6$ \\
            FQDet & \textbf{V2} & $43.6$ & $62.1$ & $46.7$ & $26.4$ & $47.2$ & $59.4$ & $33.9$ M & $99.0$ G & $21.6$ \\
            \bottomrule
        \end{tabular}}
    \label{tab:ch4_cls_loss_v2}
\end{table*}

\paragraph{Classification loss.} The results of the FQDetV2 design experiments varying the classification loss, are found in Table~\ref{tab:ch4_cls_loss_v2}. Here, the baseline V1 classification loss uses the sigmoid focal loss~\cite{lin2017focal} with $\alpha=0.25$, $\gamma=2$, and loss weight equal to $1.0$. The new V2 classification loss on the other hand uses the quality focal loss~\cite{li2020generalized} with box IoU soft labels, $\beta=2$, and loss weight equal to~$0.5$. From Table~\ref{tab:ch4_cls_loss_v2}, we can see that the new V2 classification loss increases the performance from $43.0$~AP to $43.6$~AP, while leaving the inference computational cost unchanged. This performance gain can be attributed to the use of soft classification labels, where the box IoU value between predicted and target bounding box is used for positive matches instead of the value $1.0$. Because of this, the classification scores now also take the expected bounding box accuracy into account, leading to improved rankings of the different detections (with better detections coming first), in turn yielding increased AP values. Based on these findings, the new V2 classification loss is chosen for the FQDetV2 head.

\begin{table*}[t!]
    \centering
    \caption[FQDetV2 design - Box regression network]{FQDetV2 design experiments varying the box regression network.}
    \vspace{0.2cm}
    \resizebox{\columnwidth}{!}{
        \begin{tabular}{c c|c c c c c c|c c c}
            \toprule
            Head & Box net & AP & AP$_{50}$ & AP$_{75}$ & AP$_S$ & AP$_M$ & AP$_L$ & Params & FLOPs & FPS \\ 
            \midrule
            FQDet & V1 & $43.0$ & $61.8$ & $46.1$ & $24.9$ & $46.8$ & $58.5$ & $33.9$ M & $99.0$ G & $21.6$ \\
            FQDet & \textbf{V2} & $43.5$ & $62.5$ & $46.4$ & $26.4$ & $47.3$ & $59.0$ & $34.0$ M & $99.0$ G & $21.6$ \\
            \bottomrule
        \end{tabular}}
    \label{tab:ch4_box_net_v2}
\end{table*}

\paragraph{Box regression network.} In Table~\ref{tab:ch4_box_net_v2}, we provide the results of the FQDetV2 design experiments varying the box regression network. For the baseline V1 box regression network, we use an MLP with one hidden layer, and for the new V2 box regression network, we use an MLP with two hidden layers, similar to~\cite{zhang2022dino}. Replacing the V1 box regression network with the V2 box regression network, increases the performance from $43.0$~AP to $43.5$~AP, with only a marginal amount of additional computation. For the FQDetV2 head, we hence switch to the V2 box regression network.

\begin{table*}[t!]
    \centering
    \caption[FQDetV2 design - Box regression loss]{FQDetV2 design experiments varying the box regression loss.}
    \vspace{0.2cm}
    \resizebox{\columnwidth}{!}{
        \begin{tabular}{c c|c c c c c c|c c c}
            \toprule
            Head & Box loss & AP & AP$_{50}$ & AP$_{75}$ & AP$_S$ & AP$_M$ & AP$_L$ & Params & FLOPs & FPS \\ 
            \midrule
            FQDet & L1 & $43.0$ & $61.8$ & $46.1$ & $24.9$ & $46.8$ & $58.5$ & $33.9$ M & $99.0$ G & $21.6$ \\
            \midrule
            FQDet & CIoU & $43.3$ & $62.0$ & $46.5$ & $24.7$ & $46.9$ & $58.8$ & $33.9$ M & $99.0$ G & $21.6$ \\
            FQDet & GIoU & $43.4$ & $62.1$ & $46.3$ & $24.8$ & $47.2$ & $58.6$ & $33.9$ M & $99.0$ G & $21.6$ \\
            FQDet & DIoU & $43.5$ & $62.4$ & $46.7$ & $25.9$ & $47.3$ & $58.3$ & $33.9$ M & $99.0$ G & $21.6$ \\
            \midrule
            FQDet & \textbf{EIoU} & $43.6$ & $62.4$ & $46.7$ & $24.9$ & $47.4$ & $59.6$ & $33.9$ M & $99.0$ G & $21.6$ \\
            \bottomrule
        \end{tabular}}
    \label{tab:ch4_box_loss_v2}
\end{table*}

\paragraph{Box regression loss.} We provide the results of the FQDetV2 design expiriments varying the box regression loss in Table~\ref{tab:ch4_box_loss_v2}. The baseline loss is the L1 loss with loss weight~$1.0$. As new box regression losses, we consider the CIoU loss~\cite{zheng2021enhancing}, the GIoU loss~\cite{rezatofighi2019generalized}, the DIoU loss~\cite{zheng2020distance}, and the EIoU loss~\cite{peng2021systematic}, with loss weights equal to $1.3$, $1.3$, $1.3$, and $1.85$ respectively. Here, the loss weights are determined by making sure that the loss values are roughly the same compared to the baseline loss values. From the results of Table~\ref{tab:ch4_box_loss_v2}, we can see that all of the above IoU-based losses outperform the baseline L1 loss, with the EIoU working the best increasing the performance from $43.0$~AP to $43.6$~AP. For this reason, we decide to use the EIoU box regression loss for the FQDetV2 head.

\begin{table*}[t!]
    \centering
    \caption[FQDetV2 design - Box scoring]{FQDetV2 design experiments with and without box scoring.}
    \vspace{0.2cm}
    \resizebox{\columnwidth}{!}{
        \begin{tabular}{c c|c c c c c c|c c c}
            \toprule
            Head & Scoring & AP & AP$_{50}$ & AP$_{75}$ & AP$_S$ & AP$_M$ & AP$_L$ & Params & FLOPs & FPS \\ 
            \midrule
            FQDet & \xmark & $43.0$ & $61.8$ & $46.1$ & $24.9$ & $46.8$ & $58.5$ & $33.9$ M & $99.0$ G & $21.6$ \\
            FQDet & \cmark & $43.6$ & $61.5$ & $47.0$ & $25.8$ & $47.5$ & $59.4$ & $34.0$ M & $99.0$ G & $21.4$ \\
            \bottomrule
        \end{tabular}}
    \label{tab:ch4_box_scoring_v2}
\end{table*}

\paragraph{Box scoring.} In Table~\ref{tab:ch4_box_scoring_v2}, the results are found of the FQDetV2 design experiments with and without box scoring mechanism. The box scoring mechanism consists in producing a box confidence score, based on the \gls{IoU} between predicted and target bounding boxes. For this, we use an MLP with two hidden layers (similar to the V2 box regression network), supervised to predict the inverse sigmoid~\cite{zhu2020deformable} of the predicted and target box IoU during training. As box scoring loss, we use an L1 loss with loss weight~$0.2$, and during inference, we obtain the final detection scores by multiplying the classification scores with the box scores.

From the results in Table~\ref{tab:ch4_box_scoring_v2}, we observe an increase in performance from $43.0$~AP to $43.6$~AP at a small cost, when adding the box scoring mechanism. Especially on the AP$_{75}$ metric, we see a clear improvement from $46.1$ to $47.0$. This performance gain can be explained by the better ranking of the different detections, with better (\ie high IoU) detections coming first, leading to increased AP values. Given its effectiveness, we add the box scoring mechanism to the FQDetV2 head.

\begin{table*}[t!]
    \centering
    \caption[FQDetV2 design - NMS threshold]{FQDetV2 design experiments varying the NMS threshold.}
    \vspace{0.2cm}
    \resizebox{\columnwidth}{!}{
        \begin{tabular}{c c|c c c c c c|c c c}
            \toprule
            Head & NMS & AP & AP$_{50}$ & AP$_{75}$ & AP$_S$ & AP$_M$ & AP$_L$ & Params & FLOPs & FPS \\
            \midrule
            FQDet & $0.45$ & $42.7$ & $61.7$ & $45.8$ & $24.9$ & $46.5$ & $58.3$ & $33.9$ M & $99.0$ G & $21.6$ \\
            FQDet & $0.50$ & $43.0$ & $61.8$ & $46.1$ & $24.9$ & $46.8$ & $58.5$ & $33.9$ M & $99.0$ G & $21.6$ \\
            FQDet & $0.55$ & $43.0$ & $61.7$ & $46.2$ & $25.1$ & $46.8$ & $58.8$ & $33.9$ M & $99.0$ G & $21.6$ \\
            FQDet & $0.60$ & $43.1$ & $61.5$ & $46.6$ & $25.1$ & $47.0$ & $58.9$ & $33.9$ M & $99.0$ G & $21.6$ \\
            \midrule
            FQDet & $\mathbf{0.65}$ & $43.2$ & $61.2$ & $47.0$ & $25.0$ & $47.2$ & $59.0$ & $33.9$ M & $99.0$ G & $21.6$ \\
            FQDet & $\mathbf{0.70}$ & $43.1$ & $60.6$ & $47.2$ & $24.9$ & $47.2$ & $58.9$ & $33.9$ M & $99.0$ G & $21.6$ \\
            \midrule
            FQDet & $0.75$ & $42.9$ & $59.8$ & $47.2$ & $24.7$ & $47.1$ & $58.7$ & $33.9$ M & $99.0$ G & $21.6$ \\
            FQDet & $0.80$ & $42.5$ & $58.7$ & $47.0$ & $24.5$ & $46.8$ & $58.3$ & $33.9$ M & $99.0$ G & $21.6$ \\
            \bottomrule
        \end{tabular}}
    \label{tab:ch4_nms_v2}
\end{table*}

\paragraph{NMS threshold.} In FQDet, we use \gls{NMS} as duplicate removal mechanism, with an NMS IoU threshold (or NMS threshold for short) of $0.50$. In Table~\ref{tab:ch4_nms_v2}, we provide additional results of the FQDetV2 design experiments varying the NMS threshold between $0.45$ and $0.80$. We can see that the best performance is obtained for an NMS threshold of $0.65$, increasing the performance from $43.0$~AP to $43.2$~AP \wrt to the baseline NMS threshold of $0.50$. However, we notice in the larger-scale experiments of Subsection~\ref{sec:ch4_main_v2}, that an NMS threshold of $0.70$ works better there. For the FQDetV2 head, we therefore decide to use an NMS threshold of $0.65$ for the small-scale experiments in this subsection, and an NMS threshold of $0.70$ for the larger-scale experiments in Subsection~\ref{sec:ch4_main_v2}.

Interestingly, we additionally observe from Table~\ref{tab:ch4_nms_v2} that the optimal NMS thresholds vary depending on the AP metric, with an optimal NMS threshold of $0.50$ for the AP$_{50}$ metric, and an optimal NMS threshold of $0.70$ for the AP$_{75}$ metric. These findings can be explained as follows. The stricter NMS threshold of $0.50$, removes more duplicate detections, but also more correct detections of overlapping objects, with only inaccurate boxes (if any) remaining for highly overlapping objects. Hence, the NMS threshold of $0.50$ leads to many correct detections maximizing the AP$_{50}$ metric, but also to many inaccurate detections for overlapping objects decreasing the AP$_{75}$ performance. For the more permissive NMS threshold of $0.70$, the opposite is true, where more duplicate detections are retained, but also more accurate boxes for overlapping objects. As a result, the NMS threshold of $0.70$ does not perform as well on the AP$_{50}$ metric (due to remaining the duplicates), but maximizes the AP$_{75}$ metric by retaining accurate boxes for overlapping objects. One might now wonder if a better duplicate removal mechanism could be devised, maximizing both the AP$_{50}$ and AP$_{75}$ metrics at the same time, by removing duplicate detections and still keeping correct detections of overlapping objects. We explore this in next paragraph.

\begin{table*}[t!]
    \centering
    \caption[FQDetV2 design - Duplicate removal]{FQDetV2 design experiments varying the duplicate removal.}
    \vspace{0.2cm}
    \resizebox{\columnwidth}{!}{
        \begin{tabular}{c c|c c c c c c|c c c}
            \toprule
            Head & Removal & AP & AP$_{50}$ & AP$_{75}$ & AP$_S$ & AP$_M$ & AP$_L$ & Params & FLOPs & FPS \\
            \midrule
            FQDet & \textbf{NMS} & $43.2$ & $61.2$ & $47.0$ & $25.0$ & $47.2$ & $59.0$ & $33.9$ M & $99.0$ G & $21.6$ \\
            \midrule
            FQDet & Learned & $42.5$ & $61.1$ & $45.9$ & $24.5$ & $46.4$ & $57.7$ & $34.1$ M & $99.1$ G & $20.5$ \\
            FQDet & Soft-Learned & $42.8$ & $61.5$ & $46.3$ & $24.8$ & $46.7$ & $57.9$ & $34.1$ M & $99.1$ G & $10.0$ \\
            FQDet & Learn+NMS & $42.8$ & $60.9$ & $46.7$ & $24.7$ & $46.8$ & $57.9$ & $34.1$ M & $99.0$ G & $21.4$ \\
            \midrule
            FQDet & UpBnd-50 & $42.9$ & $63.1$ & $45.8$ & $25.1$ & $46.8$ & $58.4$ & $33.9$ M & $99.0$ G & $21.6$ \\
            FQDet & UpBnd-75 & $44.0$ & $61.0$ & $51.6$ & $25.7$ & $48.1$ & $59.7$ & $33.9$ M & $99.0$ G & $21.6$ \\
            \bottomrule
        \end{tabular}}
    \label{tab:ch4_dup_v2}
\end{table*}

\paragraph{Duplicate removal.} In Table~\ref{tab:ch4_dup_v2}, we provide the results of the FQDetV2 design experiments varying the duplicate removal mechanism. Our baseline is the NMS mechanism equipped with the optimal NMS threshold of $0.65$.

Firstly, we consider a learned duplicate removal mechanism. Here, we add a duplicate prediction network (or sub-head) to the FQDet head, which predicts for any pair of detections whether they are duplicates or not. The prediction network consists of applying an MLP with one hidden layer to each query individually, fusing query pairs using element-wise feature multiplication, followed by a linear layer obtaining the duplicate logits. During training, a positive duplicate target is assigned to detection pairs matching the same ground-truth object, and a negative duplicate target when matching different ground-truth objects. As duplicate loss, we use the cross-entropy loss function with loss weight~$1.0$.

The results of the learned duplicate removal mechanism, are found in the middle three rows of Table~\ref{tab:ch4_dup_v2}, which correspond to different ways of post-processing. The Learned variant removes the lowest-scoring duplicate prediction from pairs if the duplicate score is greater than $0.5$. In the Soft-Learned variant, instead of removing detections altogether, the detection score is altered by multiplying it with $(1 - d_{sc})^\alpha$, with $d_{sc}$ the duplicate score and $\alpha=2$ the score modification power. We can see from Table~\ref{tab:ch4_dup_v2}, that the Soft-Learned variant slightly improves the performance from $42.5$~AP to $42.8$~AP. However, note that the Soft-Learned variant is significantly slower compared to the other duplicate removal mechanisms, due to the use of a for-loop in its implementation. Further research is required to investigate whether the inference speed of the Soft-Learned variant can be increased or not.

We also report the results of the Learn+NMS variant, where we apply NMS (with optimal NMS threshold) on the FQDet model with the duplicate sub-head. We can see from Table~\ref{tab:ch4_dup_v2} that the variant also reaches $42.8$~AP, similar to the Soft-Learned variant. Learning which detections are duplicates hence performs on par with the hand-crafted NMS mechanism. However, by introducing an additional loss, the FQDet head with duplicate sub-head (reaching $42.8$~AP) performs slightly worse compared to the FQDet head without duplicate sub-head (reaching $43.2$~AP). Because of this, the FQDet head with the vanilla NMS duplicate removal mechanism remains the best option. Further exploring the learned duplicate removal mechanism is left as future work (see Section~\ref{sec:ch4_limit_future}).

Note that given the success of the Soft-Learned variant over the Learned variant, we also attempted to use Soft-NMS~\cite{bodla2017soft} to improve over base NMS. However, no Soft-NMS settings were found reaching higher performance compared to base NMS (only settings reaching equal performance).

Finally, in the last two rows of Table~\ref{tab:ch4_dup_v2}, we also provide results for the UpBnd-50 and UpBnd-75 variants, which are duplicate removal mechanisms with an upper-bound performance for the AP$_{50}$ and AP$_{75}$ metrics respectively (obtained by using ground-truth information). The UpBnd duplicate removal mechanism consists of removing detections which (1) do not sufficiently overlap (\eg $0.50$ and $0.75$) with any unmatched ground-truth detection, and (2) do have an IoU overlap of at least $0.5$ with a non-removed higher-scoring detection. From Table~\ref{tab:ch4_dup_v2}, we can see from the UpBnd variants that there are still some potential improvements left \wrt the duplicate removal mechanisms. However, it is unclear how much of this potential can realistically be acquired in practice.

\begin{table*}[t!]
    \centering
    \caption[FQDetV2 design - FQDet vs. FQDetV2]{Comparison between FQDet and FQDetV2 heads.}
    \vspace{0.2cm}
    \resizebox{\columnwidth}{!}{
        \begin{tabular}{c|c c c c c c|c c c}
            \toprule
            Head & AP & AP$_{50}$ & AP$_{75}$ & AP$_S$ & AP$_M$ & AP$_L$ & Params & FLOPs & FPS \\ \midrule
            FQDet & $43.0$ & $61.8$ & $46.1$ & $24.9$ & $46.8$ & $58.5$ & $33.9$ M & $99.0$ G & $21.6$ \\
            \textbf{FQDetV2} & $47.0$ & $64.3$ & $51.7$ & $30.8$ & $51.4$ & $62.0$ & $37.9$ M & $117.4$ G & $17.7$ \\
            \bottomrule
        \end{tabular}}
    \label{tab:ch4_FQDet_v2}
\end{table*}

\paragraph{FQDetV2.} Finally, when we combine all of the above design improvements, we obtain the \gls{FQDetV2} head. In Table~\ref{tab:ch4_FQDet_v2}, we compare the baseline FQDet head with the FQDetV2 head. We can see that the FQDetV2 head reaches $47.0$~AP, outperforming the FQDet baseline by a sizable $4.0$~AP. Especially for the AP$_{75}$, AP$_S$ and AP$_M$ metrics, we observe large improvements. These FQDetV2 performance improvements come with a higher but reasonable computational cost, increasing the number of parameters with $4.0$~M, increasing the FLOPs with $18.4$~G, and decreasing the FPS from $21.6$ to $17.7$.

\subsection{FQDetV2 main experiments} \label{sec:ch4_main_v2}

\begin{table*}[t!]
    \centering
    \caption[FQDetV2 main - FQDetV2 vs. SOTA]{Comparison between FQDetV2 and SOTA object detection heads.}
    \vspace{0.2cm}
    \resizebox{\columnwidth}{!}{
    \begin{tabular}{c c c c|c c c c c c|c}
        \toprule
        Bbone & Neck & Head & Ep. & AP & AP$_{50}$ & AP$_{75}$ & AP$_S$ & AP$_M$ & AP$_L$ & Params \\
        \midrule
        R50 & DefEnc-P3 & DINO~\cite{zhang2022dino} & $12$ & $49.0$ & $66.6$ & $53.5$ & $32.0$ & $52.3$ & $63.0$ & $47$ M \\
        R50 & DefEnc-P3 & Stable-DINO~\cite{liu2023detection} & $12$ & $50.4$ & $67.4$ & $55.0$ & $32.9$ & $54.0$ & $65.5$ & $-$ M \\
        R50 & DefEnc-P3 & \textbf{FQDetV2} & $12$ & $50.8$ & $67.2$ & $56.3$ & $33.6$ & $54.7$ & $65.5$ & $48$ M \\
        \midrule
        R50 & DefEnc-P2 & DINO~\cite{zhang2022dino} & $12$ & $49.4$ & $66.9$ & $53.8$ & $32.3$ & $52.5$ & $63.9$ & $47$ M \\
        R50 & DefEnc-P2 & Stable-DINO~\cite{liu2023detection} & $12$ & $50.5$ & $66.8$ & $55.3$ & $32.6$ & $54.0$ & $65.3$ & $-$ M \\
        R50 & DefEnc-P2 & \textbf{FQDetV2} & $12$ & $51.7$ & $67.7$ & $57.1$ & $34.8$ & $55.8$ & $65.8$ & $48$ M \\
        R50 & DefEnc-P2 & Co-DETR~\cite{zong2022detrs} & $12$ & $52.1$ & $69.4$ & $57.1$ & $35.4$ & $55.4$ & $65.9$ & $-$ M \\
        \midrule
        \midrule
        R50 & DefEnc-P3 & DINO~\cite{zhang2022dino} & $24$ & $50.4$ & $68.3$ & $54.8$ & $33.3$ & $53.7$ & $64.8$ & $47$ M \\
        R50 & DefEnc-P3 & Stable-DINO~\cite{liu2023detection} & $24$ & $51.5$ & $68.5$ & $56.3$ & $35.2$ & $54.7$ & $66.5$ & $-$ M \\
        R50 & DefEnc-P3 & \textbf{FQDetV2} & $24$ & $52.1$ & $68.4$ & $57.4$ & $35.5$ & $55.8$ & $66.9$ & $48$ M \\
        \midrule
        R50 & DefEnc-P2 & DINO~\cite{zhang2022dino} & $24$ & $51.3$ & $69.1$ & $56.0$ & $34.5$ & $54.2$ & $65.8$ & $47$ M \\
        R50 & DefEnc-P2 & \textbf{FQDetV2} & $24$ & $52.8$ & $69.0$ & $58.2$ & $36.9$ & $56.1$ & $66.8$ & $48$ M \\
        \midrule
        \midrule
        SwL & DefEnc-P3 & DINO~\cite{zhang2022dino} & $12$ & $56.8$ & $75.6$ & $62.0$ & $40.0$ & $60.5$ & $73.2$ & $-$ M \\
        SwL & DefEnc-P3 & Stable-DINO~\cite{liu2023detection} & $12$ & $57.7$ & $75.7$ & $63.4$ & $39.8$ & $62.0$ & $74.7$ & $-$ M \\
        SwL & DefEnc-P3 & \textbf{FQDetV2} & $12$ & $58.2$ & $75.4$ & $64.6$ & $41.2$ & $63.2$ & $74.7$ & $219$ M \\
        \midrule
        SwL & DefEnc-P3 & Stable-DINO~\cite{liu2023detection} & $24$ & $58.6$ & $76.7$ & $64.1$ & $41.8$ & $63.0$ & $74.7$ & $-$ M \\
        SwL & DefEnc-P3 & \textbf{FQDetV2} & $24$ & $58.7$ & $75.9$ & $65.0$ & $41.7$ & $63.5$ & $74.9$ & $219$ M \\
        \bottomrule
    \end{tabular}}
    \label{tab:ch4_FQDetV2_main}
\end{table*}

\paragraph{Performance.} In Table~\ref{tab:ch4_FQDetV2_main}, we provide the main experiment results comparing the FQDetV2 head with the \gls{SOTA} object detection heads DINO~\cite{zhang2022dino}, Stable-DINO~\cite{liu2023detection}, and Co-DETR~\cite{zong2022detrs}. Most notably, after training for $24$~epcohs, the FQDetV2 head obtains $52.8$~AP when using the ResNet-50 backbone~\cite{he2016deep} with the DefEnc-P2 neck~\cite{zhu2020deformable}, and $58.7$~AP when using the Swin-L backbone~\cite{liu2021swin} with the DefEnc-P3 neck.

The FQDetV2 head outperforms the DINO and Stable-DINO heads, both when using the ResNet-50 and Swin-L~\cite{liu2021swin} backbones. The DINO heads are popular query-based two-stage object detection heads that are built from the DETR line of work~\cite{carion2020end, zhu2020deformable, liu2022dab, li2022dn}. Different from the FQDetV2 head, the DINO heads use Hungarian matching, classification and box regression networks after each decoder layer supervised by auxiliary losses, intermediate bounding box regression (\gls{IBBR}), and query denoising~\cite{li2022dn}. Many of these techniques such as query denoising are meant to improve the inductive biases and speed up the convergence. In the FQDet heads, these challenges are met by introducing anchors of various sizes and aspect ratios in the query-based object detection paradigm, which we believe to be a simpler solution. Moreover, note that the DINO heads in fact also use some kind of anchoring mechanism, but only with anchors of a single size and aspect ratio. All in all, the FQDetV2 head proposes a nice alternative to the DINO heads (currently dominating the high-performance object detection field) using a different approach, while achieving similar or better performance.

We also compare the FQDetV2 head with the current SOTA object detection head Co-DETR. The Co-DETR head is in fact the DINO head, trained in conjunction with the ATSS~\cite{zhang2020bridging} and Faster R-CNN~\cite{ren2015faster} heads. These auxiliary heads provide additional supervision to the backbone, neck, and decoder layers resulting in faster convergence and increased performance. The Co-DETR head outperforms the FQDetV2 head with $0.4$~AP. Especially on the AP$_{50}$ metric, the Co-DETR head performs significantly better with $69.4$~AP$_{50}$ compared to $67.7$~AP$_{50}$ for the FQDetV2 head. Unfortunately, we only have a single point of reference to compare with Co-DETR, as most of its results are obtained using the large-scale jittering (LSJ) data augmentation scheme~\cite{ghiasi2021simple}.

\begin{table*}[t!]
    \centering
    \caption[FQDetV2 main - FQDetV2 cost metrics]{Computational cost metrics of the FQDetV2 models.}
    \vspace{0.2cm}
    \resizebox{0.95\columnwidth}{!}{
    \begin{tabular}{c c c c|c|c c c}
        \toprule
        Bbone & Neck & Head & Ep. & AP & Params & FLOPs & FPS \\
        \midrule
        R50 & DefEnc-P3 & FQDetV2 & $24$ & $52.1$ & $48.1$ M & $256.1$ G & $15.5$ \\
        R50 & DefEnc-P2 & FQDetV2 & $24$ & $52.8$ & $48.4$ M & $747.0$ G & $6.8$ \\
        SwL & DefEnc-P3 & FQDetV2 & $24$ & $58.7$ & $218.7$ M & $875.4$ G & $5.8$ \\
        \bottomrule
    \end{tabular}}
    \label{tab:ch4_FQDetV2_cost}
\end{table*}

\paragraph{Computational cost.} In Table~\ref{tab:ch4_FQDetV2_cost}, we provide the computational cost metrics of the different FQDetV2 models for completeness. Unfortunately, we cannot compare with the DINO, Stable-DINO, Co-DETR heads, as the computational cost metrics are not reported in their respective papers. 

From Table~\ref{tab:ch4_FQDetV2_cost}, we can see that the cost metrics are quite different for the three models. Out of the three, the FQDetV2 model with the ResNet-50 backbone and the DefEnc-P3 neck has the lowest cost, with $48.1$~M parameters, $256.1$~G FLOPs, and $15.5$~FPS. When replacing the DefEnc-P3 neck with the DefEnc-P2 neck, we can see that FLOPs and FPS costs increase by a substantial amount, while the number of parameters remain roughly the same. The addition of the high-resolution $P_2$ feature map to the feature pyramid hence has a big impact on the computational cost. With a performance gain of only $0.7$~AP, we do not believe this increase is worth the added computational cost.

Finally, we compare the FQDetV2 models with the ResNet-50 and Swin-L backbones, both equipped with the DefEnc-P3 neck. We can see that the Swin-L backbone significantly increases the computational cost, especially for the number of parameters metric increasing from $48.1$~M to $218.7$~M. However, as opposed to the model with the DefEnc-P2 neck, the added computation results in a significant performance gain from $52.1$~AP to $58.7$~AP. Replacing the ResNet-50 backbone with the Swin-L backbone is hence to be preferred over replacing the DefEnc-P3 neck with the DefEnc-P2 neck, by reaching a better performance vs. cost trade-off.

\subsection{FQDetV2 qualitative results} \label{sec:ch4_qualitative_v2}

In what follows, we provide some qualitative results of the FQDetV2 model with the ResNet-50 backbone and the FPN neck from Subsection~\ref{sec:ch4_design_v2}.

\begin{figure}
    \centering
    \includegraphics[scale=0.27]{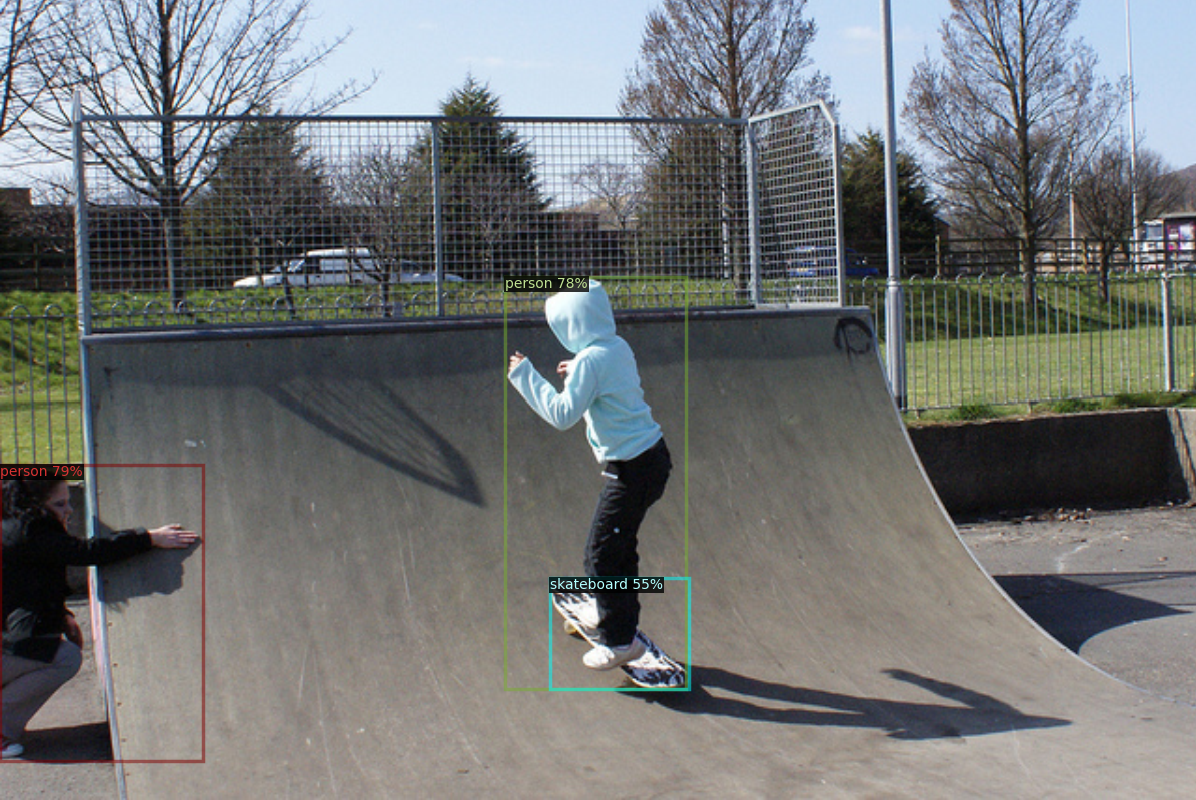} \\
    \vspace{0.4cm}
    \includegraphics[scale=0.302]{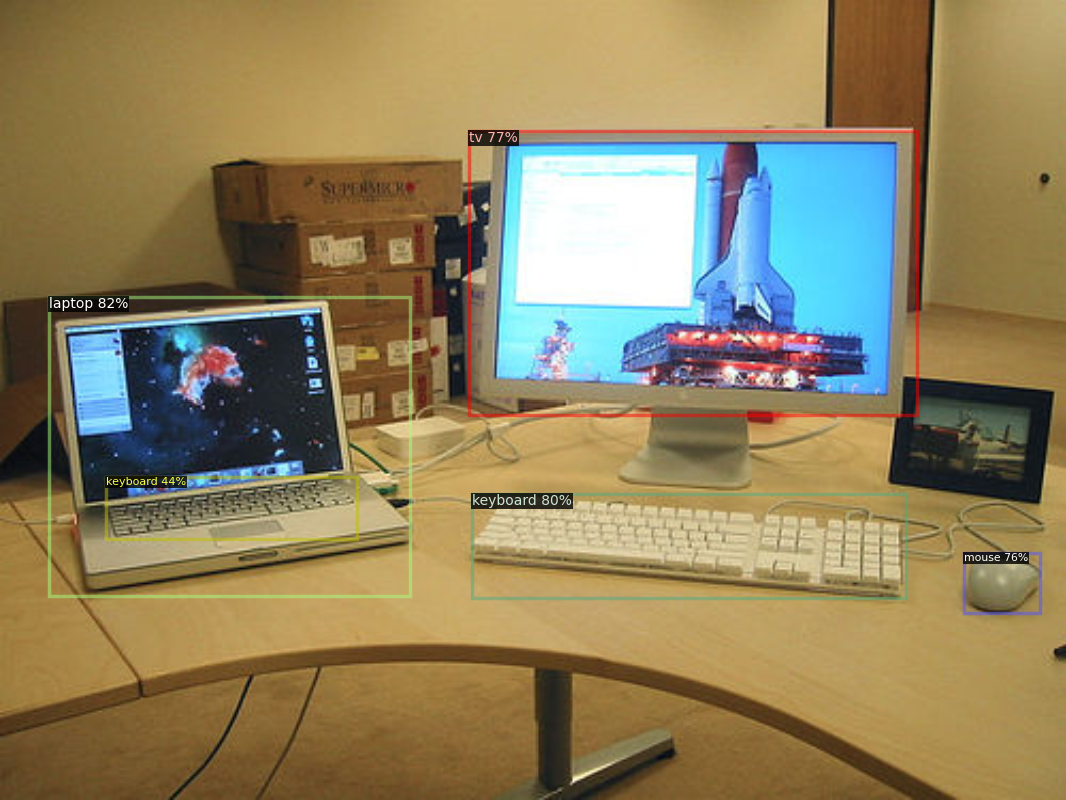}
    \caption[FQDetV2 qualitative - Correct detections]{Example images with correct detections.}
    \label{fig:ch4_correct_dets}
\end{figure}

\begin{figure}
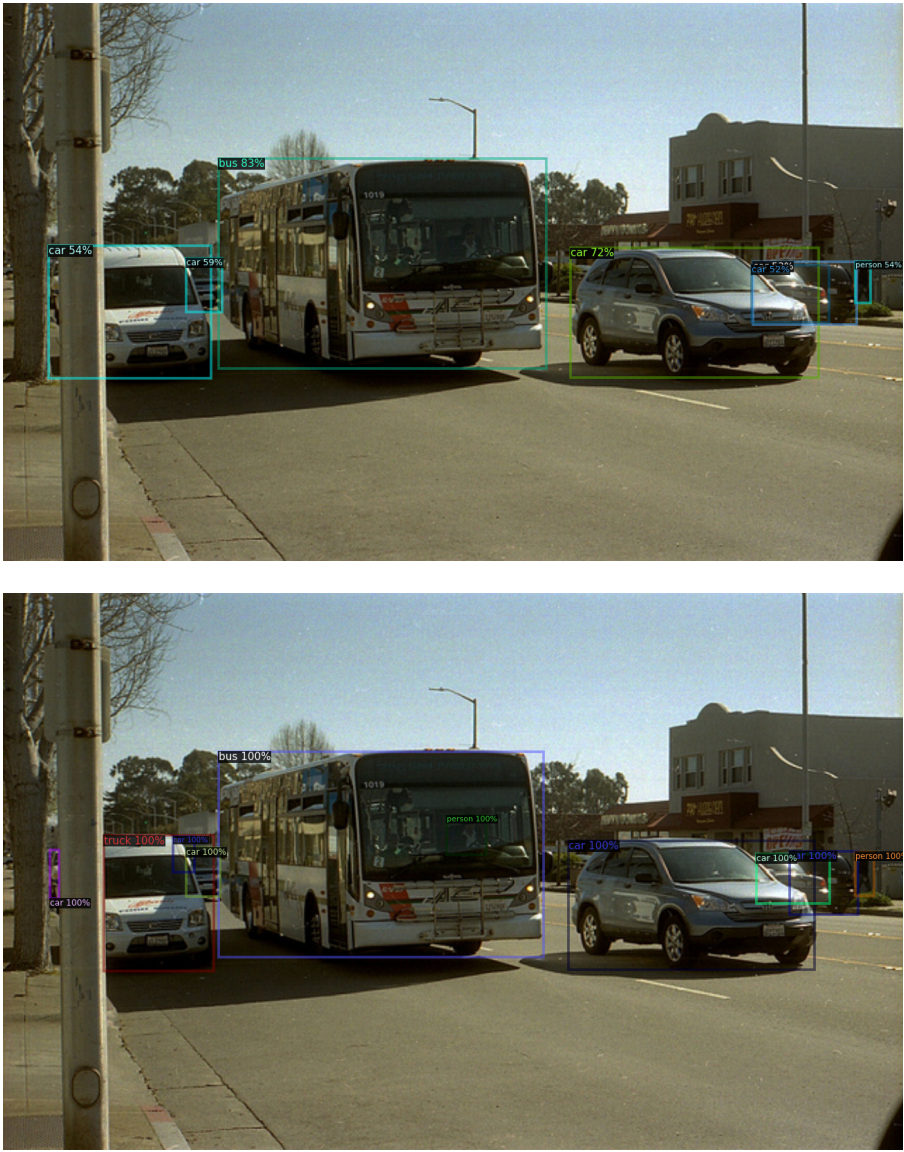

    \centering
    \includegraphics[scale=0.26]{incorrect_cls_pred} \\
    \vspace{0.4cm}
    \includegraphics[scale=0.26]{incorrect_cls_tgt}
    \caption[FQDetV2 qualitative - Incorrect classification]{Example image with incorrect classification. The top image contains the predicted detections and the bottom image the ground-truth detections.}
    \label{fig:ch4_incorrect_cls}
\end{figure}

\begin{figure}
    \centering
    \includegraphics[scale=0.27]{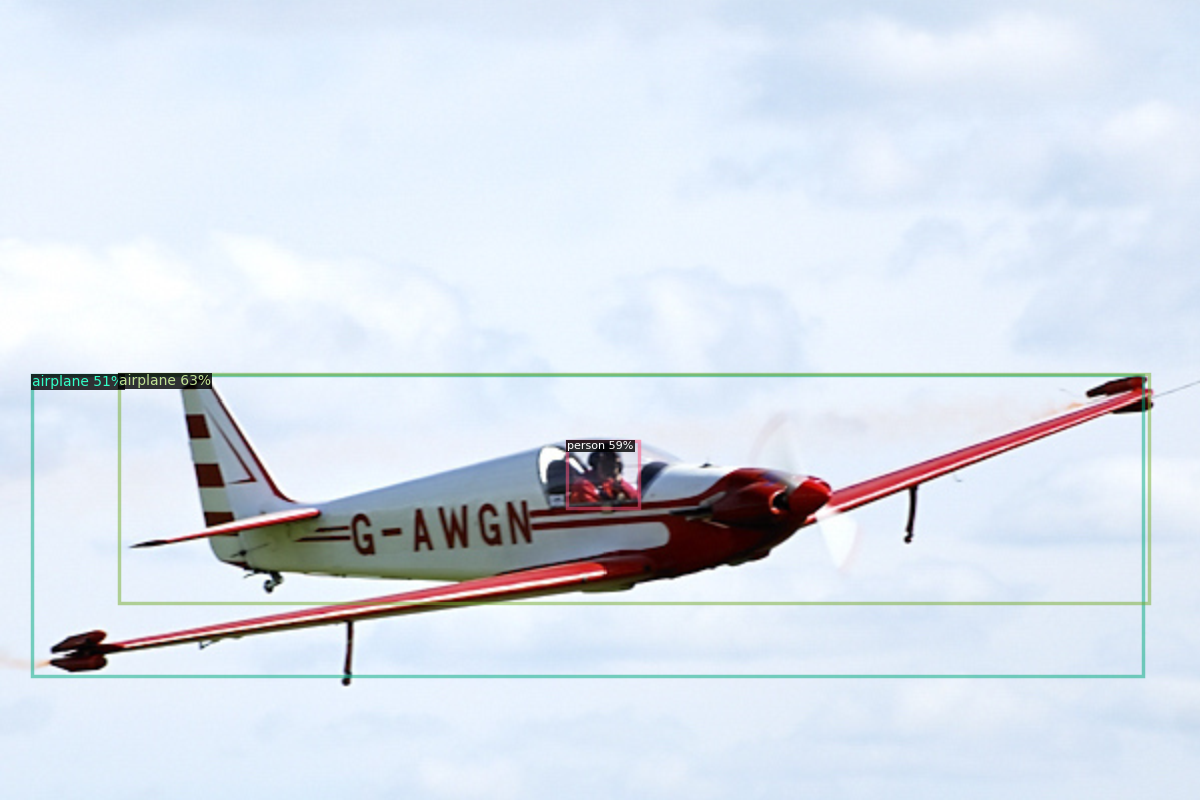} \\
    \vspace{0.4cm}
    \includegraphics[scale=0.27]{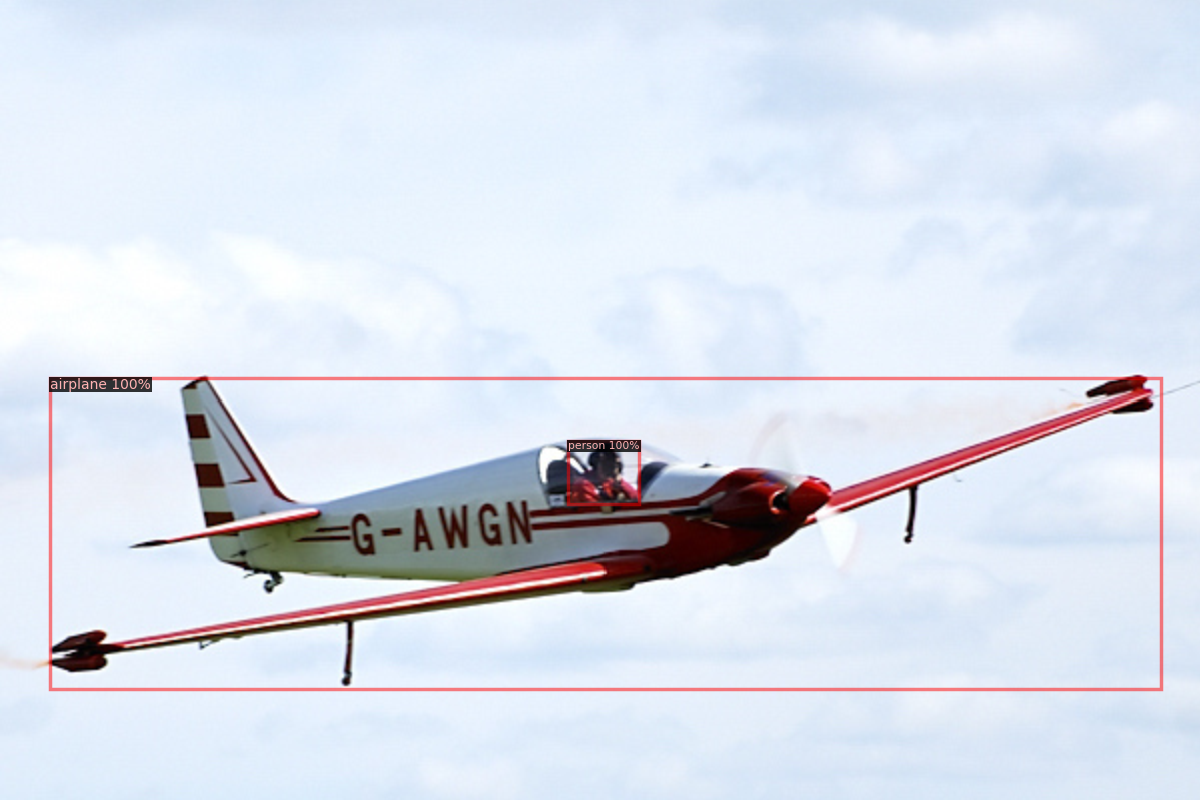}
    \caption[FQDetV2 qualitative - Incorrect bonding box]{Example image with incorrect bounding box. The top image contains the predicted detections and the bottom image the ground-truth detections.}
    \label{fig:ch4_incorrect_box}
\end{figure}

\begin{figure}
    \centering
    \includegraphics[scale=0.27]{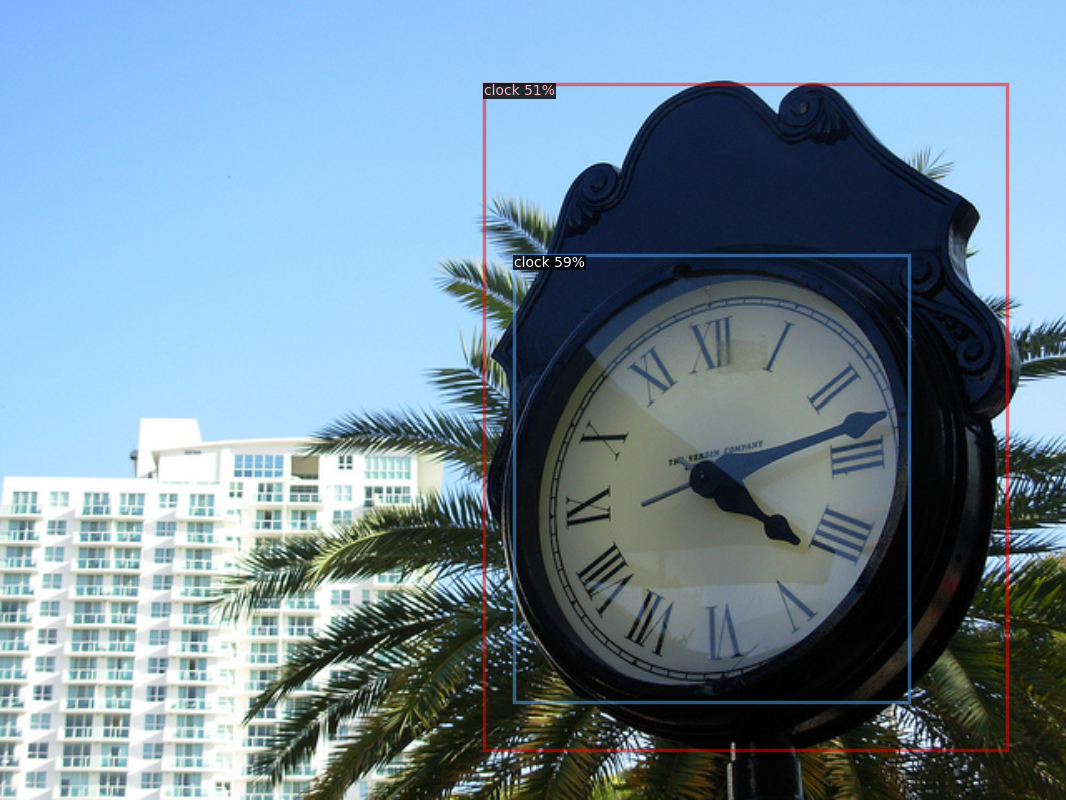} \\
    \vspace{0.4cm}
    \includegraphics[scale=0.27]{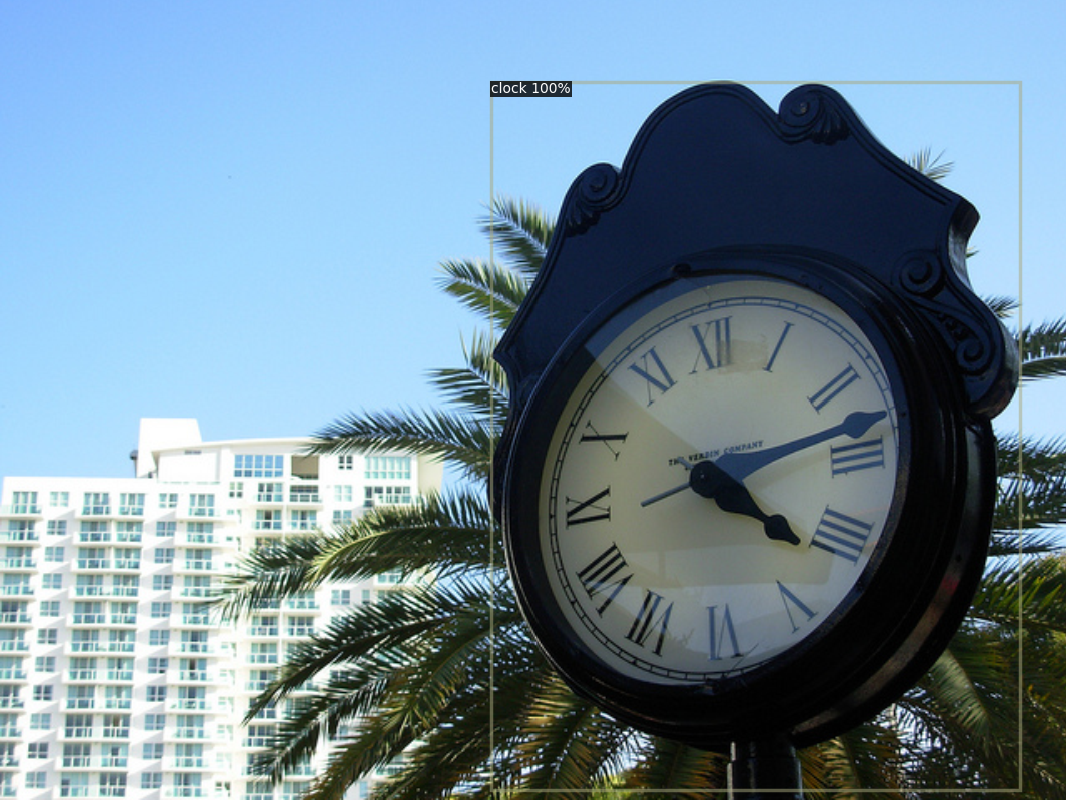}
    \caption[FQDetV2 qualitative - Ambiguous box]{Example image with ambiguous box. The top image contains the predicted detection and the bottom image the ground-truth detection.}
    \label{fig:ch4_ambiguous_box}
\end{figure}

\paragraph{Correct detections.} In Figure~\ref{fig:ch4_correct_dets}, we provide two example images with correct detections. We can see that the correct classification labels are predicted and that the bounding boxes are tight around the objects of interest.

\paragraph{Incorrect classification.} We give an example image containing an incorrect classification in Figure~\ref{fig:ch4_incorrect_cls}. Here, the vehicle on the left is classified as a car, but should have been predicted as a truck instead. Moreover, we also observe that some bounding boxes are slightly imprecise, such as the bounding box of the truck on the left and the bounding box of the car on the far right.

\paragraph{Incorrect box.} In Figure~\ref{fig:ch4_incorrect_box}, we provide an example image with an incorrect bounding box prediction. Here, we can see that the box of the highest-scoring detection misses a large part of the plane, by neglecting the right wing. However, note how the box of a lower-scoring detection does a much better job, covering the whole plane. If the box of the highest-scoring detection does not sufficiently overlap with the corresponding ground-truth box, then the object is still found by the lower-scoring detection, given its much more accurate bounding box. This shows that removing duplicate detections is not always beneficial for the AP metric, which is one of the reasons why it so difficult to do better than the hand-crafted NMS duplicate removal mechanism.

\paragraph{Ambiguous box.} We give an example image containing an ambiguous box in Figure~\ref{fig:ch4_ambiguous_box}. We can see that it is unclear whether the outer part is considered as belonging to the clock or not. The ground-truth considers the whole structure as part of the clock, while the highest-scoring prediction only considers the inner part. Fortunately, a lower-scoring prediction also contains the outer part, similar to the ground-truth annotation. Hence, we again see that in some cases it pays off to keep duplicate detections alive, as the boxes of some objects might be ambiguous.

\section{Limitations and future work} \label{sec:ch4_limit_future}

In following paragraphs, we discuss some of the limitations and give directions for future work.

\paragraph{Dataset overfitting.} All of our experiments were performed on the COCO object detection dataset~\cite{lin2014microsoft}. Because of this, some of the found hyperparameters might be specifically tuned towards this dataset, and therefore might not generalize to other object detection datasets. This for example could be the case for the hyperparameters controlling the number of queries and the NMS threshold, which depend on how many objects are expected in the image and to what extent these objects overlap. As future work, it would therefore be useful to perform experiments on other datasets such as ADE20k~\cite{zhou2017scene}, Cityscapes~\cite{cordts2016cityscapes} and Objects365~\cite{shao2019objects365}, to investigate which FQDet and FQDetV2 hyperparameters are sensitive to changes in data distribution.

\paragraph{Metric overfitting.} All of our models were evaluated using the COCO AP metric. Based on this AP metric, design decisions were made, eventually culminating in the FQDet and FQDetV2 heads. However, a different performance metric might have resulted in different choices, leading to different heads which are not necessarily worse from a practical point of view.

For example, some of the elements added to the FQDetV2 head are especially geared towards improving the AP metric. It is namely important for the AP metric to make sure that the most accurate detections come first, where slight inaccuracies in the ranking of detections leads to noticeable drops in AP performance. To mitigate this, the FQDetV2 head replaces the sigmoid focal loss~\cite{lin2017focal} with the quality focal loss~\cite{li2020generalized}, and adds the box scoring network estimating bounding box accuracies. While these techniques improve the AP performance, it is questionable whether these changes truly result in better object detection methods. It would therefore be useful to design an object detection metric which is less sensitive to the precise ranking of detections (\ie less sensitive to the detection scores).

Additionally, the AP metric does not penalize false positive detections, as long as they have lower detection scores compared to true positive detections. Because of this, the object detection heads are not required to filter out false positives, simplifying the task of the head. However, the removal of false positives is often needed in practical applications, and it would therefore be useful to introduce an object detection metric that penalizes false positives, even if they have lower detection scores compared to true positives. Of course, this requires all objects to be labeled within images, which is not the case for all datasets.

\paragraph{Duplicate removal.} In Table~\ref{tab:ch4_dup_v2} of Subsection~\ref{sec:ch4_design_v2}, we saw that the NMS duplicate removal mechanism works best, but that the performance of the learned variant is close. We believe that additional research into learning which detections are duplicates, should further close this gap, and eventually should outperform the NMS mechanism. Note that the SOTA heads Stable-DINO~\cite{liu2023detection} and Co-DETR~\cite{zong2022detrs} which use Hungarian matching to avoid duplicate detections, also still report improvements of $0.2$~AP when applying NMS.

\paragraph{Additional improvements.} Following three techniques could be added to the FQDet heads, possibly further increasing the performance. Firstly, some dynamic matching ideas from~\cite{ge2021ota, ge2021yolox, lyu2022rtmdet} could be used to adapt the current static top-k matching scheme. Secondly, the query denoising technique from~\cite{li2022dn} could be used to increase the convergence speed even more, and possibly also increase performance on objects which are hard to find. Finally, the sigmoid focal loss~\cite{lin2017focal} used in Stage~1 could be replaced by a different loss such as the DICE loss~\cite{milletari2016v} or Seesaw loss~\cite{wang2021seesaw}, which are losses specifically designed to deal with large imbalances between the number of positives and negatives. 

\section{Conclusion}
We propose the FQDet and FQDetV2 heads, new query-based two-stage object detection heads combining the best of both classical and DETR-based detectors. Our FQDet heads improve the cross-attention prior with anchors and introduce the effective top-k matching scheme. We perform various ablation studies and design experiments validating the FQDet design choices and show improvements \wrt existing two-stage heads when compared on the COCO object detection benchmark after training for $12$ and $24$ epochs.

\section*{Acknowledgements}
This work was partly supported by the Ford-KU Leuven alliance program.

\cleardoublepage
  \graphicspath{{\chaptersdir/segmentation/image/}}%
  \chapter[EffSeg: Efficient Fine-Grained Segmentation]{EffSeg: Efficient Fine-Grained Segmentation using Structure-Preserving Sparsity} \label{ch:segmentation}

\definecolor{shadecolor}{gray}{0.85}
\begin{shaded}
    The work in this chapter was adapted from following preprint: \\ \vspace{-0.1cm} \\
    \bibentry{picron2023effseg}
\end{shaded}

\section{Introduction}

Instance segmentation is a fundamental computer vision task assigning a semantic category (or background) to each image pixel, while differentiating between instances of a same category. Many high-performing instance segmentation methods \cite{he2017mask, cai2019cascade, chen2019hybrid, kirillov2020pointrend, vu2021scnet, zhang2021refinemask} follow the two-stage paradigm. This paradigm consists in first predicting an axis-aligned bounding box called Region of Interest (\gls{RoI}) for each detected instance, and then segmenting each pixel within the RoI as belonging to the detected instance or not.

Most two-stage instance segmentation heads \cite{he2017mask, cai2019cascade, chen2019hybrid, vu2021scnet} predict a $28\times28$ mask (within the RoI) per instance, which is too coarse to capture the fine-grained details of many objects. PointRend~\cite{kirillov2020pointrend} and RefineMask~\cite{zhang2021refinemask} both address this issue by predicting a $112\times112$ mask instead, resulting in higher quality segmentations. In both methods, these $112\times112$ masks are obtained by using a multi-stage refinement procedure, first predicting a coarse mask and then iteratively upsampling this mask by a factor~$2$ while overwriting the predictions in uncertain (PointRend) or boundary (RefineMask) locations. However, both methods have some limitations.

PointRend~\cite{kirillov2020pointrend} on the one hand overwrites predictions by sampling coarse-fine feature pairs from the most uncertain locations and by processing these pairs \textit{individually} using an MLP. Despite only performing computation at the desired locations and hence being efficient, PointRend is unable to access information from neighboring features during the refinement process, resulting in sub-optimal segmentation performance.

RefineMask~\cite{zhang2021refinemask} on the other hand processes dense feature maps and obtains new predictions in all locations, though only uses these predictions to overwrite in the boundary locations of the current prediction mask. Operating on dense feature maps enables RefineMask to use 2D convolutions allowing information to be exchanged between neighboring features, which results in improved segmentation performance \wrt PointRend. However, this also means that all computation is performed on all spatial locations within the RoI at all times, which is computationally inefficient.

\begin{table}
    \centering
    \caption[EffSeg - Comparison of fine-grained segmentation methods]{Comparison between fine-grained segmentation methods.}
    \vspace{0.2cm}
    \resizebox{0.75\columnwidth}{!}{
    \begin{tabular}{c c c}
        \toprule
        \multirow{3}{*}{Head} & Computation at  & Access to \\
        & sparse locations & neighboring features \\
        & (\ie efficient) & (\ie good performance) \\
        \midrule
        PointRend~\cite{kirillov2020pointrend} & \cmark & \xmark \\
        RefineMask~\cite{zhang2021refinemask} & \xmark & \cmark \\
        EffSeg (ours) & \cmark & \cmark \\
        \bottomrule
    \end{tabular}}
    \label{tab:ch5_fine_grained}
\end{table}

In this work, we propose EffSeg which combines the strengths and eliminates the weaknesses of PointRend and RefineMask by only performing computation at the desired locations, while still being able to access features of neighboring locations (see Table~\ref{tab:ch5_fine_grained}). This is challenging, as it requires a mechanism to perform sparse computations efficiently. For dense computations as in RefineMask, highly optimized dense convolutions can be used. Likewise, the $1\times1$ convolutions in PointRend can easily be computed after a simple data reorganization. But what about non $1\times1$ convolution filters that need to be computed only for a sparse set of pixels or locations? 

For this, we introduce our Structure Preserving Sparsity method (\gls{SPS}). SPS separately stores the active features (\ie the features in spatial locations requiring new predictions), the passive features (\ie the non-active features) and a dense 2D index map. More specifically, the active and passive features are stored in $N_A \times F$ and $N_P \times F$ matrices respectively, with $N_A$ the number of active features, $N_P$ the number of passive features, and $F$ the feature size. The index map stores the feature indices (as opposed to the features themselves) in a 2D map, preserving information about the 2D spatial structure between the different features in a compact way. This allows SPS to have access to neighboring features such that any 2D operation can still be performed at the sparse locations of interest. In EffSeg, we hence combine the desirable properties of both PointRend and RefineMask by introducing our novel SPS method, which is a stand-alone method different from combining the PointRend and RefineMask methods. 

We evaluate EffSeg and its baselines on the COCO~\cite{lin2014microsoft} instance and panoptic segmentation benchmarks. Experiments show that EffSeg achieves similar segmentation performance compared to RefineMask (\ie the best-performing baseline), while reducing the number of FLOPs up to $70$\% and increasing the FPS up to $44$\%.

\section{Related work}

\paragraph{Instance segmentation.} Instance segmentation methods can be divided into two-stage (or box-based) methods and one-stage (or box-free) methods. Two-stage approaches \cite{he2017mask, cai2019cascade, chen2019hybrid, kirillov2020pointrend, zhang2021refinemask} first predict an axis-aligned bounding box called Region of Interest (\gls{RoI}) for each detected instance and subsequently categorize each pixel as belonging to the detected instance or not. One-stage approaches \cite{tian2020conditional, wang2020solov2, zhang2021k, cheng2022masked} on the other hand directly predict instance masks over the whole image without using intermediate bounding boxes.

One-stage approaches have the advantage that they are similar to semantic segmentation methods by predicting masks over the whole image instead of inside the RoI, allowing for a natural extension to the more general panoptic segmentation task \cite{kirillov2019panoptic}. Two-stage approaches have the advantage that by only segmenting inside the RoI, there is no wasted computation outside the bounding box. As EffSeg aims to only perform computation there where it is needed, the two-stage approach is chosen.

\paragraph{Fine-grained instance segmentation.} Many two-stage instance segmentation methods such as Mask R-CNN~\cite{he2017mask} predict rather coarse segmentation masks. There are two main reasons why the predicted masks are coarse. First, segmentation masks of large objects are computed using features pooled from low resolution feature maps. A first improvement found in many methods \cite{kirillov2020pointrend, cheng2020boundary, zhang2021refinemask, ke2022mask}, consists in additionally using features from the high-resolution feature maps of the feature pyramid. Second, Mask R-CNN only predicts a $28\times28$ segmentation mask inside each RoI, which is too coarse to capture the fine details of many objects. Methods such as PointRend~\cite{kirillov2020pointrend}, RefineMask~\cite{zhang2021refinemask} and Mask Transfiner~\cite{ke2022mask} therefore instead predict a $112\times112$ mask within each RoI, allowing for fine-grained segmentation predictions. PointRend achieves this by using an \gls{MLP}, RefineMask by iteratively using their SFM module consisting of parallel convolutions with different dilations, and Mask Transfiner by using a transformer. However, all of these methods have limitations. PointRend has no access to neighboring features, RefineMask performs computation at all locations within each RoI at all times, and Mask Transfiner performs attention over \textit{all} active features instead of over neighboring features only and it does not have access to passive features. EffSeg instead performs local computation at sparse locations, while keeping access to both active and passive features.

Another family of methods obtaining fine-grained segmentation masks, are contour-based methods \cite{peng2020deep, liu2021dance, zhu2022sharpcontour}. Contour-based methods first fit a polygon around an initial mask prediction, and then iteratively update the polygon vertices to improve the segmentation mask. Contour-based methods can hence be seen as a post-processing method to improve the quality of the initial mask. Contour-based methods obtain good improvements in mask quality when the initial mask is rather coarse~\cite{zhu2022sharpcontour} (\eg a mask predicted by Mask R-CNN~\cite{he2017mask}), but improvements are limited when the initial mask is already of high-quality~\cite{zhu2022sharpcontour} (\eg a mask predicted by RefineMask~\cite{zhang2021refinemask}).

\paragraph{Panoptic Segmentation.} The panoptic segmentation task is introduced in~\cite{kirillov2019panoptic} in an attempt to unify the semantic segmentation and the instance segmentation tasks. Originally, many methods such as~\cite{kirillov2019panoptic2, xiong2019upsnet} relied on a separate semantic and instance segmentation branch (or head) in order to solve the panoptic segmentation task. However, more recent approaches such as \cite{zhang2021k, cheng2021per, cheng2022masked, li2022mask} instead use a one-stage transformer decoder in order to obtain the panoptic segmentation masks. Different from existing methods, we tackle the panoptic segmentation task by using the same two-stage RoI-based approach as for the instance segmentation task.

\paragraph{Spatial-wise dynamic networks.} In order to be efficient, EffSeg only performs processing at those spatial locations that are needed to obtain a fine-grained segmentation mask, avoiding unnecessary computation in the bulk of the object. EffSeg could hence be considered as a spatial-wise dynamic network. Spatial-wise dynamic networks have been used in many other computer vision tasks such as image classification~\cite{verelst2020dynamic}, object detection~\cite{yang2022querydet}, and video recognition~\cite{wang2022adafocus}. However, these methods differ from EffSeg, as they apply an operation at sparse locations on a dense tensor (see SparseOnDense method from Subsection~\ref{sec:ch5_sps}), whereas EffSeg uses the Structure-Preserving Sparsity (SPS) method separately storing the active features, the passive features, and a 2D index map containing the feature indices. This brings in two advantages: (1) passive features are not copied between subsequent sparse operations leading to increased storage efficiency, and (2) pointwise operations such as linear layers can directly be applied on the active features (instead of first having to select these from the 2D map) leading to increased processing speeds.

\section{EffSeg}

\begin{figure*}
    \centering
    \includegraphics[scale=0.6]{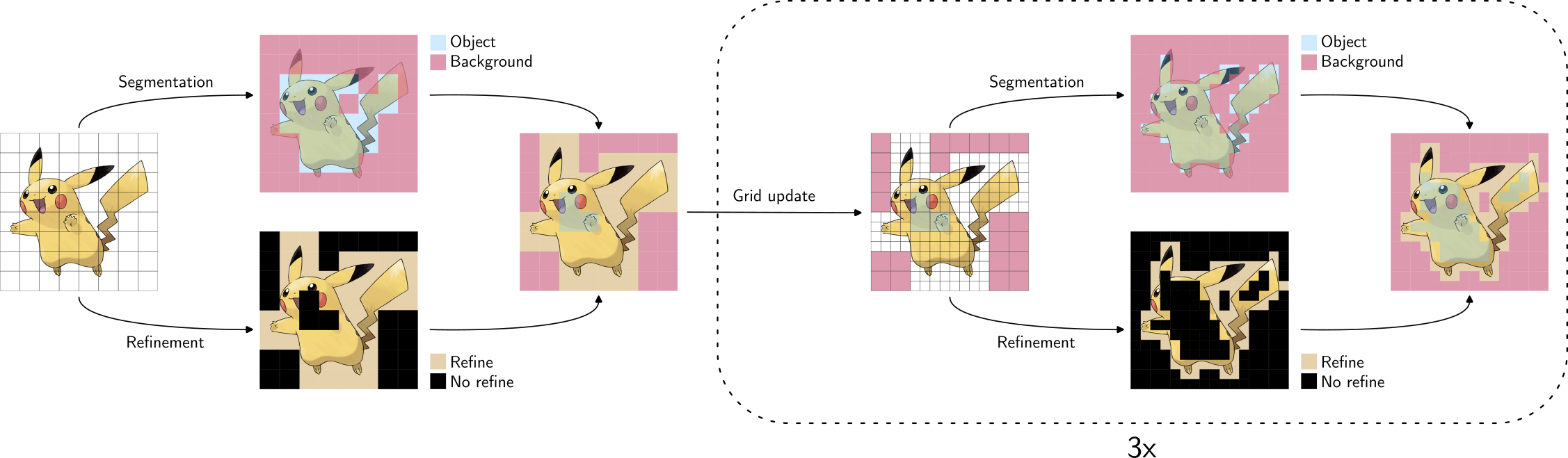}
    \caption[EffSeg - High-level overview]{High-level overview of how EffSeg efficiently predicts fine-grained segmentation masks. EffSeg consists of multiple stages refining the predicted segmentation masks, with each stage consisting of a segmentation prediction (top branch) and a refinement prediction (bottom branch). By only refining (\ie upsampling) at a select number of locations, EffSeg efficiently predicts fine-grained segmentations. In the example shown above, only $65$\% of the bounding box area needs to be processed in the second stage, and only roughly $35$\% and $20$\% in the third and fourth stages respectively.}
    \label{fig:ch5_pikachu}
\end{figure*}

\subsection{High-level overview}
EffSeg is a two-stage segmentation head obtaining fine-grained segmentation masks by using a multi-stage refinement procedure similar to the one used in PointRend~\cite{kirillov2020pointrend} and RefineMask~\cite{zhang2021refinemask}. For each detected object, EffSeg first predicts a $14\times14$ mask within the RoI and iteratively upsamples this mask by a factor~$2$ to obtain a fine-grained $112\times112$ mask. See Figure~\ref{fig:ch5_pikachu} for an illustration of how EffSeg efficiently predicts fine-grained segmentation masks.

The $14\times14$ mask is computed by working on a dense 2D feature map of shape $[N_R, F_0, 14, 14]$, with $N_R$ the number of RoIs and $F_0$ the feature size at refinement stage~$0$. However, the $14\times14$ mask is too coarse to obtain accurate segmentation masks, as a single cell from the $14\times14$ grid might contain both object and background pixels, rendering a correct assignment impossible. To solve this issue, higher resolution masks are needed, reducing the fraction of ambiguous cells which contain both foreground and background.

The $14\times14$ feature grid is therefore upsampled to a $28\times28$ feature grid. After processing the $28\times28$ feature grid in sparse locations, new segmentation predictions are made in these sparse locations, and the mask predictions in the remaining locations are inferred from existing predictions (\ie those of the $14\times14$ mask). Features corresponding to the mask locations for which a new prediction is made, are called \textit{active} features, whereas features corresponding to the remaining mask locations, are called \textit{passive} features (as they are not updated). Given that new segmentation predictions are only required for a subset of spatial locations within the $28\times28$ grid, it is inefficient to use a dense feature map of shape $[N_R, F_1, 28, 28]$ to represent this $28\times28$ feature grid (as done in RefineMask~\cite{zhang2021refinemask}). Moreover, when upsampling by a factor~$2$, every grid cell gets subdivided in a $2\times2$ grid of smaller cells, with the feature from the parent cell copied to the $4$ children cells. The dense feature map of shape $[N_R, F_1, 28, 28]$ hence contains many duplicate features, which is a second source of inefficiency. EffSeg therefore introduces the Structure-Preserving Sparsity (SPS) method, which separately stores the active features, the passive features (without duplicates), and a 2D index map containing the feature indices (see Subsection~\ref{sec:ch5_sps} for more information).

EffSeg repeats this upsampling process two more times, resulting in the fine-grained $112\times112$ mask. Further upsampling the predicted mask is undesired, as $224\times224$ masks typically do not yield performance gains~\cite{kirillov2020pointrend, ke2022mask} while requiring additional computation. At last, the final segmentation mask is obtained by pasting the predicted $112\times112$ mask inside the corresponding RoI box using bilinear interpolation.

\subsection{Structure-preserving sparsity} \label{sec:ch5_sps}

\begin{table*}
    \centering
    \caption[EffSeg - Comparison of dense and various sparse methods]{Comparison between dense and various sparse methods.}
    \vspace{0.2cm}
    \resizebox{\columnwidth}{!}{
    \begin{tabular}{c c c c c c}
        \toprule
        \multirow{2}{*}{Method} & Example & Computationally & Access to & Supports any & Storage \\
        & where used & efficient & neighbors & 2D operation & efficient \\
        \midrule
        Dense & RefineMask \cite{zhang2021refinemask} & \xmark & \cmark & \cmark & \xmark \\
        Pointwise & PointRend \cite{kirillov2020pointrend} & \cmark & \xmark & \xmark & \cmark \\
        Neighbors & - & \cmark & \cmark & \xmark & \cmark \\
        SparseOnDense & - & \cmark & \cmark & \cmark & \xmark \\
        SPS & EffSeg & \cmark & \cmark & \cmark & \cmark \\
        \bottomrule
    \end{tabular}}
    \label{tab:ch5_dense_sparse}
\end{table*}

\paragraph{Motivation.} When upsampling a segmentation mask by a factor~$2$, new predictions are only required in a subset of spatial locations. The \textbf{Dense} method, which consists of processing dense 2D feature maps as done in RefineMask~\cite{zhang2021refinemask}, is inefficient as new predictions are computed over all spatial locations instead of only over the spatial locations of interest. A method capable of performing computation in a sparse set of 2D locations is therefore required. We distinguish following four sparse methods, where we first propose three baseline methods before introducing our SPS method.

First, the \textbf{Pointwise} method selects features from the desired spatial locations (called \textit{active} features) and only processes these using pointwise networks such as MLPs or FFNs~\cite{vaswani2017attention}, as done in PointRend~\cite{kirillov2020pointrend}. Given that the pointwise networks do not require access to neighboring features, there is no need to store passive features, nor information about the 2D spatial relationship between features, making this method simple and efficient. However, the features solely processed by pointwise networks miss context information, resulting in inferior segmentation performance as empirically shown in Subsection~\ref{sec:ch5_comp_proc}. The Pointwise method is hence simple and efficient, but does not perform that well.

Second, the \textbf{Neighbors} method consists in both storing the active features, as well as their $8$ neighboring features. This allows the active features to be processed by pointwise operations, as well as by non-dilated 2D convolutions with $3\times3$ kernel by accessing the neighboring features. The Neighbors method hence combines efficiency with access to the $8$ neighboring features, yielding improved segmentation performance \wrt the Pointwise method. However, this approach is limited in the 2D operations it can perform. The $8$ neighboring features for example do not suffice for 2D convolutions with kernels larger than $3\times3$ or dilated convolutions, nor do they suffice for 2D deformable convolutions which require features to be sampled from arbitrary locations. The Neighbors method hence lacks generality in the 2D operations it can perform.

Third, the \textbf{SparseOnDense} method consists in applying traditional operations such as 2D convolutions at sparse locations of a dense 2D feature map, as \eg done in \cite{verelst2020dynamic}. This method allows information to be exchanged between neighboring features (as opposed to the Pointwise method) and is compatible with any 2D operation (as opposed to the Neighbors method). Moreover, it is \textit{computationally efficient} as it only performs computation there where it is needed. However, the use of a dense 2D feature map of shape $[N_R, F, H, W]$ as data structure is \textit{storage inefficient}, given that only a subset of the dense 2D feature map gets updated each time, with unchanged features copied from one feature map to the other. Additionally, the dense 2D feature map also contains multiple duplicate features due to passive features covering multiple cells of the 2D grid, leading to a second source of storage inefficiency. Hence, while having good performance and while being computationally efficient, the SparseOnDense method is not storage efficient.

Fourth, the \textbf{Structure-Preserving Sparsity (SPS)} method stores a $N_A \times F$ matrix containing the active features, a $N_P \times F$ matrix containing the passive features (without duplicates), and a dense 2D index map of shape $[N_R, H, W]$ containing the feature indices. The goal of the index map is to \textit{preserve} the 2D spatial configuration or \textit{structure} of the features, such that any 2D operation can still be performed (as opposed to the Neighbors method). Separating the storage of active and passive features, enables SPS to update the active features without requiring to copy the unchanged passive features (as opposed to the SparseOnDense method). Moreover, by storing the active features in a dense $N_A \times F$ matrix, pointwise operations such linear layers can be applied without any data reorganization (as opposed to the SparseOnDense method), leading to increased processing speeds. The SPS method hence allows for fast and storage efficient sparse processing, while being computationally efficient and supporting any 2D operation thanks to the 2D index map.

An overview of the different methods with their properties is found in Table~\ref{tab:ch5_dense_sparse}. The SPS method will be used in EffSeg as it ticks all the boxes.

\begin{figure*}
    \centering
    \includegraphics[scale=0.72]{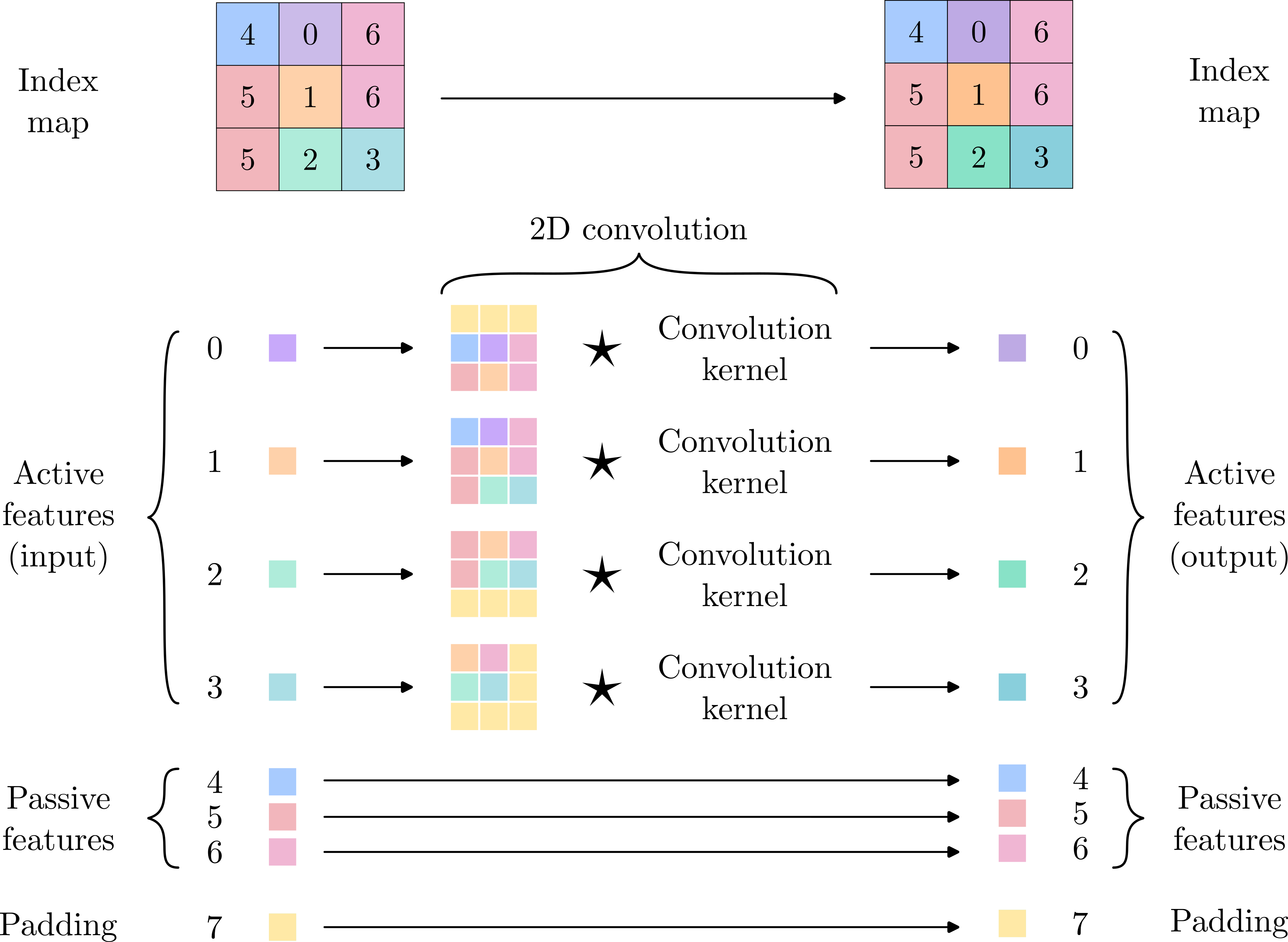}
    \caption[EffSeg - SPS toy example]{Toy example illustrating how a non-dilated 2D convolution operation with $3\times3$ kernel is performed using the SPS method. The colored squares represent the different feature vectors and the numbers correspond to the feature indices.}
    \label{fig:ch5_sps_conv}
\end{figure*}

\paragraph{Toy example of SPS.} In Figure~\ref{fig:ch5_sps_conv}, a toy example is shown illustrating how a non-dilated 2D convolution operation with $3\times3$ kernel is performed using the Structure-Preserving Sparsity (SPS) method. The example contains $4$~active features and $3$~passive features, organized in a $3\times3$ grid according to the dense 2D index map. Notice how the index map contains duplicate entries, with passive feature indices $5$ and~$6$ appearing twice in the grid.

The SPS method applies the 2D convolution operation with $3\times3$ kernel and dilation~$1$ to each of the active features, by first gathering its neighboring features into a $3\times3$ grid, and then convolving this feature grid with the learned $3\times3$ convolution kernel. When a certain neighbor feature does not exist as it lies outside of the 2D index map, a padding feature is used instead. In practice, this padding feature corresponds to the zero vector.

As a result, each of the active features are sparsely updated by the 2D convolution operation, whereas the passive features and the dense 2D index map remain unchanged. Note that performing other types of 2D operations such as dilated or deformable~\cite{dai2017deformable} convolutions occurs in similar way, with the only difference being which neighboring features are gathered and how they are processed.

\subsection{Detailed overview} \label{sec:ch5_detail}

\begin{figure*}
    \centering
    \includegraphics[scale=0.81]{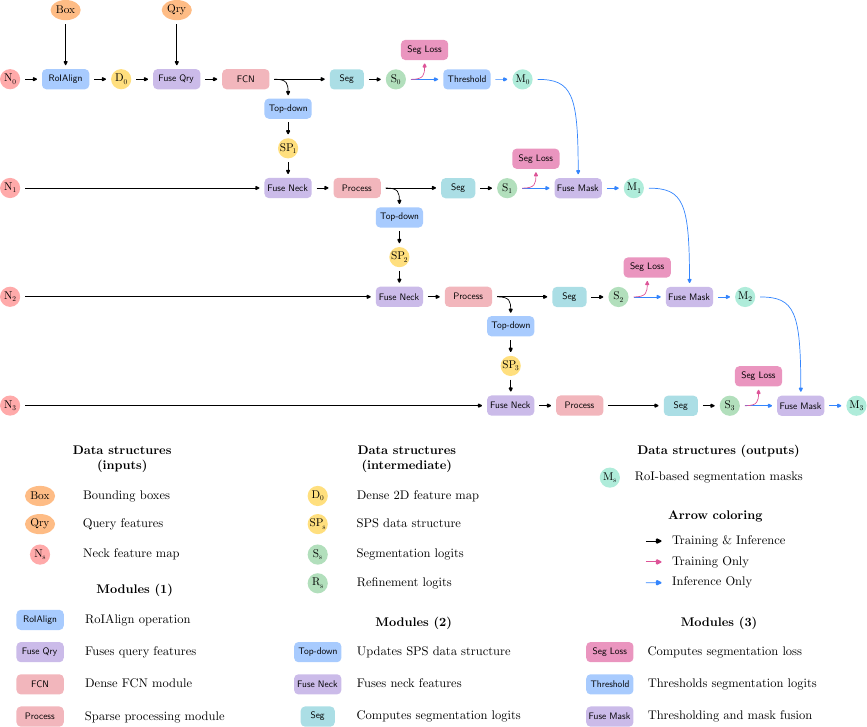}
    \caption[EffSeg - Detailed overview]{Detailed overview of the EffSeg architecture (the refinement branches and RoI mask pasting are omitted for clarity).}
    \label{fig:ch5_arch}
\end{figure*}

Figure~\ref{fig:ch5_arch} shows a detailed overview of the EffSeg architecture. The overall architecture is similar to the one used in RefineMask~\cite{zhang2021refinemask}, with some small tweaks as detailed below. In what follows, we provide more information about the various data structures and modules used in EffSeg.

\paragraph{Inputs.} The inputs of EffSeg are the neck feature maps, the predicted bounding boxes, and the query features. The neck feature maps are feature maps coming from the $P_2$-$P_7$ neck feature pyramid, with neck feature map~$N_s$ corresponding to refinement stage~$s$. The initial neck feature map~$N_0$ is determined based on the size of the predicted bounding box (and hence varies for different detected objects), following the same scheme as in Mask R-CNN~\cite{lin2017feature, he2017mask} where $N_0 = P_{k_0}$ with
\begin{equation}
    k_0 = 2 + \min \left( \lfloor \log_2(\sqrt{wh} / 56) \rfloor, \,3 \right),
\end{equation}
and with $w$ and $h$ the width and height of the predicted bounding box respectively. For subsequent stages, the neck feature map is used which has twice the resolution compared to the map of previous stage, unless no higher resolution neck feature map is available. In general, we hence have $N_s = P_{k_s}$ with
\begin{equation}
    k_s = \max (k_0 - s, 2).
\end{equation}
Note that this is different from RefineMask~\cite{zhang2021refinemask}, which uses $k_s = 2$ for stages $1$, $2$, and $3$.

The remaining two inputs are the predicted bounding boxes and the query features, with one predicted bounding box and one query feature per detected object. The query feature is used by the detector to predict the class and bounding box of each detected object, and hence carries useful instance-level information condensed into a single feature.

\paragraph{Dense processing.} The first refinement stage (\ie stage~0) solely consists of dense processing on a 2D feature map.

At first, EffSeg applies the RoIAlign operation~\cite{he2017mask} on the $N_0$ neck feature maps to obtain the initial RoI-based 2D feature map of shape $[N_R, F_0, H_0, W_0]$ with $N_R$ the number of RoIs (\ie the number of detected objects), $F_0$ the feature size, $H_0$ the height of the map, and $W_0$ the width of the map. Note that the numeral subscripts, as those found in $F_0$, $H_0$, and $W_0$, indicate the refinement stage. In practice, EffSeg uses $F_0=256$, $H_0=14$, and $W_0=14$.

Next, the query features from the detector are fused with the 2D feature map obtained by the RoIAlign operation. The fusion consists in concatenating each of the RoI features with their corresponding query feature, processing the concatenated features using a two-layer \gls{MLP}, and adding the resulting features to the original RoI features. Fusing the query features allows to explicitly encode which object within the RoI box is considered the object of interest, as opposed to implicitly inferring this from the delineation of the RoI box. This is especially useful when having overlapping objects with similar bounding boxes.

After the query fusion, the 2D feature map gets further processed by a Fully Convolutional Network (FCN)~\cite{long2015fully}, similar to the one used in Mask R-CNN~\cite{he2017mask}, consisting of $4$ convolution layers separated by ReLU activations.

Finally, the resulting 2D feature map is used to obtain the coarse $14\times14$ segmentation mask predictions with a two-layer MLP. Additionally, EffSeg also uses a two-layer MLP to make refinement predictions, which are used to identify the cells (\ie locations) from the $14\times14$ grid that require a higher resolution and hence need to be refined.

\paragraph{Sparse processing.} The subsequent refinement stages (\ie stages 1, 2 and 3) solely consist of sparse processing using the Structure-Preserving Sparsity (\gls{SPS}) method (see Subsection~\ref{sec:ch5_sps} for more information about SPS).

At first, the SPS data structure is constructed or updated from previous stage. The $N_A$ features corresponding to the cells with the $10.000$ highest refinement scores, are categorized as active features, whereas the remaining $N_P$ features are labeled as passive features. The active and passive features are stored in $N_A \times F_{s-1}$ and $N_P \times F_{s-1}$ matrices respectively, with active feature indices ranging from $0$ to $N_A - 1$ and with passive feature indices ranging from $N_A$ to $N_A + N_P - 1$. The dense 2D index map of the SPS data structure is constructed from the stage~0 dense 2D feature map, or from the index map from previous stage, while taking the new feature indices into consideration due to the new split between active and passive features.

Thereafter, the SPS data structure is updated based on the upsampling of the feature grid by a factor~$2$. The number of active features~$N_A$ increases by a factor~$4$, as each parent cell gets subdivided into $4$ children cells. The children active features are computed from their parent active feature using a two-layer MLP, with a different MLP for each of the $4$ children. The dense 2D index map is updated based on the new feature indices (as the number of active features increased) and by copying the feature indices from the parent cell of passive features to its children cells. Note that the passive features themselves remain unchanged.

Next, the active features are fused with their corresponding neck feature, which is sampled from the neck feature map $N_s$ in the center of the active feature cell. The fusion consists in concatenating each of the active features with their corresponding neck feature, processing the concatenated features using a two-layer MLP, and adding the resulting features to the original active features.

Afterwards, the feature size of the active and passive features are divided by $2$ using a shared one-layer MLP. We hence have $F_{s+1} = F_s / 2$, decreasing the feature size by a factor~$2$ every refinement stage, as done in RefineMask~\cite{zhang2021refinemask}.

After decreasing the feature sizes, the active features are further updated using the processing module, which does most of the heavy computation. The processing module supports any 2D operation thanks to the versatility of the SPS method. Our default EffSeg implementation uses the Semantic Fusion Module (SFM) from RefineMask~\cite{zhang2021refinemask}, which fuses (\ie adds) the features obtained by three parallel convolution layers using a $3\times3$ kernel and dilations $1$, $3$, and~$5$. In  Subsection~\ref{sec:ch5_comp_proc}, we compare the performance of EffSeg heads using different processing modules.

Finally, the resulting active features are used to obtain the new segmentation and refinement predictions in their corresponding cells. Both the segmentation branch and the refinement branch use a two-layer MLP, as in stage~0. 

\paragraph{Training.} During training, EffSeg applies segmentation and refinement losses on the segmentation and refinement predictions from each EffSeg stage~$s$, where each of these predictions are made for a particular cell from the 2D grid. The ground-truth segmentation targets are obtained by sampling the ground-truth mask in the center of the cell, and the ground-truth refinement targets are determined by evaluating whether the cell contains both foreground and background or not. We use the cross-entropy loss for both the segmentation and refinement losses, with loss weights $(0.25, 0.375, 0.375, 0.5)$ and $(0.25, 0.25, 0.25, 0.25)$ for stages 0 to 3 respectively.

\paragraph{Inference.} During inference, EffSeg additionally constructs the desired segmentation masks based on the segmentation predictions from each stage. The segmentation predictions from stage~0 already correspond to dense $14\times14$ segmentation masks, and hence do not require any post-processing. In each subsequent stage, the segmentation masks from previous stage are upsampled by a factor~$2$, and the sparse segmentation predictions in active feature locations are used to overwrite the old segmentation predictions in their corresponding cells. After performing this process for three refinement stages, the coarse $14\times14$ masks are upsampled to fine-grained $112\times112$ segmentation masks. Finally, the image-size segmentation masks are obtained by pasting the RoI-based $112\times112$ segmentation masks inside their corresponding RoI boxes using bilinear interpolation.

The segmentation confidence scores~$s_\text{seg}$ are computed by taking the product of the classification score~$s_\text{cls}$ and the mask score~$s_\text{mask}$ averaged over the predicted foreground pixels, which gives
\begin{equation}
    s_\text{seg} = s_\text{cls} \cdot \frac{1}{|\mathcal{F}|} \sum_i^\mathcal{F} s_{\text{mask}, i}
\end{equation}
with $\mathcal{F}$ the set of all predicted foreground pixels.

\section{Experiments}

\subsection{Experimental setup} \label{sec:ch5_setup}

\paragraph{Datasets.} We perform experiments on the COCO~\cite{lin2014microsoft} instance and panoptic segmentation benchmarks. We train on the 2017 training set and evaluate on the 2017 validation set.

\paragraph{Implementation details.} We follow the standardized settings mentioned in Section~\ref{sec:ch2_settings} for the backbone settings, neck settings, data settings, and (some of the) training settings. We use the same FPN~\cite{lin2017feature} settings as in Subsection~\ref{sec:ch4_setup_v2}, but applied to a $P_2$-$P_7$ feature pyramid instead of a $P_3$-$P_7$ feature pyramid. For the FQDet head, we use the settings from Subsection~\ref{sec:ch4_setup_v1} and the improvements from Subsection~\ref{sec:ch4_improving}. For the FQDetV2 head, we use the settings from Subsection~\ref{sec:ch4_design_v2}, except that we use $300$ or $600$~queries instead of $1500$, $3$~different aspect ratios instead of~$7$, and no box scoring mechanism. For the EffSeg head, we refer to the settings mentioned in Subsection~\ref{sec:ch5_detail}.

We train our EffSeg models using a 1x or 2x training schedule. The 1x schedule contains 12 training epochs with a learning rate decrease after the 9th epoch with a factor~$0.1$, and the 2x schedule consists of 24 training epochs with learning rate drops after the 18th and 22nd epochs with a factor~$0.1$. The models are trained on two GPUs, with each a batch size of one. For the remaining training settings, see Subsection~\ref{sec:ch2_settings}.

In order to obtain the panoptic segmentation predictions during inference, we change the IoU threshold used during box \gls{NMS} to $0.5$, and additionally perform following six post-processing steps:

\begin{enumerate}
    \vspace{-0.1cm}
    \item Predictions are removed with a classification score lower than $0.3$.
    \item Mask \gls{NMS} with an IoU threshold of $0.75$ is applied on predictions.
    \item Overlapping segmentation masks are transformed into non-overlapping segmentation masks as requested by the panoptic segmentation task. This is done by choosing for every pixel the mask with the highest combined classification and mask score (\ie maximum $s_{\text{cls}} \,\cdot\, s_{\text{mask}}$), unless the highest score is lower than $0.35$ in which case no mask is assigned to the pixel.
    \item Predictions are removed for which the non-overlapping mask no longer sufficiently resembles the original overlapping mask, \ie when the IoU of the non-overlapping mask and the original mask is lower than $0.6$.
    \item Predictions are removed with a mask containing fewer than $150$ pixels.
    \item Different masks of a same stuff class are merged into one.
\end{enumerate}

\paragraph{Metrics.} As performance metrics for the instance segmentation task, we use the COCO \gls{AP} metrics evaluated using the original COCO annotations, as well as the AP$^*$ and the boundary-based AP$^{B*}$ metrics evaluated using the higher quality LVIS annotations (see Subsection~\ref{sec:ch2_metrics}). As performance metrics for the panoptic segmentation task, we use the \gls{PQ}, SQ, RQ, PQ$^{\text{Th}}$, and PQ$^{\text{St}}$ metrics as introduced in Subsection~\ref{sec:ch2_metrics}.

For the computational cost metrics, we report the Params, iFLOPs and iFPS metrics. Given that there are no training-based cost metrics, the iFLOPs and iFPS metrics can be abbreviated to FLOPs and FPS respectively. The FLOPs and FPS metrics are computed based on the average over the first $100$ images of the validation set, where the inference speeds are measured on an NVIDIA GeForce RTX 3060 Ti using a batch size of one. See Subsection~\ref{sec:ch2_metrics} for more information about these metrics.

\paragraph{Baselines.} Our baselines are the Mask R-CNN~\cite{he2017mask}, PointRend~\cite{kirillov2020pointrend}, and RefineMask~\cite{zhang2021refinemask} heads. Mask R-CNN could be considered as the entry-level baseline without any enhancements towards fine-grained segmentation. PointRend and RefineMask on the other hand are two baselines with improvements towards fine-grained segmentation, with RefineMask our main baseline due to its superior performance. We use the implementations from MMDetection~\cite{mmdetection} for both the Mask R-CNN and PointRend heads, whereas for RefineMask we use the latest version from the official implementation~\cite{zhang2021refinemask}.

In order to provide a fair comparison with EffSeg, we (only) consider the enhanced versions of above baselines, called Mask R-CNN++, PointRend++, and RefineMask++. The enhanced versions additionally perform query fusion and mask-based score weighting as done in EffSeg (see Subsection~\ref{sec:ch5_detail}). For PointRend++, we moreover replace the coarse MLP-based head by the same FCN-based head as used in Mask R-CNN, yielding improved performance without significant changes in computational cost.

Note that Mask Transfiner~\cite{ke2022mask} is not used as baseline, due to irregularities in the reported experimental results and due to non-standard experimental settings as discussed in \cite{transfiner2022irregularities}.

\subsection{Instance segmentation experiments} \label{sec:ch5_ins_seg}

\begin{table*}
    \centering
    \caption[EffSeg - Instance segmentation results]{Results of the instance segmentation experiments.}
    \vspace{0.2cm}
    \resizebox{\columnwidth}{!}{
    \begin{tabular}{c c c c|c c c c c c|c c|c c c}
        \toprule
        Detector & Seg. head & Qrys & Ep. & AP & AP$_{50}$ & AP$_{75}$ & AP$_S$ & AP$_M$ & AP$_L$ & AP$^*$ & AP$^{B*}$ & Params & FLOPs & FPS \\
        \midrule
        FQDet & Mask R-CNN++ & $300$ & $12$ & $40.0$ & $61.0$ & $43.7$ & $20.4$ & $42.8$ & $58.4$ & $42.2$ & $29.6$ & $37.5$ M & $220.5$ G & $10.7$ \\
        FQDet & PointRend++ & $300$ & $12$ & $40.8$ & $61.2$ & $44.3$ & $20.5$ & $43.9$ & $59.9$ & $43.7$ & $32.9$ & $37.8$ M & $288.0$ G & $6.8$ \\
        FQDet & RefineMask++ & $300$ & $12$ & $41.3$ & $61.4$ & $\mathbf{45.1}$ & $21.0$ & $44.0$ & $\mathbf{61.0}$ & $\mathbf{44.4}$ & $\mathbf{33.5}$ & $41.2$ M & $446.4$ G & $6.5$ \\
        FQDet & EffSeg (ours) & $300$ & $12$ & $\mathbf{41.5}$ & $\mathbf{61.7}$ & $45.0$ & $\mathbf{21.1}$ & $\mathbf{44.5}$ & $60.5$ & $44.3$ & $33.4$ & $38.8$ M & $245.5$ G & $8.0$ \\
        \midrule
        FQDet & Mask R-CNN++ & $300$ & $24$ & $40.9$ & $62.3$ & $44.5$ & $20.6$ & $43.6$ & $58.6$ & $43.0$ & $30.4$ & $37.5$ M & $220.0$ G & $10.7$ \\
        FQDet & PointRend++ & $300$ & $24$ & $41.9$ & $62.8$ & $45.7$ & $\mathbf{22.0}$ & $44.8$ & $61.1$ & $44.5$ & $33.5$ & $37.8$ M & $287.4$ G & $6.8$ \\
        FQDet & RefineMask++ & $300$ & $24$ & $42.1$ & $62.8$ & $45.7$ & $21.6$ & $44.9$ & $61.7$ & $\mathbf{45.4}$ & $\mathbf{34.4}$ & $41.2$ M & $445.8$ G & $6.5$ \\
        FQDet & EffSeg (ours) & $300$ & $24$ & $\mathbf{42.4}$ & $\mathbf{63.1}$ & $\mathbf{46.2}$ & $21.7$ & $\mathbf{45.2}$ & $\mathbf{61.9}$ & $45.1$ & $34.1$ & $38.8$ M & $244.4$ G & $8.0$ \\
        \midrule
        \midrule
        FQDetV2 & Mask R-CNN++ & $300$ & $12$ & $40.0$ & $60.2$ & $44.0$ & $20.5$ & $43.2$ & $58.4$ & $42.6$ & $29.7$ & $40.5$ M & $223.3$ G & $10.6$ \\
        FQDetV2 & PointRend++ & $300$ & $12$ & $40.9$ & $60.2$ & $45.0$ & $20.8$ & $43.9$ & $60.1$ & $43.7$ & $32.9$ & $40.8$ M & $292.7$ G & $6.6$ \\
        FQDetV2 & RefineMask++ & $300$ & $12$ & $41.3$ & $60.5$ & $45.4$ & $20.5$ & $\mathbf{44.3}$ & $\mathbf{61.3}$ & $\mathbf{44.4}$ & $33.6$ & $44.2$ M & $451.6$ G & $6.4$ \\
        FQDetV2 & EffSeg (ours) & $300$ & $12$ & $\mathbf{41.5}$ & $\mathbf{60.9}$ & $\mathbf{45.6}$ & $\mathbf{21.6}$ & $44.1$ & $61.1$ & $\mathbf{44.4}$ & $\mathbf{33.7}$ & $41.8$ M & $251.8$ G & $7.8$ \\
        \midrule
        FQDetV2 & Mask R-CNN++ & $300$ & $24$ & $41.5$ & $62.5$ & $45.4$ & $21.8$ & $44.4$ & $59.6$ & $43.6$ & $30.7$ & $40.5$ M & $223.0$ G & $10.6$ \\
        FQDetV2 & PointRend++ & $300$ & $24$ & $42.4$ & $62.5$ & $46.5$ & $22.2$ & $45.0$ & $61.7$ & $45.0$ & $34.0$ & $40.8$ M & $292.7$ G & $6.7$ \\
        FQDetV2 & RefineMask++ & $300$ & $24$ & $\mathbf{42.8}$ & $\mathbf{62.8}$ & $\mathbf{46.8}$ & $\mathbf{23.0}$ & $\mathbf{45.6}$ & $\mathbf{62.6}$ & $\mathbf{45.6}$ & $\mathbf{34.6}$ & $44.2$ M & $451.6$ G & $6.4$ \\
        FQDetV2 & EffSeg (ours) & $300$ & $24$ & $42.6$ & $62.4$ & $46.8$ & $22.2$ & $45.2$ & $62.3$ & $45.3$ & $34.3$ & $41.8$ M & $251.3$ G & $7.8$ \\
        \midrule
        \midrule
        FQDetV2 & Mask R-CNN++ & $600$ & $12$ & $41.3$ & $62.2$ & $45.3$ & $21.8$ & $44.2$ & $60.0$ & $44.0$ & $31.0$ & $40.5$ M & $226.7$ G & $10.4$ \\
        FQDetV2 & PointRend++ & $600$ & $12$ & $42.0$ & $62.2$ & $45.9$ & $22.1$ & $44.9$ & $60.8$ & $45.0$ & $34.0$ & $40.8$ M & $296.2$ G & $6.6$ \\
        FQDetV2 & RefineMask++ & $600$ & $12$ & $\mathbf{42.7}$ & $\mathbf{62.8}$ & $\mathbf{46.9}$ & $\mathbf{23.2}$ & $\mathbf{45.8}$ & $61.7$ & $\mathbf{46.1}$ & $\mathbf{35.1}$ & $44.2$ M & $455.1$ G & $6.3$ \\
        FQDetV2 & EffSeg (ours) & $600$ & $12$ & $42.4$ & $62.5$ & $46.4$ & $22.5$ & $45.0$ & $\mathbf{62.1}$ & $45.6$ & $34.6$ & $41.8$ M & $262.6$ G & $7.5$ \\
        \midrule
        FQDetV2 & Mask R-CNN++ & $600$ & $24$ & $42.7$ & $64.2$ & $46.9$ & $23.7$ & $45.2$ & $61.2$ & $45.1$ & $31.9$ & $40.5$ M & $226.7$ G & $10.4$ \\
        FQDetV2 & PointRend++ & $600$ & $24$ & $43.2$ & $64.1$ & $47.2$ & $23.3$ & $45.9$ & $62.5$ & $46.4$ & $35.3$ & $40.8$ M & $296.0$ G & $6.6$ \\
        FQDetV2 & RefineMask++ & $600$ & $24$ & $\mathbf{44.0}$ & $\mathbf{64.6}$ & $\mathbf{48.5}$ & $\mathbf{23.6}$ & $\mathbf{47.1}$ & $\mathbf{63.2}$ & $\mathbf{47.2}$ & $\mathbf{36.2}$ & $44.2$ M & $455.0$ G & $6.3$ \\
        FQDetV2 & EffSeg (ours) & $600$ & $24$ & $43.7$ & $64.3$ & $48.0$ & $23.5$ & $46.5$ & $\mathbf{63.2}$ & $46.8$ & $35.7$ & $41.8$ M & $262.3$ G & $7.6$ \\
        \bottomrule
    \end{tabular}}
    \label{tab:ch5_ins_seg}
\end{table*}

We perform our instance segmentation experiments using the ResNet-50 backbone~\cite{he2016deep} and the FPN neck~\cite{lin2017feature}. The results are found in Table~\ref{tab:ch5_ins_seg}, from which we make following observations.

\paragraph{Performance.} Performance-wise, we can see that Mask R-CNN++ performs the worst, that RefineMask++ and EffSeg perform the best, and that PointRend++ performs somewhere in between. This is in line with the arguments presented earlier, which we reiterate here for convenience. 

Mask R-CNN++ predicts a $28\times28$ mask per RoI, which is too coarse to capture the fine details of many objects. This is especially true for large objects, as can be seen from the significantly lower AP$_{L}$ values compared to the other segmentation heads. 

PointRend++ performs better compared to Mask R-CNN++ by predicting a $112\times112$ mask, yielding significant gains in the boundary accuracy AP$^{B*}$. However, PointRend++ does not access neighboring features during the refinement process, resulting in lower segmentation performance compared RefineMask++ and Effseg, which both do leverage the context provided by neighboring features. 

Finally, we can see that RefineMask++ and EffSeg perform the best, both reaching similar segmentation performance.

\begin{table}
    \centering
    \caption[EffSeg - Computational cost metrics (instance head only)]{Computational cost metrics of the instance segmentation heads alone. The relative metrics (three rightmost columns) are \wrt to RefineMask++. We highlight the EffSeg relative metrics in bold (not necessarily the best).}
    \vspace{0.2cm}
    \resizebox{0.9\columnwidth}{!}{
    \begin{tabular}{c|c c c|c c c}
        \toprule
         \multirow{2}{*}{Seg. head} & \multirow{2}{*}{Params} & \multirow{2}{*}{FLOPs} & \multirow{2}{*}{FPS} & Params & FLOPs & FPS \\
         & & & & decrease & decrease & gain \\
         \midrule
         Mask R-CNN++ & $2.9$ M & $57.0$ G & $35.0$ & $56$\% & $80$\% & $215$\% \\
         PointRend++ & $3.2$ M & $126.7$ G & $12.0$ & $52$\% & $56$\% & $8$\% \\
         RefineMask++ & $6.6$ M & $285.6$ G & $11.1$ & $0$\% & $0$\% & $0$\% \\
         EffSeg (ours) & $4.2$ M & $85.3$ G & $16.0$ & $\mathbf{36}$\textbf{\%} & $\mathbf{70}$\textbf{\%} & $\mathbf{44}$\textbf{\%} \\
         \bottomrule
    \end{tabular}}
    \label{tab:ch5_eff_ins}
\end{table}

\paragraph{Efficiency.} In Table~\ref{tab:ch5_ins_seg}, we can find the computational cost metrics of the different models \textit{as a whole}, \ie containing both the computational costs originating from the segmentation head, as well as those from the backbone, neck, and detector (\ie detection head). To provide a better comparison between the different segmentation heads, we also report the computational cost metrics of the instance segmentation heads \textit{alone} in Table~\ref{tab:ch5_eff_ins}. More specifically, we provide the absolute cost metrics for each of the instance segmentation heads, as well as some relative cost metrics measuring the head's efficiency compared to RefineMask++. Note that these cost metrics were obtained by applying the instance segmentation heads on the FQDetV2 detector with $300$ queries, and that similar results are found when using a different detector or a different number of queries.

As expected, we can see that Mask R-CNN++ is computationally the cheapest, given that it only predicts a $28\times28$ mask instead of a $112\times112$ mask. From the three remaining heads, RefineMask++ is clearly the most expensive one, as it performs computation on all locations within the RoI, instead of sparsely. PointRend++ and EffSeg are lying somewhere in between, being more computationally expensive than Mask R-CNN++, but cheaper than RefineMask++.

Finally, when comparing RefineMask++ with EffSeg, we can see that EffSeg uses $36$\% fewer parameters, reduces the number of inference FLOPs by $70$\%, and increases the inference FPS by $44$\%.

\paragraph{Performance vs. Efficiency.} When we consider both the performance and the efficiency metrics, we can see that the EffSeg head comes out on top. On the one hand, we notice that EffSeg clearly outperforms Mask R-CNN++, especially on large objects and near object boundaries, with only a small amount of additional computation. On the other hand, we observe that EffSeg obtains excellent instance segmentation performance similar to RefineMask++ (\ie the best performing baseline), while reducing the inference FLOPs by $70$\% and increasing the number of inference FPS by $44$\%.

\paragraph{Qualitative results.} In what follows, we provide some qualitative results to visually assess the instance segmentation performance. For this, we use the segmentation models with the FQDetV2 detector and $300$ queries, trained for 24 epochs.

\begin{figure}
    \centering
    \includegraphics[scale=0.185]{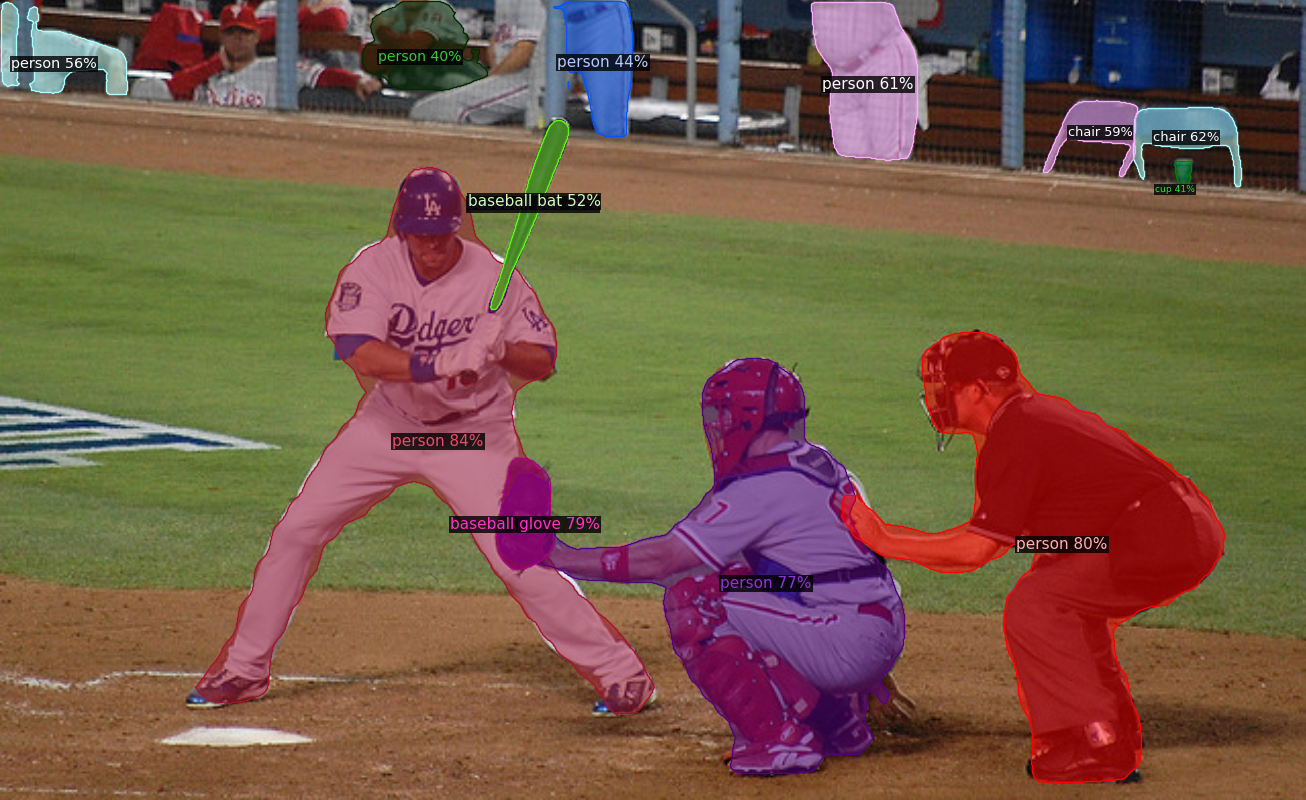} \\
    \vspace{0.2cm}
    \includegraphics[scale=0.185]{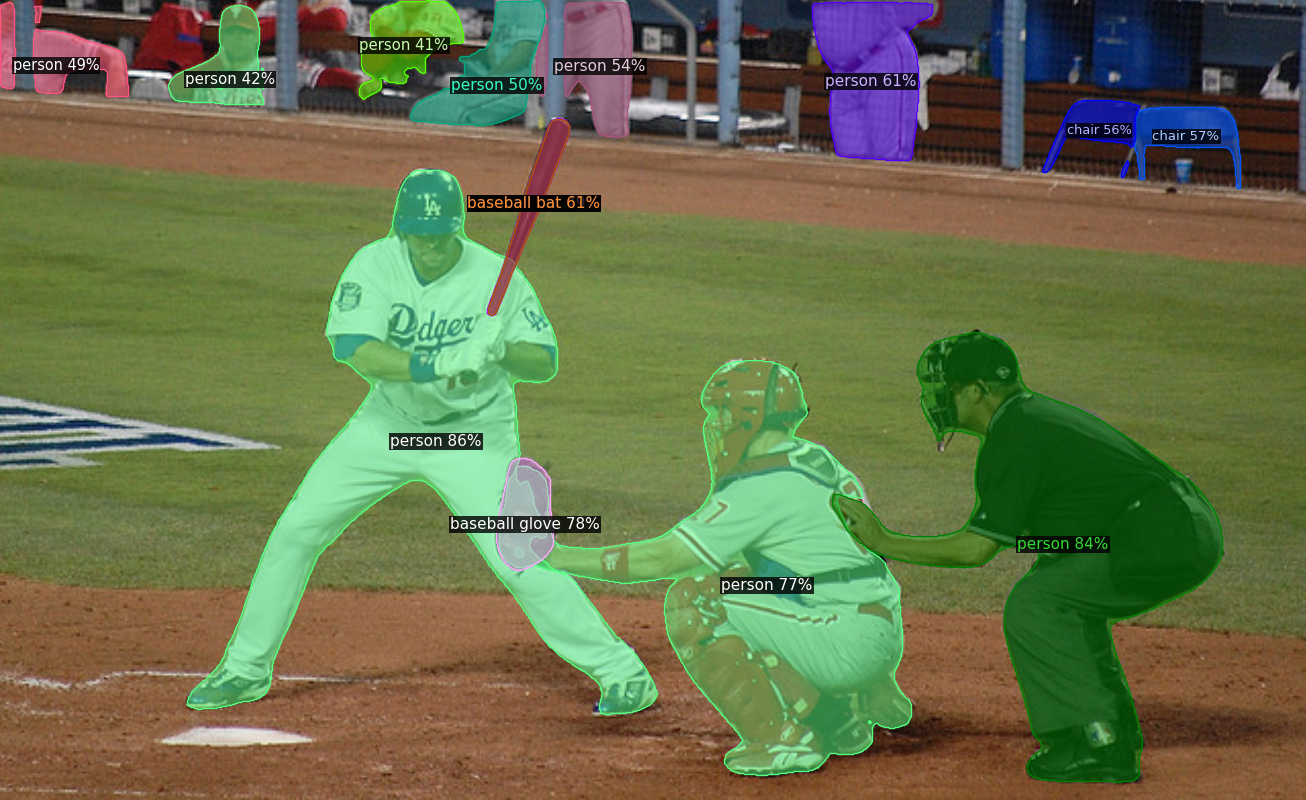} \\
    \vspace{0.2cm}
    \includegraphics[scale=0.185]{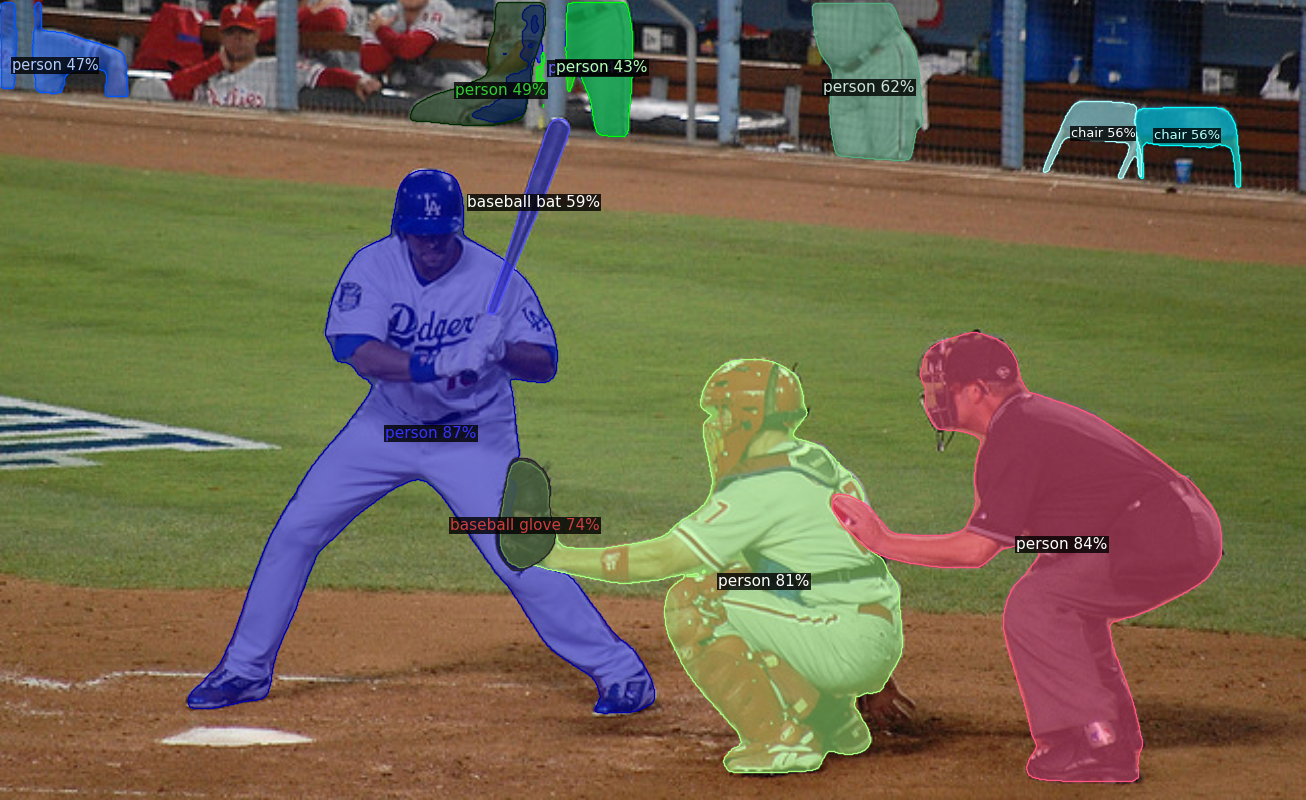}
    \caption[EffSeg qualitative (instance) - Example 1]{Example image with instance segmentation predictions using the Mask R-CNN++ (\textit{top}), RefineMask++ (\textit{middle}), and EffSeg (\textit{bottom}) segmentation heads.}
    \label{fig:ch5_ins_ex1}
\end{figure}

\begin{figure}
    \centering
    \includegraphics[scale=0.185]{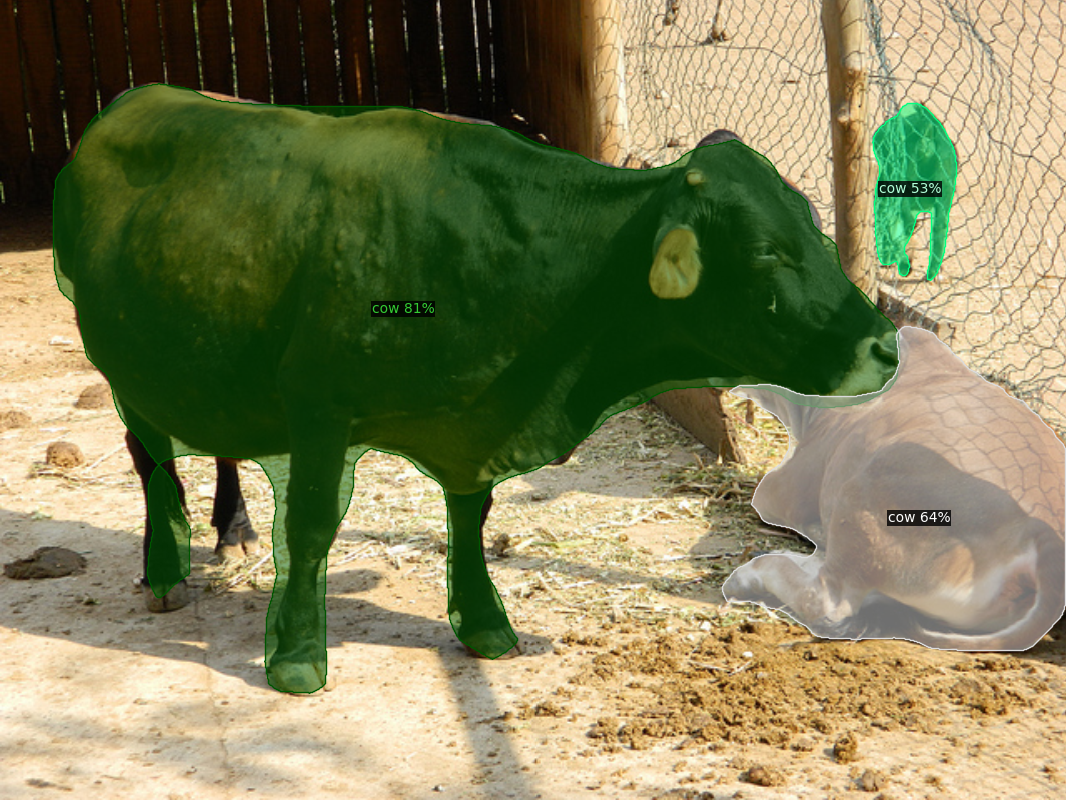} \\
    \vspace{0.2cm}
    \includegraphics[scale=0.185]{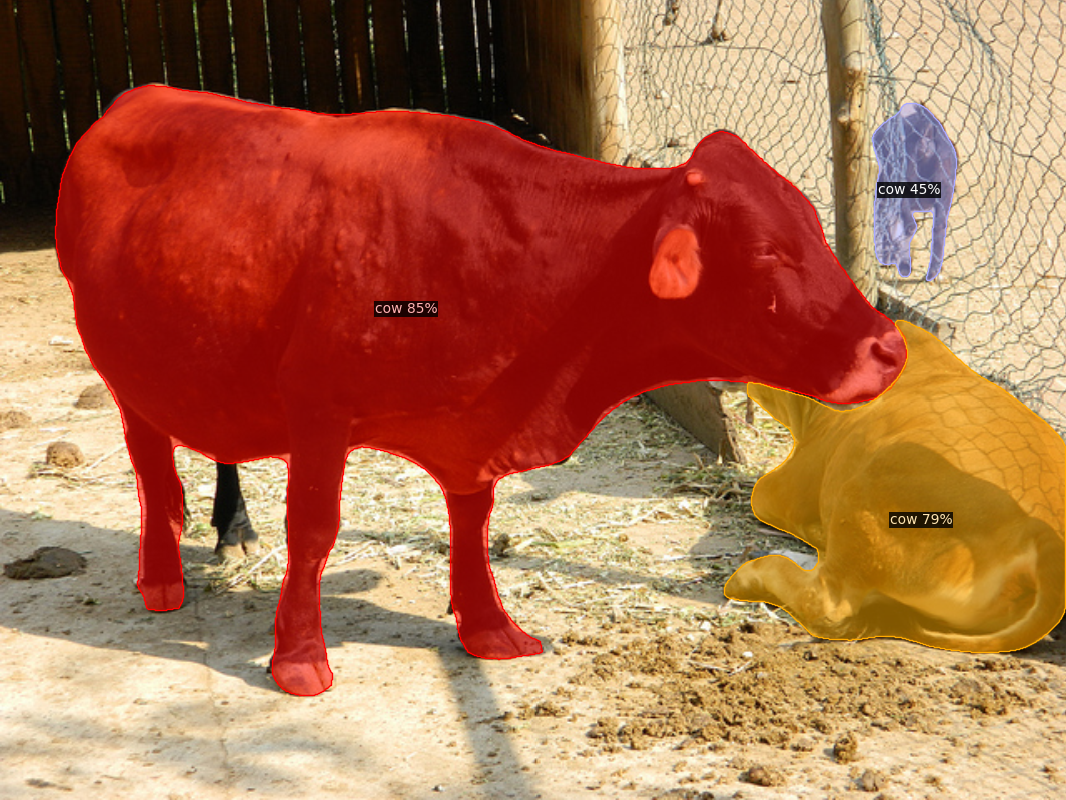} \\
    \vspace{0.2cm}
    \includegraphics[scale=0.185]{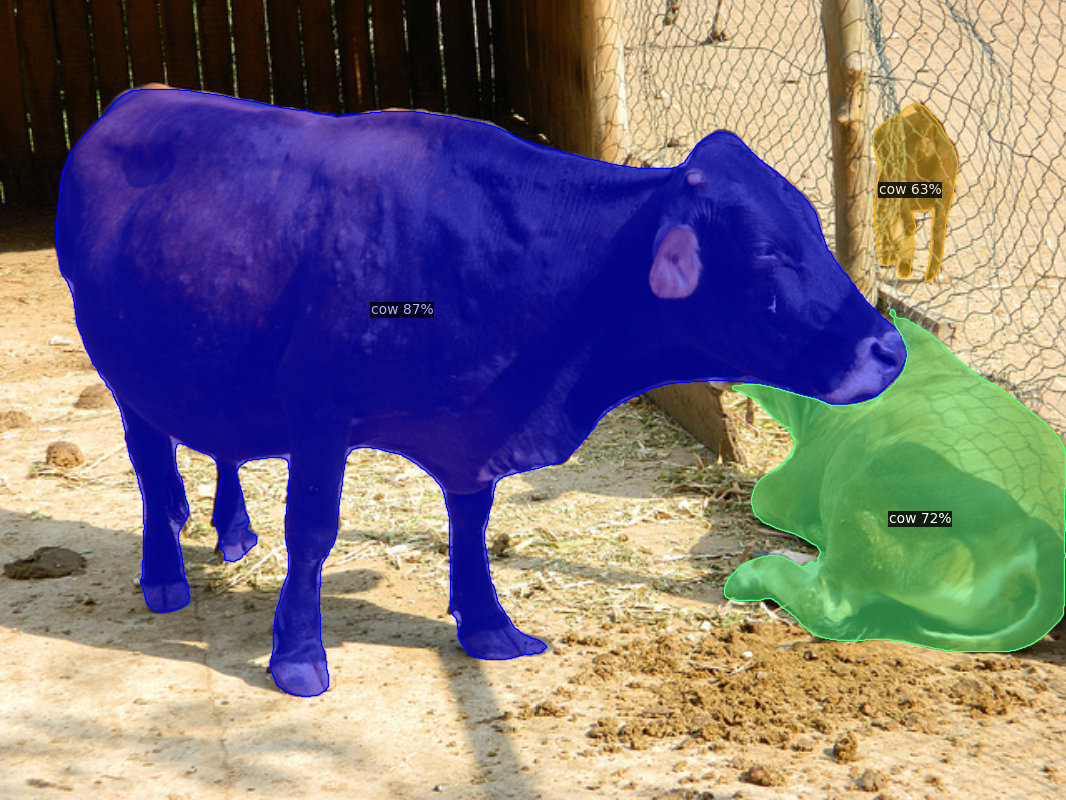}
    \caption[EffSeg qualitative (instance) - Example 2]{Example image with instance segmentation predictions using the Mask R-CNN++ (\textit{top}), RefineMask++ (\textit{middle}), and EffSeg (\textit{bottom}) segmentation heads.}
    \label{fig:ch5_ins_ex2}
\end{figure}

In Figure~\ref{fig:ch5_ins_ex1} and Figure~\ref{fig:ch5_ins_ex2}, we show two example images containing the instance segmentation predictions of the Mask R-CNN++ (top), RefineMask++ (middle), and EffSeg (bottom) segmentation heads. We can see that the segmentation masks from the Mask R-CNN++ head are relatively inaccurate at times, containing undulating mask boundaries and mask inaccuracies near junctions (\eg near the neck of the left baseball player in Figure~\ref{fig:ch5_ins_ex1} or near the transition between body and legs of the left cow in Figure~\ref{fig:ch5_ins_ex2}). The RefineMask++ and EffSeg heads on the other hand provide more accurate segmentation masks, with smoother mask boundaries and crispier mask predictions near junctions. Additionally, we can see that both the Mask R-CNN++ and RefineMask++ heads miss one leg of the left cow in Figure~\ref{fig:ch5_ins_ex2}, whereas the segmentation mask of the EffSeg head correctly contains all four legs.

In what follows, we also provide some failure cases of the EffSeg segmentation head. Note that these failures cases are not specific to the EffSeg head and are also found in the baseline heads.

\begin{figure}
    \centering
    \includegraphics[scale=0.28]{inc_box_box} \\
    \vspace{0.4cm}
    \includegraphics[scale=0.28]{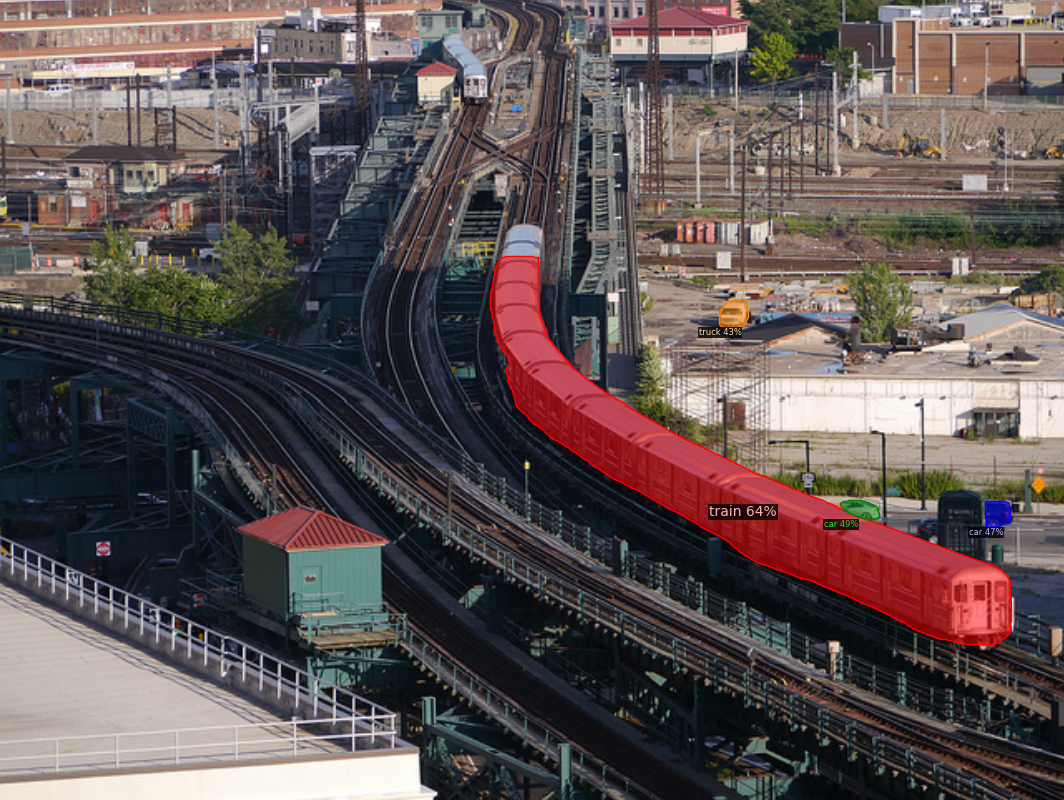}
    \caption[EffSeg qualitative (instance) - Incorrect box]{Example image with incorrect bounding box prediction (\textit{top}) leading to incorrect segmentation mask (\textit{bottom}).}
    \label{fig:ch5_inc_box}
\end{figure}

In Figure~\ref{fig:ch5_inc_box}, we show an example image where the predicted bounding box only covers part of the object of interest (\ie part of the train in the middle). As the EffSeg head is a \gls{RoI}-based segmentation head meaning that it only segments inside the predicted bounding box, the resulting segmentation mask is incomplete by missing part of the train object lying outside of the predicted bounding box. This is a clear limitation of RoI-based (or two-stage) segmentation heads as they strongly rely on the accuracy of the predicted bounding boxes by assuming the boxes fully cover the objects to segment. A possible way to mitigate this problem, is to increase the size of the used RoI boxes with a fixed factor (\eg with a factor~$1.1$), and as such increasing the probability that RoI boxes fully contain the objects to be segmented.

\begin{figure}
    \centering
    \begin{minipage}{0.49\textwidth}
        \centering
        \includegraphics[scale=0.2]{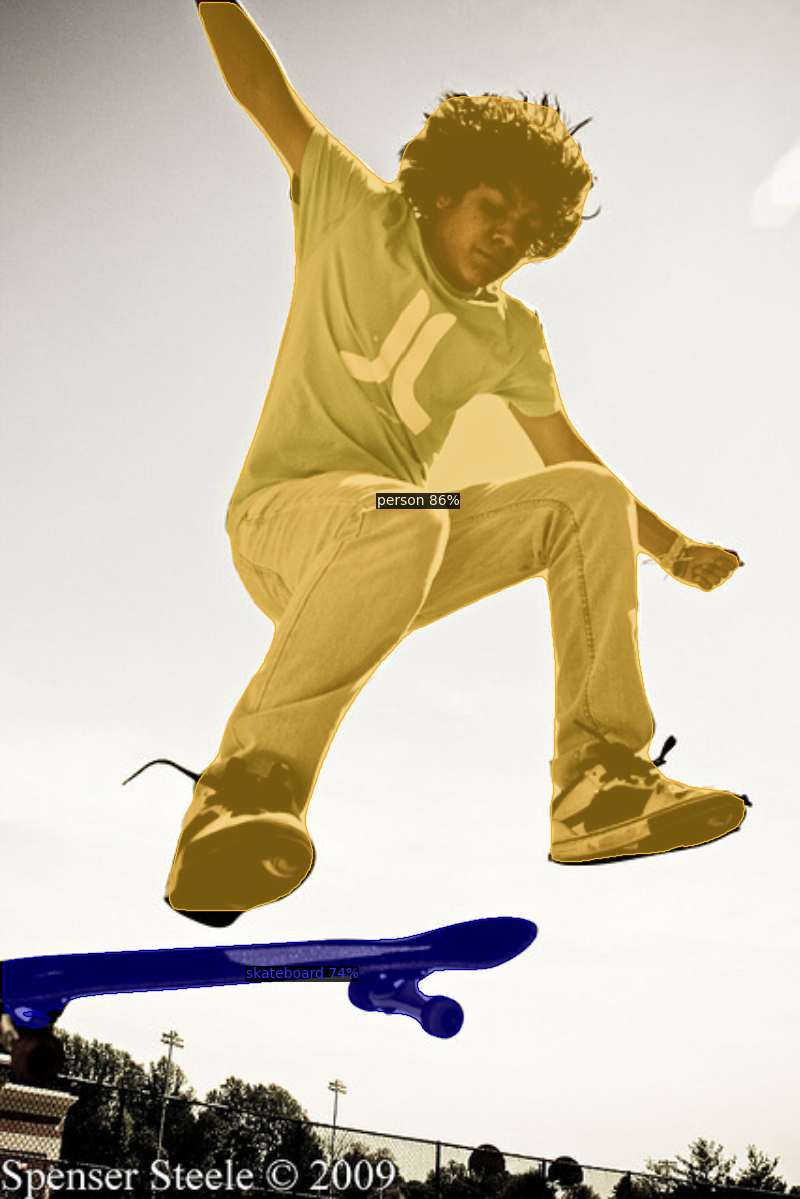}
    \end{minipage} \hfill
    \begin{minipage}{0.49\textwidth}
        \centering
        \includegraphics[scale=0.2]{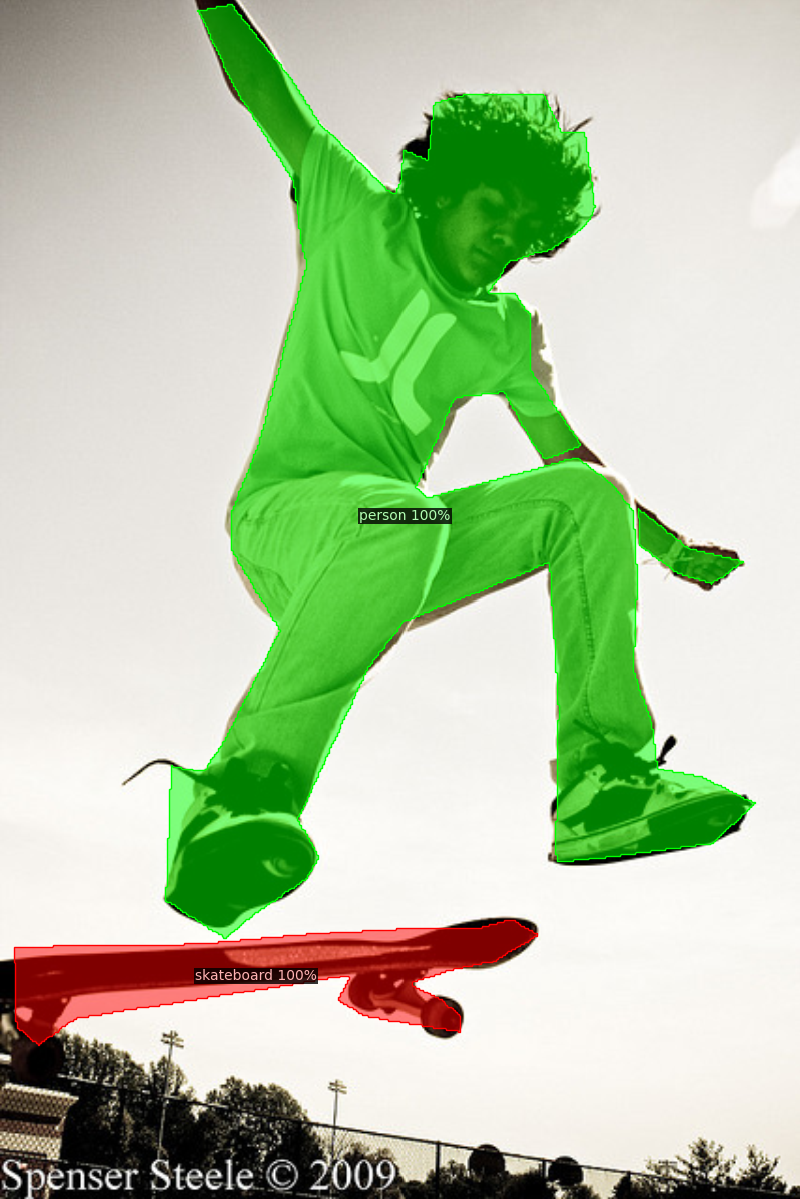}
    \end{minipage}
    \caption[EffSeg qualitative (instance) - Ambiguous segmentation mask]{Example image with ambiguous segmentation mask. The left image contains the predicted segmentation masks and the right image the ground-truth segmentation masks.}
    \label{fig:ch5_amb_seg}
\end{figure}

In Figure~\ref{fig:ch5_amb_seg}, we provide an example image where the segmentation mask to be predicted is ambiguous, due to annotation inconsistencies. It is namely unclear whether the triangular area (of background) enclosed by the boy, should be part of the boy segmentation mask or not. Intuitively, it would be logical to exclude it from the boy's mask, as this image area simply does not correspond to the boy. However, in the training set, similar enclosed areas were sometimes annotated as belonging to the object and sometimes not, leading to ambiguity at inference. In Figure~\ref{fig:ch5_amb_seg}, we can see that EffSeg guessed wrong by predicting the triangular area as belonging to the boy mask, whereas the ground-truth mask excludes it. Moreover, note from Figure~\ref{fig:ch5_amb_seg} that the provided ground-truth masks are not very accurate, explaining why training and evaluating fine-grained segmentation methods with COCO annotations is not ideal.

\subsection{Panoptic segmentation experiments}

\begin{table*}
    \centering
    \caption[EffSeg - Panoptic segmentation results]{Results of the panoptic segmentation experiments.}
    \vspace{0.2cm}
    \resizebox{\columnwidth}{!}{
    \begin{tabular}{c c c c|c c c|c c|c c c}
        \toprule
        Detector & Seg. head & Qrys & Ep. & PQ & SQ & RQ & PQ$^{Th}$ & PQ$^{St}$ & Params & FLOPs & FPS \\
        \midrule
        FQDetV2 & Mask R-CNN++ & $300$ & $12$ & $44.8$ & $80.8$ & $54.1$ & $50.7$ & $35.9$ & $40.6$ M & $191.5$ G & $11.6$ \\
        FQDetV2 & PointRend++ & $300$ & $12$ & $46.3$ & $81.1$ & $55.6$ & $52.5$ & $37.1$ & $40.9$ M & $229.6$ G & $8.1$ \\
        FQDetV2 & RefineMask++ & $300$ & $12$ & $46.1$ & $81.5$ & $55.2$ & $52.0$ & $37.3$ & $44.2$ M & $371.8$ G & $7.2$ \\
        FQDetV2 & EffSeg (ours) & $300$ & $12$ & $46.4$ & $82.2$ & $55.3$ & $52.8$ & $36.7$ & $41.8$ M & $231.5$ G & $8.1$ \\
        \midrule
        FQDetV2 & Mask R-CNN++ & $300$ & $24$ & $46.5$ & $80.8$ & $56.1$ & $52.2$ & $37.9$ & $40.6$ M & $191.7$ G & $11.2$ \\
        FQDetV2 & PointRend++ & $300$ & $24$ & $47.7$ & $82.3$ & $57.1$ & $53.7$ & $38.7$ & $40.9$ M & $229.0$ G & $7.8$ \\
        FQDetV2 & RefineMask++ & $300$ & $24$ & $47.3$ & $82.5$ & $56.4$ & $53.2$ & $38.5$ & $44.2$ M & $370.1$ G & $7.7$ \\
        FQDetV2 & EffSeg (ours) & $300$ & $24$ & $47.6$ & $82.5$ & $56.7$ & $53.8$ & $38.2$ & $41.8$ M & $231.3$ G & $8.3$ \\
        \midrule
        \midrule
        FQDetV2 & Mask R-CNN++ & $600$ & $12$ & $45.8$ & $80.6$ & $55.4$ & $51.9$ & $36.6$ & $40.6$ M & $218.6$ G & $9.9$ \\
        FQDetV2 & PointRend++ & $600$ & $12$ & $47.0$ & $81.1$ & $56.4$ & $53.1$ & $37.7$ & $40.9$ M & $289.7$ G & $6.3$ \\
        FQDetV2 & RefineMask++ & $600$ & $12$ & $47.2$ & $81.5$ & $56.5$ & $53.0$ & $38.4$ & $44.2$ M & $433.2$ G & $6.3$ \\
        FQDetV2 & EffSeg (ours) & $600$ & $12$ & $47.0$ & $82.1$ & $56.2$ & $53.4$ & $37.3$ & $41.8$ M & $262.6$ G & $6.7$ \\
        \midrule
        FQDetV2 & Mask R-CNN++ & $600$ & $24$ & $47.3$ & $81.6$ & $57.0$ & $53.2$ & $38.4$ & $40.6$ M & $218.2$ G & $10.5$ \\
        FQDetV2 & PointRend++ & $600$ & $24$ & $48.4$ & $82.0$ & $58.1$ & $54.3$ & $39.5$ & $40.9$ M & $290.2$ G & $6.3$ \\
        FQDetV2 & RefineMask++ & $600$ & $24$ & $48.6$ & $82.2$ & $58.1$ & $54.4$ & $39.8$ & $44.2$ M & $432.7$ G & $6.6$ \\
        FQDetV2 & EffSeg (ours) & $600$ & $24$ & $48.4$ & $82.3$ & $57.8$ & $54.6$ & $39.1$ & $41.8$ M & $262.8$ G & $7.4$ \\
        \bottomrule
    \end{tabular}}
    \label{tab:ch5_pan_seg}
\end{table*}

Additionally, we also perform panoptic segmentation experiments using the ResNet-50 backbone~\cite{he2016deep} and the FPN neck~\cite{lin2017feature}. The results are found in Table~\ref{tab:ch5_pan_seg}, from which we make following observations.

\paragraph{Performance.} Performance-wise, we can see that the Mask R-CNN++ head still performs worse compared to the other segmentation heads, as it only predicts a $28\times28$ mask instead of a $112\times112$ mask. Additionally, we also observe that the RefineMask++ and the EffSeg heads reach similar performance, as during the instance segmentation experiments.

However, this time the PointRend++ head also reaches performance close to the RefineMask++ and EffSeg heads, while PointRend++ performed clearly worse during the instance segmentation experiments. It is currently unclear to us why this is the case. We hypothesize that by tuning the hyperparameters (which were left unchanged from the instance segmentation experiments), similar performance gaps could arise as experienced during the instance segmentation experiments. However, additional experimentation would be required to verify this claim.

We also notice some small but noticeable differences between the different heads regarding the SQ and RQ metrics, with EffSeg typically obtaining the highest segmentation quality SQ, and with PointRend++ typically obtaining the highest recall quality RQ. Also here, it is currently unclear to us why these small differences emerge, with further experimentation needed to gain a better understanding of these empircal results.

\begin{table}
    \centering
    \caption[EffSeg - Computational cost metrics (panoptic head only)]{Computational cost metrics of the panoptic segmentation heads alone. The relative metrics (three rightmost columns) are \wrt to RefineMask++. We highlight the EffSeg relative metrics in bold (not necessarily the best).}
    \vspace{0.2cm}
    \resizebox{0.9\columnwidth}{!}{
    \begin{tabular}{c|c c c|c c c}
        \toprule
         \multirow{2}{*}{Seg. head} & \multirow{2}{*}{Params} & \multirow{2}{*}{FLOPs} & \multirow{2}{*}{FPS} & Params & FLOPs & FPS \\
         & & & & decrease & decrease & gain \\
         \midrule
         Mask R-CNN++ & $3.0$ M & $26.2$ G & $40.4$ & $55$\% & $87$\% & $164$\% \\
         PointRend++ & $3.3$ M & $63.5$ G & $15.7$ & $50$\% & $69$\% & $0$\% \\
         RefineMask++ & $6.6$ M & $204.6$ G & $15.3$ & $0$\% & $0$\% & $0$\% \\
         EffSeg (ours) & $4.2$ M & $65.8$ G & $17.9$ & $\mathbf{36}$\textbf{\%} & $\mathbf{68}$\textbf{\%} & $\mathbf{17}$\textbf{\%} \\
         \bottomrule
    \end{tabular}}
    \label{tab:ch5_eff_pan}
\end{table}

\paragraph{Efficiency.} In Table~\ref{tab:ch5_pan_seg}, we report the computational cost metrics of the different panoptic segmentation models \textit{as a whole}. However, to get a better understanding of the actual computational cost generated by the different panoptic segmentation heads, we also report the cost metrics of the panoptic heads \textit{alone} in Table~\ref{tab:ch5_eff_pan}. Here, we consider both the absolute cost metrics of the different heads, as well as relative cost metrics measured \wrt the RefineMask++ head. Note that these cost metrics were obtained by applying the panoptic segmentation heads on the FQDetV2 detector with $300$ queries, with differences in the detector or in the number of queries only resulting in minor changes.

From Table~\ref{tab:ch5_eff_pan}, we observe that the Mask R-CNN++ head has the lowest computational cost as expected, given it only predicts a $28\times28$ mask inside the \gls{RoI} instead of a $112\times112$ mask. We also can see that once again the RefineMask++ head is computationally the most expensive one, followed by the PointRend++ and EffSeg heads with cost metrics lying between the Mask R-CNN++ and RefineMask++ cost metrics. More specifically, the EffSeg head uses $68$\% fewer inference FLOPs and is $17$\% faster during inference compared to the RefineMask++ head. Moreover, the PointRend++ head also uses $69$\% fewer FLOPs during inference, but is unable to run faster compared to RefineMask++. Additionally, the GPU memory usage during inference is much higher for the PointRend++ head compared to the EffSeg head (not shown in Table~\ref{tab:ch5_eff_pan}), with the former having a peak GPU memory usage of 4.2~GB and the latter a peak GPU memory usage of only 3.0~GB.

\paragraph{Performance vs. Efficiency.} When we consider both the performance and the efficiency metrics, both the PointRend++ and EffSeg heads come out on top. On the one hand, they clearly outperform the Mask R-CNN++ head, while adding only a limited amount of computational cost. On the other hand, the PointRend++ and EffSeg heads reach similar performance as the RefineMask++ head, whilst clearly being more cost-effective. Whether the PointRend++ head should be preferred or the EffSeg head, is arguable, as both of their performance and cost metrics are very similar. However, we believe that the EffSeg head has a small edge over the PointRend++ head, as EffSeg runs faster and has a significantly lower (peak) GPU memory usage.

\paragraph{Qualitative results.} In what follows, we provide some qualitative results to visually assess the panoptic segmentation performance. For this, we use the panoptic segmentation models containing the FQDetV2 detector, using $300$~queries, and trained for $24$~epochs.

\begin{figure}
    \centering
    \includegraphics[scale=0.185]{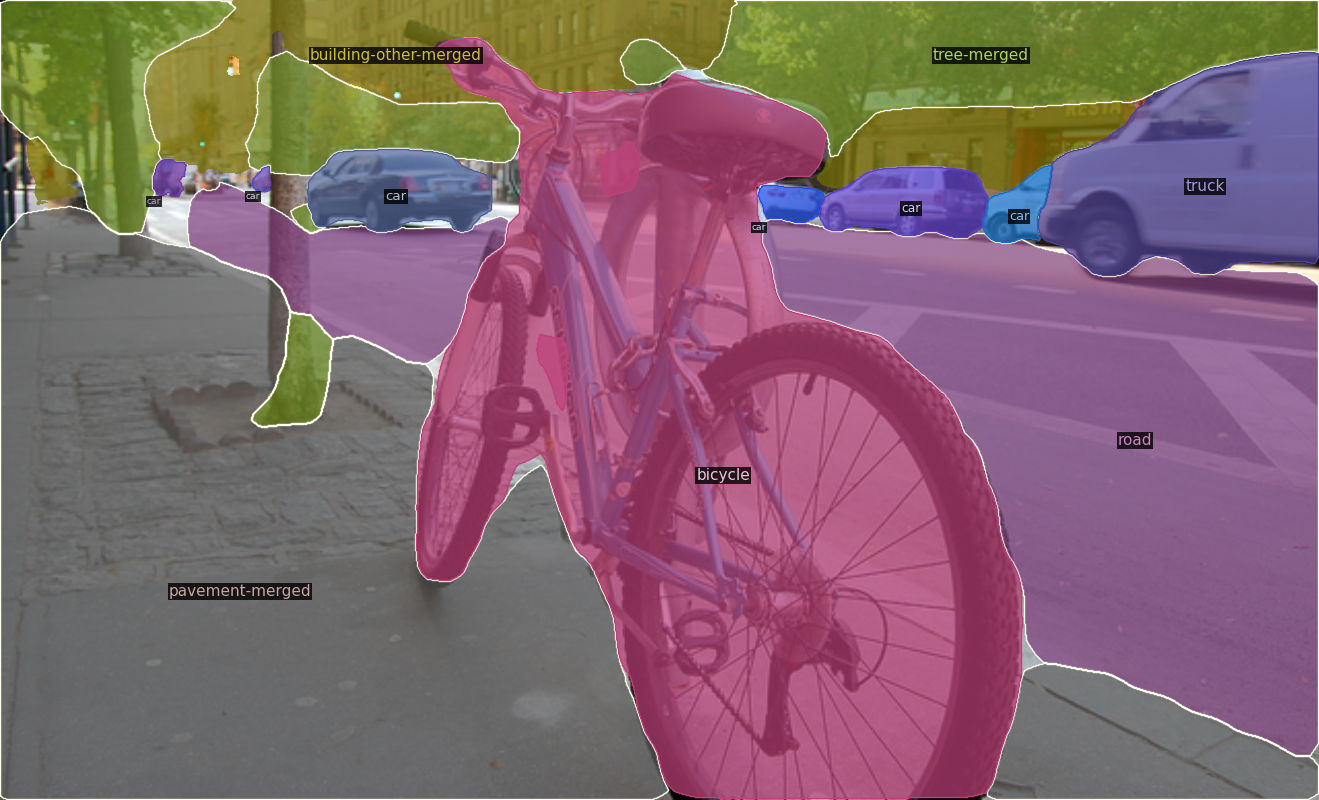} \\
    \vspace{0.2cm}
    \includegraphics[scale=0.185]{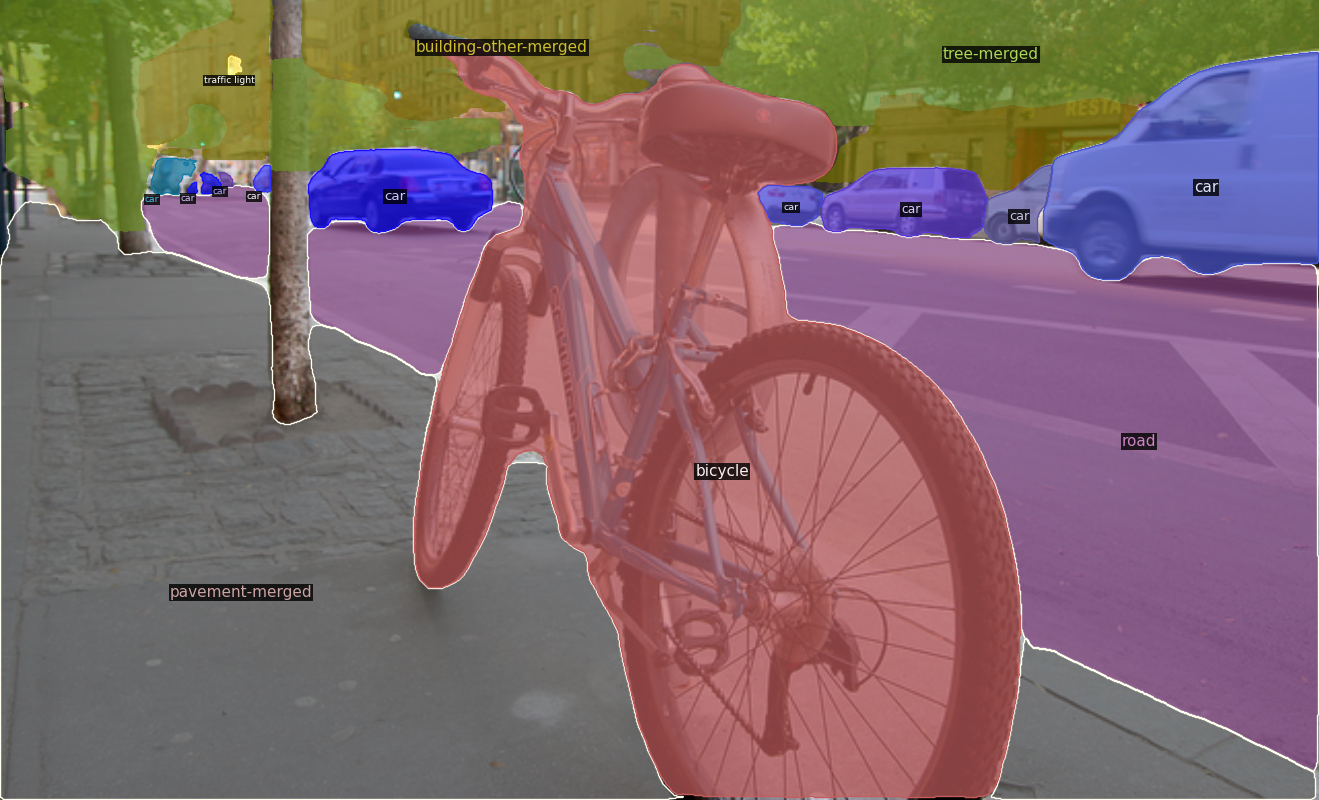} \\
    \vspace{0.2cm}
    \includegraphics[scale=0.185]{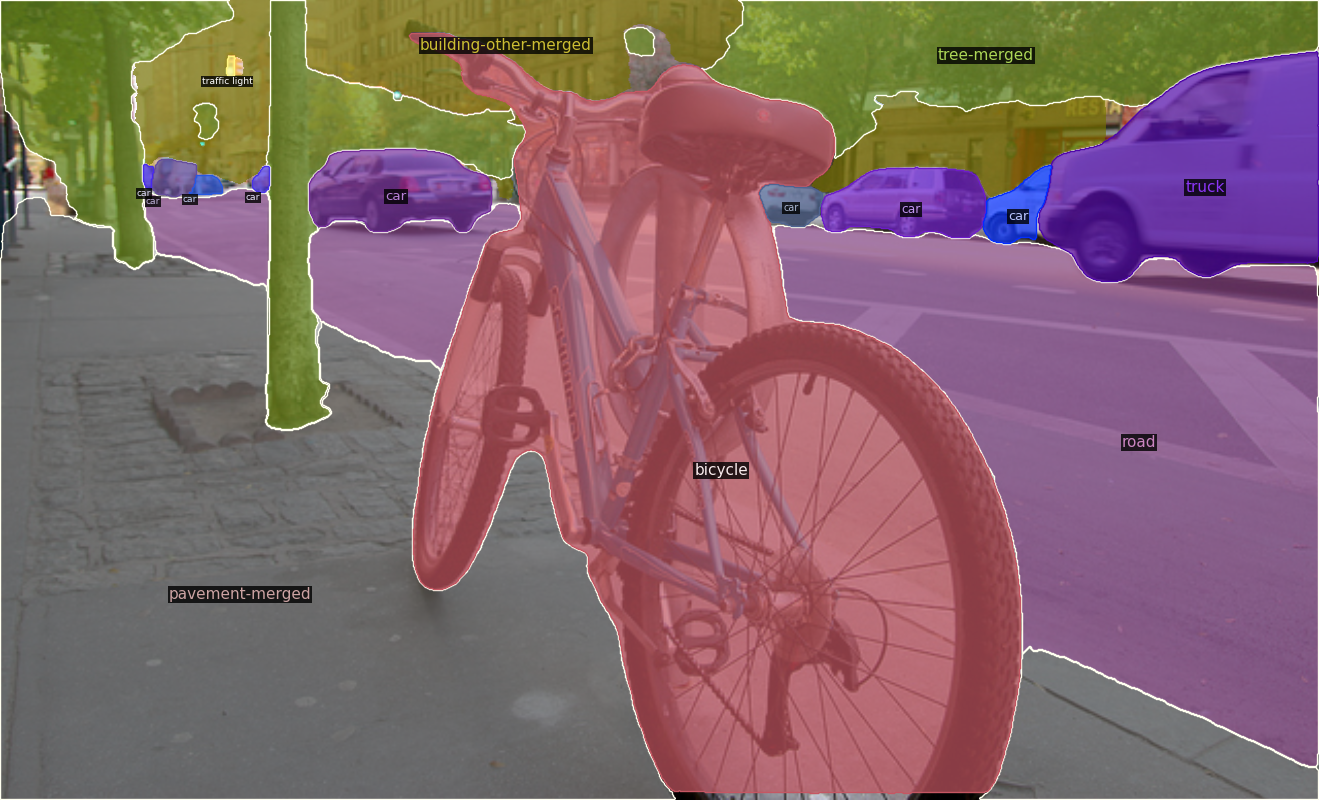}
    \caption[EffSeg qualitative (panoptic) - Example 1]{Example image with panoptic segmentation predictions using the Mask R-CNN++ (\textit{top}), RefineMask++ (\textit{middle}), and EffSeg (\textit{bottom}) segmentation heads.}
    \label{fig:ch5_pan_ex1}
\end{figure}

\begin{figure}
    \centering
    \includegraphics[scale=0.185]{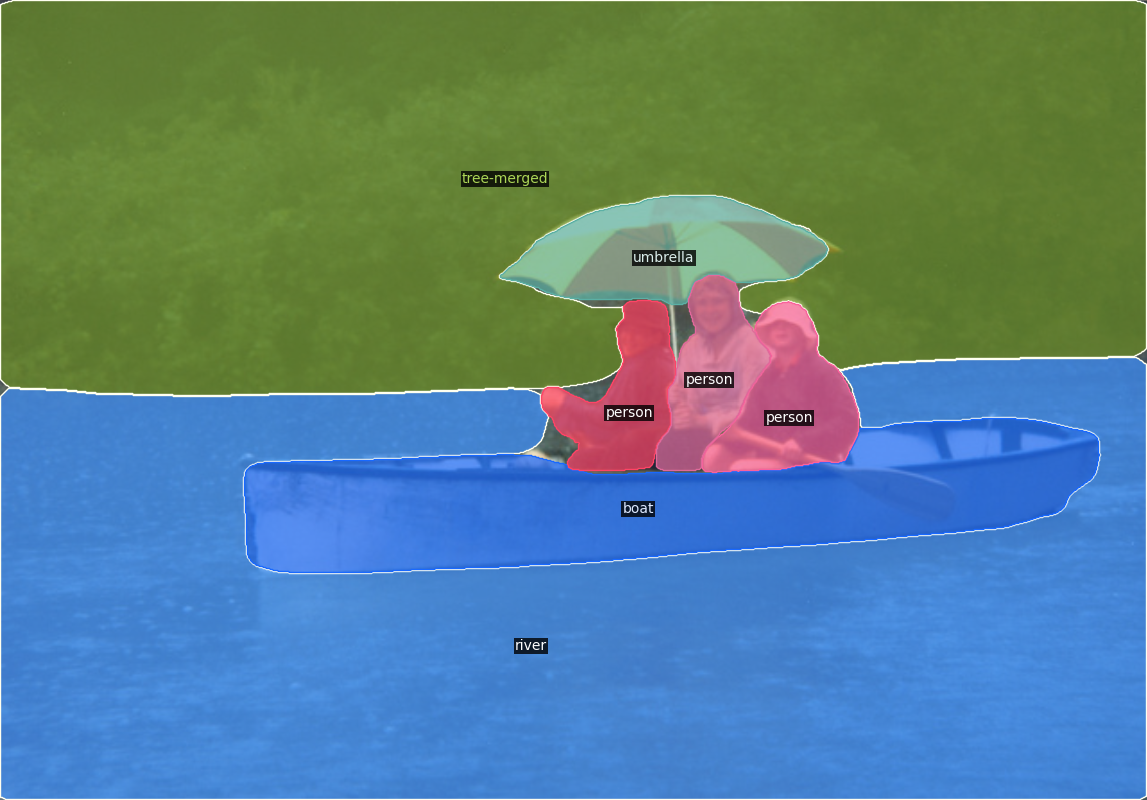} \\
    \vspace{0.2cm}
    \includegraphics[scale=0.185]{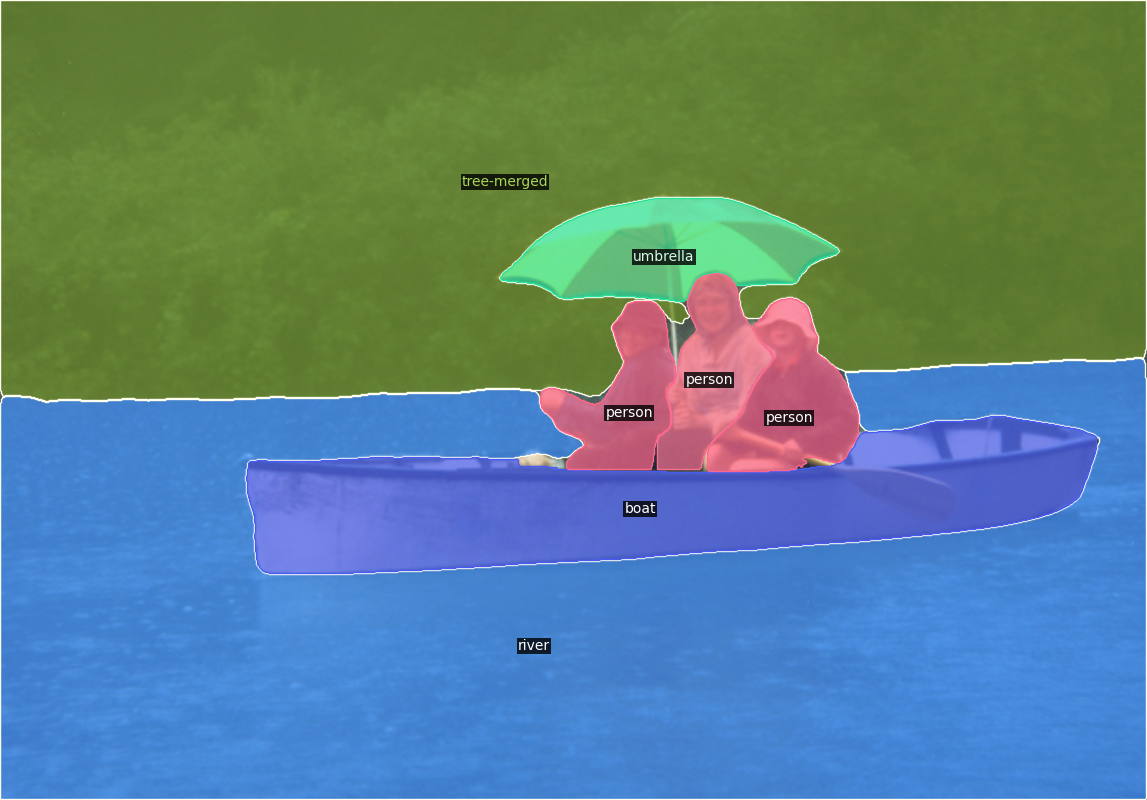} \\
    \vspace{0.2cm}
    \includegraphics[scale=0.185]{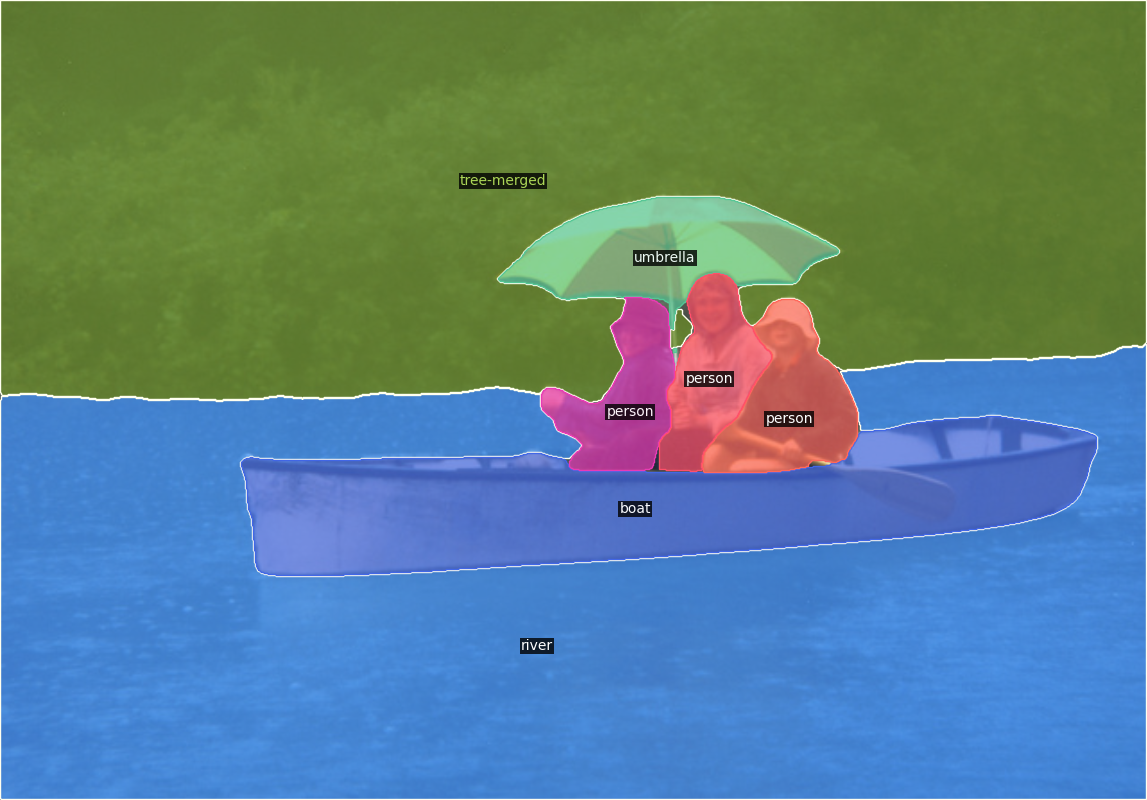}
    \caption[EffSeg qualitative (panoptic) - Example 2]{Example image with panoptic segmentation predictions using the Mask R-CNN++ (\textit{top}), RefineMask++ (\textit{middle}), and EffSeg (\textit{bottom}) segmentation heads.}
    \label{fig:ch5_pan_ex2}
\end{figure}

In Figure~\ref{fig:ch5_pan_ex1} and Figure~\ref{fig:ch5_pan_ex2}, we show two example images containing the panoptic segmentation predictions of the Mask R-CNN++ (top), RefineMask++ (middle), and EffSeg (bottom) segmentation heads. We can see that the masks produced by the Mask R-CNN++ head are rather coarse, with holes (\ie areas without prediction) near boundary regions and mask boundaries which are quite rugged. The RefineMask++ and EffSeg heads on the other hand produce more accurate segmentation masks with much smoother mask boundaries, while avoiding holes near junctions of multiple objects. Here, the difference in mask qualities can for example be assessed from the bicycle segmentation masks in Figure~\ref{fig:ch5_pan_ex1}, and from the presence (or absence) of holes near the arm of the left person in Figure~\ref{fig:ch5_pan_ex2}.

In what follows, we also provide some failure cases of the EffSeg segmentation head. Note that these failures cases are not specific to the EffSeg head and are also found in the baseline segmentation heads.

\begin{figure}
    \centering
    \includegraphics[scale=0.28]{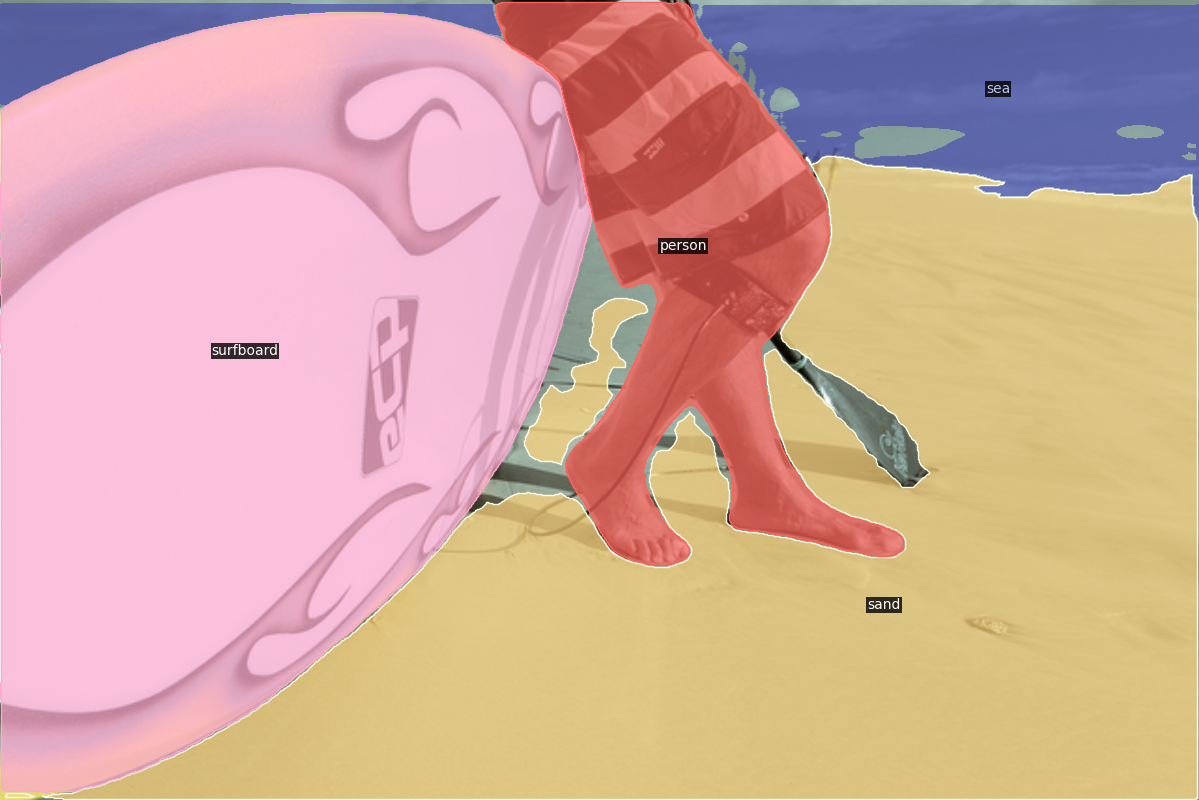} \\
    \vspace{0.4cm}
    \includegraphics[scale=0.28]{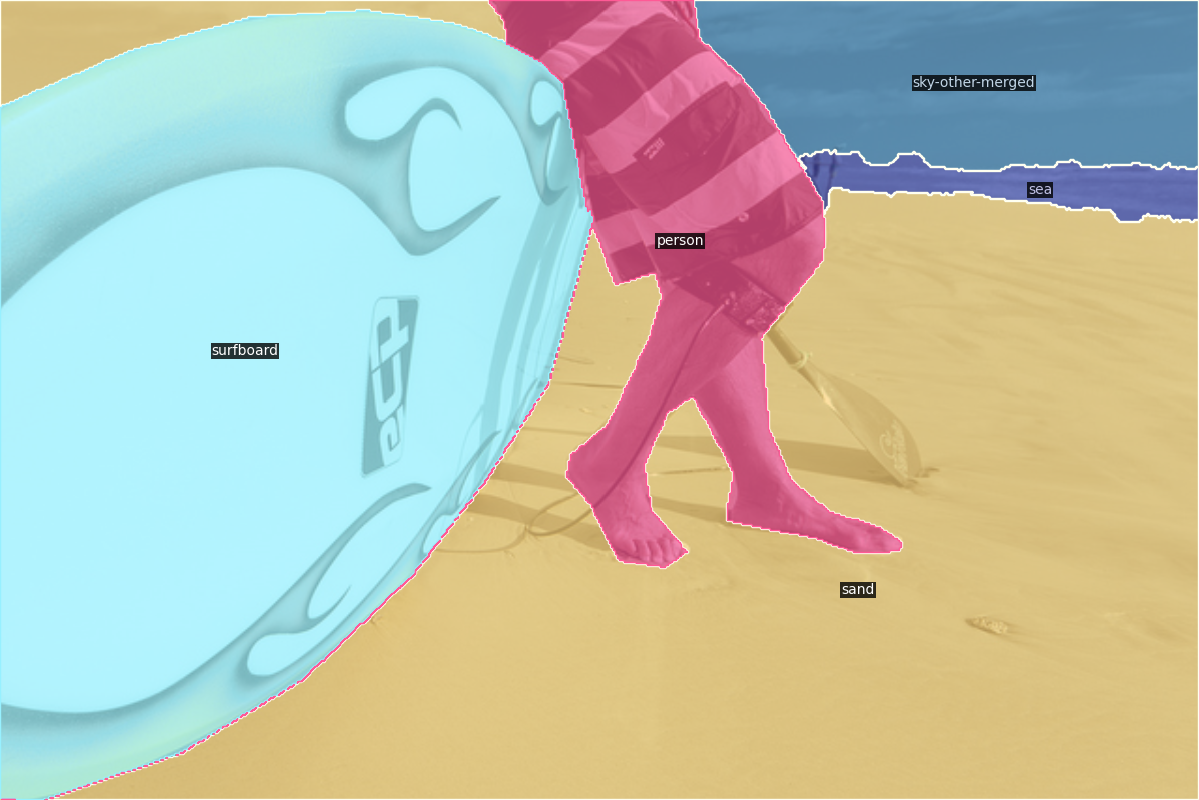}
    \caption[EffSeg qualitative (panoptic) - Incorrect stuff class]{Example image with incorrect stuff class predictions. The top image contains the predicted segmentation masks and the bottom image the ground-truth segmentation masks.}
    \label{fig:ch5_inc_stuff}
\end{figure}

In Figure~\ref{fig:ch5_inc_stuff}, we provide an example image containing incorrect stuff class predictions. Instead of separately predicting a \textit{sky} region and a \textit{sea} region, the EffSeg head predicts only a single region with the \textit{sea} class label. Moreover, the EffSeg head is also hesitant about the \textit{sand} segmentation mask, which contains various holes between the surfboard and the person. Note that the ground-truth annotations too, are far from perfect. Firstly, the ground-truth annotates the top-left region as \textit{sand}, whereas this clearly should be labeled as \textit{sky}. Secondly, we also notice that the paddle next to the person is annotated as \textit{sand}. The reason for this choice, is because there are no labels within the COCO dataset that correspond to the paddle object, and because the panoptic segmentation task requires all pixels within an image to be assigned to a class label. Hence, given that a decision for the paddle region has to be made, the annotators chose to incorporate the paddle into the \textit{sand} region. Note that the EffSeg head, also realizing that no class label corresponds to the paddle object, instead decides to leave the paddle region unlabeled (hence resulting in a hole).

\begin{figure}
    \centering
    \begin{minipage}{0.49\textwidth}
        \centering
        \includegraphics[scale=0.2]{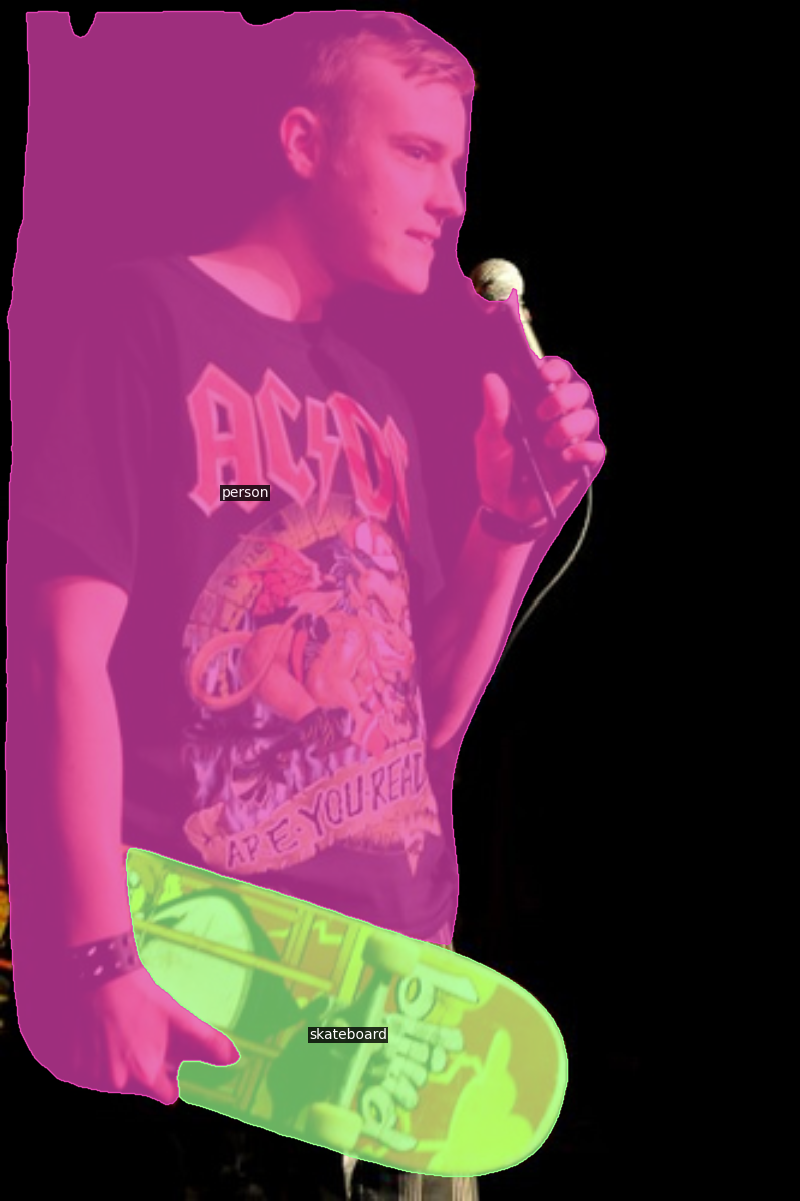}
    \end{minipage} \hfill
    \begin{minipage}{0.49\textwidth}
        \centering
        \includegraphics[scale=0.2]{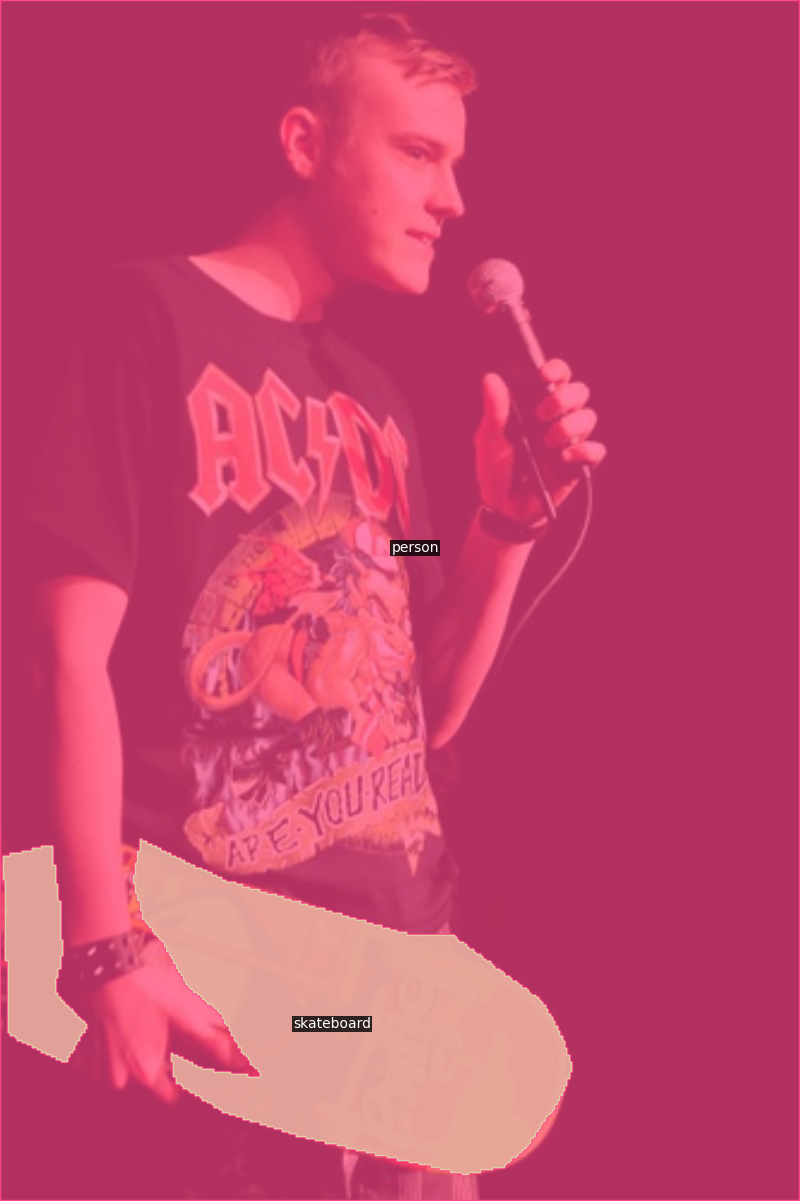}
    \end{minipage}
    \caption[EffSeg qualitative (panoptic) - Unknown background]{Example image with an unknown background. The left image contains the predicted segmentation masks and the right image the ground-truth segmentation masks.}
    \label{fig:ch5_unk_back}
\end{figure}

In some cases, it is unclear how the background should be annotated, leading to weird ground-truth panoptic segmentation masks. In Figure~\ref{fig:ch5_unk_back}, we show an example image where the background is pitch-black. Also here, no COCO class label corresponds to this background, but the panoptic segmentation task still requires a class label to be assigned to each of these background pixels. As in Figure~\ref{fig:ch5_inc_stuff}, the annotators therefore decided to extend the nearest segmentation mask to also include the background pixels, resulting in a person mask which is much larger than it actually should be. Note that the EffSeg head is hesitant to also incorporate the background region into the person mask, by adding a large part of the top-left background region to the person mask, but leaving the right part of the background region unlabeled. Previous two example images hence have shown that it can be tricky to perform panoptic segmentation when using a fixed number of predefined classes, as some image regions might not correspond to any of these class labels despite requiring one.

\subsection{Comparison between processing modules} \label{sec:ch5_comp_proc}

\begin{table*}
    \centering
    \caption[EffSeg - Comparison of processing modules]{Comparison between different EffSeg processing modules.}
    \vspace{0.2cm}
    \resizebox{\columnwidth}{!}{
    \begin{tabular}{c c|c c c c c c|c c|c c c}
        \toprule
        Seg. head & Module & AP & AP$_{50}$ & AP$_{75}$ & AP$_S$ & AP$_M$ & AP$_L$ & AP$^*$ & AP$^{B*}$ & Params & FLOPs & FPS \\
        \midrule
        EffSeg & MLP & $40.8$ & $61.0$ & $44.4$ & $19.9$ & $43.9$ & $59.3$ & $43.9$ & $33.1$ & $38.4$ M & $227.4$ G & $8.7$ \\
        \midrule
        EffSeg & Conv & $41.1$ & $61.4$ & $44.6$ & $20.4$ & $43.7$ & $60.3$ & $43.9$ & $33.1$ & $38.5$ M & $234.0$ G & $8.5$ \\
        EffSeg & DeformConv & $41.1$ & $61.2$ & $45.0$ & $20.9$ & $44.1$ & $60.0$ & $43.8$ & $32.8$ & $38.5$ M & $235.1$ G & $7.4$ \\
        EffSeg & SFM & $\mathbf{41.5}$ & $\mathbf{61.7}$ & $\mathbf{45.0}$ & $\mathbf{21.1}$ & $\mathbf{44.5}$ & $\mathbf{60.5}$ & $\mathbf{44.3}$ & $\mathbf{33.4}$ & $38.8$ M & $245.5$ G & $8.0$ \\
        \midrule
        EffSeg & Dense SFM & $41.1$ & $61.1$ & $44.9$ & $20.5$ & $44.1$ & $60.0$ & $44.1$ & $\mathbf{33.4}$ & $38.9$ M & $337.8$ G & $5.6$ \\
        \bottomrule
    \end{tabular}}
    \label{tab:ch5_comp_proc}
\end{table*}

In Table~\ref{tab:ch5_comp_proc}, we show results comparing EffSeg models with different processing modules on the instance segmentation task (see Subsection~\ref{sec:ch5_detail} for more information about the processing module). All models were trained for $12$~epochs using the ResNet-50 backbone~\cite{he2016deep}, the FPN neck~\cite{lin2017feature}, and the FQDet detector (see Chapter~\ref{ch:detection}). We make following observations.

First, we can see that the \gls{MLP} processing module performs the worst. This confirms that Pointwise networks such as MLPs yield sub-optimal segmentation performance due to their inability to access information from neighboring locations, as argued in Subsection~\ref{sec:ch5_sps}.

Next, we consider the convolution (Conv), deformable convolution~\cite{dai2017deformable} (DeformConv), and Semantic Fusion Module~\cite{zhang2021refinemask} (SFM) processing modules. We can see that the Conv and DeformConv processing modules reach similar performance, whereas the SFM processing module obtains slightly higher segmentation performance. Note that the use of the DeformConv and SFM processing modules, was enabled by our SPS method, which supports any 2D operation. This is in contrast to the Neighbors method for example (see Subsection~\ref{sec:ch5_sps}), that neither supports the DeformConv processing module, nor the SFM processing module (as it contains dilated convolutions). This hence highlights the importance of the \gls{SPS} method to support any 2D operation, allowing the use of superior processing modules such as the SFM processing module.

Finally, Table~\ref{tab:ch5_comp_proc} additionally contains the DenseSFM baseline, applying the SFM processing module over all RoI locations similar to RefineMask~\cite{zhang2021refinemask}. When looking at the results, we can see that densely applying the SFM module (DenseSFM) as opposed to sparsely (SFM), does not yield any performance gains, while dramatically increasing the computational cost. We hence conclude that no performance is sacrificed when performing sparse processing instead of dense processing.

\section{Limitations and future work}
In following paragraphs, we discuss some of the limitations of EffSeg and provide directions for future work.

\paragraph{Implementation.} The 2D operations (\eg convolutions) performed on the \gls{SPS} data structure, are currently implemented in a naive way using native PyTorch~\cite{paszke2019pytorch} operations. Instead, these operations could be implemented in CUDA, which should result in additional speed-ups for our EffSeg models.

\paragraph{Active feature locations.} In EffSeg, a separate refinement branch is used to identify the spatial locations for which additional computation is performed (\ie the active feature locations). In PointRend~\cite{kirillov2020pointrend}, the active feature locations are computed based on the segmentation mask uncertainties, and in RefineMask~\cite{zhang2021refinemask} the active feature locations are determined based on the boundaries of the predicted (and ground-truth) segmentation masks. Given these varying methodologies, it would be interesting to know which approach (if any) works best.

\paragraph{Dataset overfitting.} All of the segmentation experiments were performed on the COCO dataset~\cite{lin2014microsoft}. Because of this, some of the EffSeg hyperparameters might be specifically tuned towards the COCO dataset and hence might not generalize well to other segmentation datasets. This could for example be the case for the hyperparameter controlling the number of active features per refinement stage, as the optimal value might depend on the typical number of objects present in images and on the average mask complexity. As future work, it would therefore be useful to perform experiments on other segmentation datasets such as ADE20k~\cite{zhou2017scene}, BDD100K~\cite{yu2020bdd100k}, and Cityscapes~\cite{cordts2016cityscapes}.

Additionally, note that the mask annotations provided by the COCO segmentation benchmark~\cite{lin2014microsoft} are not always of high quality (see for example the ground-truth mask of the skateboard in Figure~\ref{fig:ch5_amb_seg}), which limits the potential of the trained fine-grained segmentation heads. However, by considering additional segmentation datasets with higher quality mask annotations, the true potential of fine-grained segmentation heads could be extracted, possibly leading to bigger performance gaps between the different heads.

\paragraph{Metric overfitting.} The hyperparameters of the EffSeg head were tuned \wrt the \gls{AP} metric on the COCO instance segmentation benchmark. The found hyperparameters hence might not generalize well to other metrics, such as the \gls{PQ} metric used to evaluate the panoptic segmentation performance. Therefore, it would be useful as future work to investigate which (if any) hyperparameters are sensitive to this change in metric, and how big of an impact these have on the final performance.

Additionally, note that the PQ metric (strongly) penalizes false positives, as opposed to the AP metric which is quite lenient towards this type of error (see also Subsection~\ref{sec:ch4_limit_future} for a discussion on this topic). As the used FQDet and FQDetV2 detectors from Chapter~\ref{ch:detection} are trained with the AP metric, these detectors have no incentive to remove (low-scoring) false positive detections and hence keep them in. Therefore, in order to remove these false positives for the PQ metric, some post-processing heuristics are introduced, such as removing all detections with a classification score below~$0.3$. As future work, it would be interesting to investigate whether the FQDet heads could learn to remove these false positives automatically, similar to the way the FQDet heads can be enhanced to learn about duplicate detections. This would not only deprecate some of the post-processing heuristics and possibly increase the PQ performance, but also render the detector better suited towards practical applications, for which the removal of false positives is often needed.

\paragraph{RoI-based panoptic segmentation.} In this chapter, we performed panoptic segmentation using RoI-based instance segmentation heads. Here, stuff classes are treated in exactly the same way as thing classes, by first predicting a bounding box and by then segmenting between foreground and background within this box. Surprisingly, no results have been reported in the literature following this approach, implying that RoI-based panoptic segmentation methods do not work. Instead, methods are proposed processing thing and stuff classes separately~\cite{kirillov2019panoptic, kirillov2019panoptic2}, or one-stage segmentation methods~\cite{cheng2022masked, li2022mask} which use queries for both thing and stuff classes. However, we show that RoI-based panoptic segmentation methods can also work, giving reasonable panoptic segmentation results. As future work, additional experiments could be performed to compare RoI-based panoptic segmentation methods with existing approaches, in order to determine which approach yields the best performance vs. efficiency trade-off. 

\paragraph{Large-scale experiments.} All experiments in this chapter were performed with the ResNet-50~\cite{he2016deep} backbone and the FPN~\cite{lin2017feature} neck. Large-scale experiments with the Swin-L~\cite{liu2021swin} backbone and the DefEncoder~\cite{zhu2020deformable} neck, are left as future work. Moreover, note that in order to be consistent with \gls{SOTA} segmentation methods such as Mask2Former~\cite{cheng2022masked} and Mask DINO~\cite{li2022mask}, these large-scale experiments should be carried out with the large-scale jittering (LSJ) data augmentation scheme~\cite{ghiasi2021simple}, yielding better performance when training for 20~epochs or more~\cite{cheng2022masked}. 

\paragraph{Additional improvements.} In what follows, we propose three modifications to the EffSeg head, possibly leading to increased segmentation performance. Experimenting with these modifications is left as future work. Firstly, the size of the used RoI boxes can be increased with a fixed factor, in order to increase the probability the RoI box fully covers the object to segment, avoiding incomplete segmentation masks as in Figure~\ref{fig:ch5_inc_box}. Secondly, the EffSeg head could get rid of RoI boxes altogether (\ie of the two-stage approach), and make segmentation predictions over the whole image for each object (\ie the one-stage approach), similar to more recent works~\cite{cheng2022masked, li2022mask}. Thirdly, instead of fixing the amount of computation done at each resolution (\eg the amount of computation done at the $14\times14$ RoI resolution is always the same), this amount could be determined dynamically with an uncertainty-based, a boundary-based, or a reward-based mechanism.

\section{Conclusion}
In this work, we propose EffSeg performing fine-grained instance and panoptic segmentation in an efficient way by introducing the Structure-Preserving Sparsity (SPS) method. SPS separately stores active features, passive features, and a dense 2D index map containing the feature indices, resulting in computational and storage-wise efficiency, while supporting any 2D operation. EffSeg obtains similar segmentation performance as the highly competitive RefineMask head, while reducing the number of FLOPs up to $70$\% and increasing the FPS up to $44$\%.

\cleardoublepage
  \graphicspath{{\chaptersdir/conclusion/image/}}%
  \chapter{Conclusion} \label{ch:conclusion}
In this chapter, we give an overview of the work done throughout the thesis and highlight the main contributions. We conclude by listing some limitations of the work and provide directions for future work.

\section{Overview}
In \textbf{Chapter~\ref{ch:introduction}}, we started off by outlining the research setting and our main research goal, which consists of designing high-performing networks for multi-scale computer vision tasks such as object detection and segmentation. To asses the quality of the designed networks, we compare them with competitive baseline networks, where the goal with our designs is to lie top left \wrt to each of the baseline designs in the performance vs. cost graph. Moreover, we also highlighted the importance of correctly comparing with the baseline networks, by ensuring the used deep learning settings for our designs and the baseline designs are the same.

In \textbf{Chapter~\ref{ch:background}}, we laid the foundations of our work by introducing (1) the backbone-neck-head meta architecture, (2) the used multi-scale computer vision tasks and datasets, (3) the used performance and computational cost metrics, and (4) the standardized deep learning settings.

In \textbf{Chapter~\ref{ch:neck}}, we designed our own neck network called Trident Pyramid Network (\gls{TPN}). The TPN neck consists of multiple TPN layers, where each layer consists of sequential top-down and bottom-up operations, interleaved with parallel self-processing operations. Different from existing necks such as BiFPN~\cite{tan2020efficientdet}, our TPN neck is equipped with hyperparameters controlling the amount of communication-based processing and self-processing, enabling our TPN neck to find the optimal balance between both types of processing. Additionally, we also investigate to what extent it is more beneficial perform computation at the backbone-level, neck-level, or head-level of the backbone-neck-head meta architecture.

In \textbf{Chapter~\ref{ch:detection}}, we designed our own object detection heads (or detectors) called FQDet and FQDetV2. The FQDet heads are query-based two-stage object detection heads which combine the best of both classical and DETR-based detectors. Among other things, the FQDet heads successfully introduce the use of anchors within the query-based paradigm and the static top-k matching procedure. Extensive experiments show the effectiveness of the FQDet heads, reaching excellent object detection performance after few training epochs.

In \textbf{Chapter~\ref{ch:segmentation}}, we designed our own segmentation head called EffSeg. The EffSeg head produces both fine-grained instance and panoptic segmentation predictions in an efficient way, by leveraging our novel Structure-Preserving Sparsity (\gls{SPS}) method. Instead of performing computation at all locations within a dense 2D feature map, the SPS method processes only a specific subset of spatial locations, by separately storing the active features (\ie the features to be updated), the passive features, and a dense 2D index map. Here, the 2D index map stores the active and passive feature indices, preserving the 2D feature map structure such that any 2D operation on the active features can still be performed. Various experiments are conducted on both the instance segmentation and the panoptic segmentation tasks, validating the effectiveness of the SPS method and of the EffSeg head as a whole.

\section{Contributions}
In following paragraphs, we highlight the main contributions of the thesis.

\paragraph{Hyperparameters to balance neck.} In Chapter~\ref{ch:neck}, we introduced our own neck called \gls{TPN}. Different from existing necks such as BiFPN~\cite{tan2020efficientdet}, the TPN neck contains hyperparameters to balance the neck communication-based processing and self-processing. We have empirically shown that when these hyperparameters are carefully chosen, better object detection performance can be obtained without an increase in computational cost.

\paragraph{More neck computation.} In Chapter~\ref{ch:neck}, we additionally investigated whether it is more beneficial to put computation in the backbone, neck, or head part of a computer vision network. Experiments have shown that the neck part reaches the highest performance gains for a set amount of additional compute. This result raises the question whether it would not be more useful for the research community to put more effort into designing high-performing neck networks, rather than almost exclusively focusing on backbone networks.

\paragraph{Use of anchors.} In Chapter~\ref{ch:detection}, we introduced the \gls{FQDet} and \gls{FQDetV2} object detection heads. One of the keys behind their success, is the introduction of anchors within the query-based paradigm. The use of anchors increases the accuracy of the bounding box predictions and improves the inductive bias of the detector leading to faster convergence.

\paragraph{Static top-k matching.} In Chapter~\ref{ch:detection}, we also introduced our static top-k matching scheme, which is used in both stages of the FQDet heads. The static top-k matching scheme produces stable assignments since the beginning of the training process, leading to increased convergence speeds compared to dynamic matching schemes such as Hungarian matching.

\paragraph{More queries.} In Chapter~\ref{ch:detection}, we also performed a series of design experiments aiming to improve the FQDet head and eventually resulting in the FQDetV2 head. One of the main improvements came from using more queries, which significantly increases the detection performance at a limited computational cost.

\paragraph{SPS method.} In Chapter~\ref{ch:segmentation}, we introduced our fine-grained segmentation head called \gls{EffSeg}. In order to produce fine-grained segmentation masks in an efficient way, we introduced the \gls{SPS} method, which allows to only update a specific set of features (called active features) from a 2D feature map using any 2D operation of choice. The SPS method achieves this by separately storing the active features, the passive features, and a 2D index map, with the latter storing the 2D feature structure such that any 2D operation can still be performed.

\section{Limitations and future work}
In following paragraphs, we discuss some of the main limitations of our work and provide directions for future research.

\paragraph{TPN operations.} In Chapter~\ref{ch:neck}, we introduced the TPN neck, which uses top-down, self-processing, and bottom-up operations. During our experiments, we used one specific implementation for each type of operation, even though different implementations could have been chosen. The self-processing operation for example uses a convolution-based implementation, even though an attention-based implementation would also have been a possibility. It would hence be interesting to the experiment with different types of implementations for these operations to evaluate which ones work best.

\paragraph{FQDet duplicate removal.} In Chapter~\ref{ch:detection}, we introduced the FQDet heads. As duplicate removal mechanism, the FQDet heads still rely on the hand-crafted \gls{NMS} method. Despite the effectiveness of the NMS method, it would be more desirable that the FQDet heads learn to remove the duplicates themselves. Preliminary results show that the learned duplicate removal mechanism obtains competitive results, but still slightly lags behind the NMS method. Additional research and experimentation is hence required to further improve the learned duplicate removal mechanism and deprecate the NMS method once and for all.

\paragraph{EffSeg implementation.} In Chapter~\ref{ch:segmentation}, we introduced the EffSeg segmentation head, which performs fine-grained segmentation in an efficient way thanks to the SPS method. While the EffSeg head uses significantly fewer FLOPs compared to its main baseline RefineMask++, the corresponding increase in FPS is somewhat limited. The main reason for this is because RefineMask++ exclusively uses common data structures, such as dense 2D feature maps, for which highly optimized CUDA kernels exist, whereas EffSeg uses the SPS data structure for which such highly optimized CUDA kernels do not exist. Hence, implementing CUDA kernels for the SPS data structure would fully extract the true potential of the SPS method, and as a result reflect the \textit{true} FPS count of the EffSeg head.

\paragraph{Additional comparisons.} Our network designs could be compared with additional baselines, possibly leading to new insights. In Chapter~\ref{ch:neck}, we compared the TPN neck with the closely related BiFPN neck~\cite{tan2020efficientdet}, but a comparison with the DefEncoder neck~\cite{zhu2020deformable} is missing. Given that the TPN neck uses separate top-down, self-processing, and bottom-up operations, as opposed to the all-in-one operation from the DefEncoder neck, a comparison between these two competing approaches would be interesting. In Chapter~\ref{ch:segmentation}, experiments are also missing comparing the EffSeg head with the recent Mask2Former~\cite{cheng2022masked} and Mask DINO~\cite{li2022mask} segmentation heads.

\paragraph{Additional datasets.} All of our experiments were performed on the COCO dataset~\cite{lin2014microsoft}. Because of this, all of our network designs and especially their hyperparameters, are unwillingly tailored towards the type of images found in the COCO dataset. Hence, it would be useful to perform experiments on additional datasets, such as ADE20k~\cite{zhou2017scene}, BDD100K~\cite{yu2020bdd100k}, CityScapes~\cite{cordts2016cityscapes}, and Objects365~\cite{shao2019objects365}.

\paragraph{Additional metrics.} All of our object detection and instance segmentation experiments were evaluated using the \gls{AP} performance metric, and all of our panoptic segmentation experiments were evaluated using the \gls{PQ} performance metric. Apart from the dataset, performance metrics also leave an important mark on the obtained network designs, as they eventually determine what kind of errors are penalized the most. The AP metric for example only slightly penalizes false positives, whereas these errors are often unwanted in practical applications and hence possibly deserve a bigger penalty. It would therefore be a good idea to evaluate our network designs using additional performance metrics.

\paragraph{Additional tasks.} During our thesis, we considered the object detection, instance segmentation, and panoptic segmentation multi-scale computer vision tasks. Techniques developed throughout our thesis such as the SPS method, could be applied to other tasks though, such as the 3D instance segmentation~\cite{wang2018sgpn} and the video instance segmentation~\cite{yang2019video} tasks.

\cleardoublepage
%


\backmatter

\includebibliography
\bibliographystyle{acm}
\bibliography{allpapers}

\includecv{curriculum}
\includepublications{publications}

\makebackcoverXII

\end{document}